\documentclass[lettersize,journal]{IEEEtran}
\usepackage{amsmath,amssymb,amsfonts}
\usepackage{algorithm}
\usepackage{algpseudocode} 
\usepackage{array}
\usepackage{textcomp}
\usepackage{stfloats}
\usepackage{url}
\usepackage{verbatim}
\usepackage{graphicx}
\usepackage{cite}
\usepackage{wrapfig}
\usepackage{caption}
\usepackage{subfig}  
\usepackage{tabularx}
\usepackage[pagebackref,breaklinks,colorlinks]{hyperref}

\begin{document}

\pdfoutput=1

\title{TTS-CGAN: A Transformer Time-Series Conditional GAN for Biosignal Data Augmentation}

\author{Xiaomin Li\href{https://orcid.org/0000-0002-5738-8873}{\includegraphics[scale=0.005]{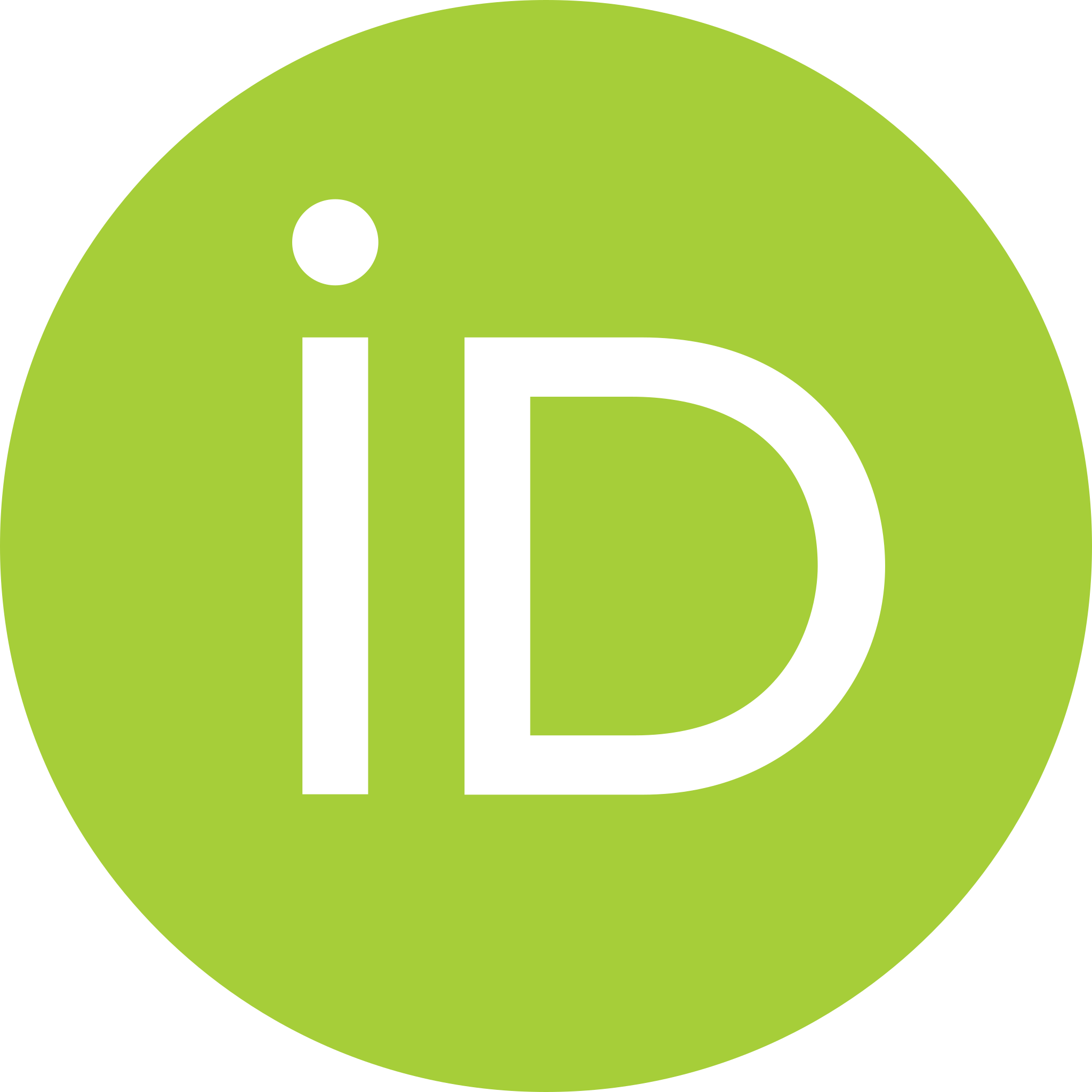}}, 
Anne Hee Hiong Ngu \href{https://orcid.org/0000-0002-5877-0230}{\includegraphics[scale=0.005]{images/ORCID_iD.png}} IEEE Member,
Vangelis Metsis\href{https://orcid.org/0000-0002-7371-8887}{\includegraphics[scale=0.005]{images/ORCID_iD.png}}
\thanks{Submission date: 06/21/2022}
\thanks{Xiaomin Li, Ph.D. candidate, e-mail:x\_l30@txsatate.edu}
\thanks{Anne Hee Hiong Ngu, Professor, e-mail: angu@txstate.edu}
\thanks{Vangelis Metsis, Associate Professor, e-mail: vmetsis@txstate.edu}
\thanks{Xiaomin Li and Vangelis Metsis are members of the Intelligent Multimodal Computing and Sensing (IMICS) Lab.
All authors are from the department of computer science, Texas State University. Address: 601 University Drive, San Marcos, TX, USA, 78666}}



\maketitle

\begin{abstract}
Signal measurement appearing in the form of time series is one of the most common types of data used in medical machine learning applications. Such datasets are often small in size, expensive to collect and annotate, and might involve privacy issues, which hinders our ability to train large, state-of-the-art deep learning models for biomedical applications. For time-series data, the suite of data augmentation strategies we can use to expand the size of the dataset is limited by the need to maintain the basic properties of the signal. Generative Adversarial Networks (GANs) can be utilized as another data augmentation tool. 
In this paper, we present TTS-CGAN, a transformer-based conditional GAN model that can be trained on existing multi-class datasets and generate class-specific synthetic time-series sequences of arbitrary length. We elaborate on the model architecture and design strategies. Synthetic sequences generated by our model are indistinguishable from real ones, and can be used to complement or replace real signals of the same type, thus achieving the goal of data augmentation. 
To evaluate the quality of the generated data, we modify the wavelet coherence metric to be able to compare the similarity between two sets of signals, and also conduct a case study where a mix of synthetic and real data are used to train a deep learning model for sequence classification. Together with other visualization techniques and qualitative evaluation approaches, we demonstrate that TTS-CGAN generated synthetic data are similar to real data, and that our model performs better than the other state-of-the-art GAN models built for time-series data generation. 
 
 TTS-CGAN source code: \href{https://github.com/imics-lab/tts-cgan}{github.com/imics-lab/tts-cgan}
\end{abstract}


\section{Introduction}\label{sec:introduction}

Shortage of training data is often a problem when analyzing health domain time-series data with deep learning models. Deep learning applications have thrived on image and text data used in computer vision (CV) and natural language processing (NLP) tasks, which are abundant on the internet, easy to collect from users' daily activities, and can be annotated with little trained labor. However, time-series data collected as sensor measurements resulting from physical or biological processes, especially when such processes involve human subjects need much more effort to collect, annotate, and interpret. For example, collecting the Electroencephalography (EEG) data used to record human brain function involves a costly endeavor and needs neuroscience experts to add annotations. Besides, data sharing in medical fields also has very strict privacy constraints which further impedes machine learning researchers from obtaining abundant medical data. Nonetheless, deep learning models require large amounts of data to train successfully. Training deep learning models on small datasets will result in over-fitting and lower generalization capabilities. As a compromise, researchers are forced to train shallower deep learning models that are not capable of capturing the full complexity of the problem at hand. In the work~\cite{lawhern2018eegnet}, the author designed two CNN models with only two and four convolution layers to classify EEG signals. The work~\cite{li2022spp} leveraged transfer learning and trained an EEG feature extractor with a vast amount of signals but still could only use a CNN model with six convolution layers. Compared with the deep learning model used in Computer Vision tasks, such models have much fewer layers and simpler designs. Because of the data shortage problem and inter-subject variance, designing a more complex deep learning model cannot help to improve the models' representation abilities on these signals. This is a common situation encountered in medical and health-related machine learning research.

Training deep learning models on multi-class imbalanced datasets will cause biased model performance in different classes~\cite{ramyachitra2014imbalanced}. However, since the probabilities of different events happening are naturally different, data imbalance is another common phenomenon that exists in multi-class medical datasets. For example, the abnormal heartbeat signals caused by diseases are rare and harder to collect compared to normal heartbeats signals, but they are essential for disease diagnosis. In practice, researchers have made various efforts to reduce such data imbalances to produce bias-free deep learning model outputs, such as undersampling the majority classes or oversampling minority classes, setting higher class weights for minority classes, and changing evaluation metrics, etc. Although these methods can relatively eliminate imbalances, they may lead to information loss that degrades the learning abilities of deep learning models. The survey paper~\cite{johnson2019survey} had thoroughly addressed this concern and listed several traditional methods and deep learning methods to relieve the data imbalance issues.   

Generative Adversarial Networks (GANs), first introduced in 2014~\cite{goodfellow2014generative}, have been gaining traction in the deep learning research field. They have successfully generated and manipulated data in Computer Vision (CV)  and Natural Language Processing (NLP) domains, such as high-quality image generation~\cite{ledig2017photo}, style transfer~\cite{bousmalis2017unsupervised}, text-to-image synthesis~\cite{zhang2017stackgan}, etc. Since GAN models can generate new synthetic data from a limited amount of data, the above-mentioned data shortage problems can be alleviated by adding reproducible synthetic data. Recently, there has been a movement toward using GANs for time-series and sequential data generation. The review paper~\cite{brophy2021generative} gives a thorough summary of GAN implementations on time-series data. It introduced the benefits of using GAN as a time-series data augmentation tool. For example, it can be used to solve data shortage issues by augmenting smaller datasets and generating new, previously unseen data. It can be used to recover missing or corrupted data and reduce data noise. It also protects data privacy by generating differentially private datasets that have no sensitive information from the source datasets. The paper also lists several state-of-the-art GAN models and algorithms for generating time-series data, such as C-RNN-GAN~\cite{mogren2016c}, RCGAN~\cite{esteban2017real}, TimeGAN~\cite{yoon2019time}, SigCWGAN~\cite{ni2020conditional}. They are all using recurrent neural networks (RNNs) as the base architecture of their GAN models. However, the RNN-based GAN models have difficulties producing long synthetic sequences that are realistic enough to be useful. This is due to the fact that the time-steps of a time-series are processed sequentially and thus more recent time-steps have a greater effect on the generation of the next time-steps. In other words, RNNs have difficulties establishing relationships between distant time-steps within a long sequence.

The transformer architecture, which relies on multiple self-attention layers~\cite{vaswani2017attention}, has recently become a prevalent deep learning model architecture. It has been shown to surpass many other popular neural network architectures, such as CNNs over images and RNNs over sequential data~\cite{dosovitskiy2020image}\cite{devlin2018bert} in classification tasks, and it has even displayed properties of a universal computation engine~\cite{lu2021pretrained}. Some works have already tried to utilize the transformer model in GAN model architecture design with the goal of either improving the quality of synthetic data or creating a more efficient training process~\cite{jiang2021transgan}\cite{diao2021tilgan} for image and text generation tasks. The work~\cite{jiang2021transgan}, for the first time, built a pure transformer-based GAN model and verified its performance on multiple image synthesis tasks. 

Since the transformer was invented to handle long sequences of text data and does not suffer from a vanishing gradient problem, theoretically, a transformer GAN model should perform better than RNN-based models on time-series data. Our previous work~\cite{li2022tts} introduced a transformer-based GAN model (TTS-GAN) to generate synthetic time-series data. In that approach, a separate GAN model is trained for each class of the dataset. A disadvantage of that approach is that if for some of the classes we only have very few training instances, it is difficult to train a GAN model to generate realistic sequences for those classes. 

In this paper, we propose a conditional GAN for time-series generation, which we call TTS-CGAN. TTS-CGAN is trained on data from all classes at the same time and we can control which class to generate data for by priming it with the right input. The benefit of that is that TTS-CGAN can take advantage of transfer learning effects between classes. Essentially, instead of training a separate model for each class, we are training a model on the whole dataset, thus allowing it to learn better low-level feature representations, while high-level features can be fine-tuned simultaneously for each class at the deeper layers of the network. We demonstrate the efficacy of TTS-CGAN with new similarity metrics and experiments showcasing the effect of synthetic data augmentation on classification tasks.

Time-series data are not easily interpretable by humans and there are no standard metrics used by researchers to compare the similarity of GAN-generated time-series data with real data. With a thorough literature review, we find that most of the existing signal similarity metrics can not maximally distinguish the differences between two sets of signals. To evaluate the similarity between real and synthetic signals, we need a metric that mostly meets the high inter-class variance and low intra-class variance standard. For this reason, we present a modified wavelet coherence~\cite{grinsted2004application} metric to compute the similarity between two sets of signals. Together with some dimensionality reduction and visualization methods we are able to demonstrate both qualitatively and quantitatively that our model produces better quality synthetic signals than other existing models.

Our contributions can be summarized as follows:
\begin{itemize}
    \item We introduce TTS-CGAN, a conditional GAN model that can generate multi-class time-series data with labels, and we conduct an empirical study about what is the best strategy to embed labels to GAN models.
    \item We modify the wavelet coherence metrics to be able to quantitatively compare the similarity between two sets of signals. 
    \item We evaluate our work with multiple qualitative metrics and compare it with other state-of-the-art GAN models.
    \item We conduct case studies to illustrate the usefulness of GAN-generated data. 
\end{itemize}

The rest of the paper is organized as follows. In section~\ref{sec:realted_work}, we introduce the most relevant research of our work. In section~\ref{sec:methodology}, we describe our proposed methodology in detail. Section~\ref{sec:experiments} presents qualitative and quantitative experiments to demonstrate the efficacy of our proposed methods. We discuss the pros and cons of our models and propose future work in section~\ref{sec:futurework}. 
Section~\ref{sec:conclusion} concludes our work.

\section{Background}
\label{sec:realted_work}
\subsection{Generative Adversarial Networks (GANs)}
GANs consist of two models, a generator and a discriminator. These two models are typically implemented by neural networks, but they can be implemented with any form of differentiable system that maps data from one space to the other. The generator tries to capture the distribution of true examples for new data example generation. The discriminator is usually a binary classifier, discriminating generated examples from the true examples as accurately as possible. The optimization of GANs is a minimax optimization problem, in which the goal is to reach Nash equilibrium~\cite{ratliff2013characterization} of the generator and discriminator. Then, the generator can be thought to have captured the real distribution of true examples. 

GANs have had many applications in different areas, but mostly in computer vision and natural language processing. For example, it can generate examples for image datasets~\cite{goodfellow2020generative}, transfer side-view faces to front-view faces~\cite{huang2017beyond}, translate texts to images~\cite{zhang2017stackgan}, etc. While these successes have drawn much attention, GANs have diversified across applications such as time-series data generation. The work~\cite{brophy2021generative} gives a thorough summary of the GAN implementations in this field. The applicability of GANs to this type of data can solve many issues that current dataset holders face. For example, GANs can augment smaller datasets by generating new, previously unseen data. GANs can replace the artifacts with information representative of clean data. And it can be used to denoise signals. GANs can also ensure an extra layer of data protection by generating deferentially private datasets containing no risk of private information linkage from source to generated datasets.

\subsection{Conditional Generative Adversarial Networks (CGANs)}
For a regular GAN model, neither the generator nor the discriminator has control of the data being generated. It never knows what kinds of data are generated, the quality of the generated data, when the training process should be stopped, etc. However, by adding additional information to the generator and discriminator, it is possible for the two networks to gain some power to direct the data generation process. Such information can be class labels, data from different modalities, etc. Shortly after the famous GAN model paper~\cite{goodfellow2014generative}, Mirza et al. proposed a conditional GAN model training strategy~\cite{mirza2014conditional}, which can use data labels during the training process to generate data belonging to specific categories. Recent work in the field of computer vision, styleGAN~\cite{karras2019style}, cycleGAN~\cite{zhu2017unpaired}, concatenate a different format of the image to the real input data, in order to create a new style of synthetic data. 

We assert that conditional GAN can be used to overcome some of the problems encountered in GAN-generated time-series data. For example, a sensor collected time-series dataset contains multiple categories of signals. With a regular GAN, to get a certain category of synthetic signal, we need to train a GAN model on only this type of real signal. Therefore, we need to train multiple GAN models to get every kind of signals for a multi-category dataset. Besides, from our prior work in TTS-GAN~\cite{li2022tts}, we notice that the less data used for training the GAN model, the easier for the discriminator to overfit the training data and the lower is the quality of generated data. Moreover, the available datasets are often small and have imbalanced numbers of samples for each category or class. So it is neither efficient nor effective to train multiple GANs for different categories of data from a single small original dataset. 
However, currently, no work has been done on using labels as control conditions for GAN models to generate multi-category time-series data. 

\begin{figure}
\centering
\subfloat[]
{\includegraphics[width=0.47\columnwidth,keepaspectratio]{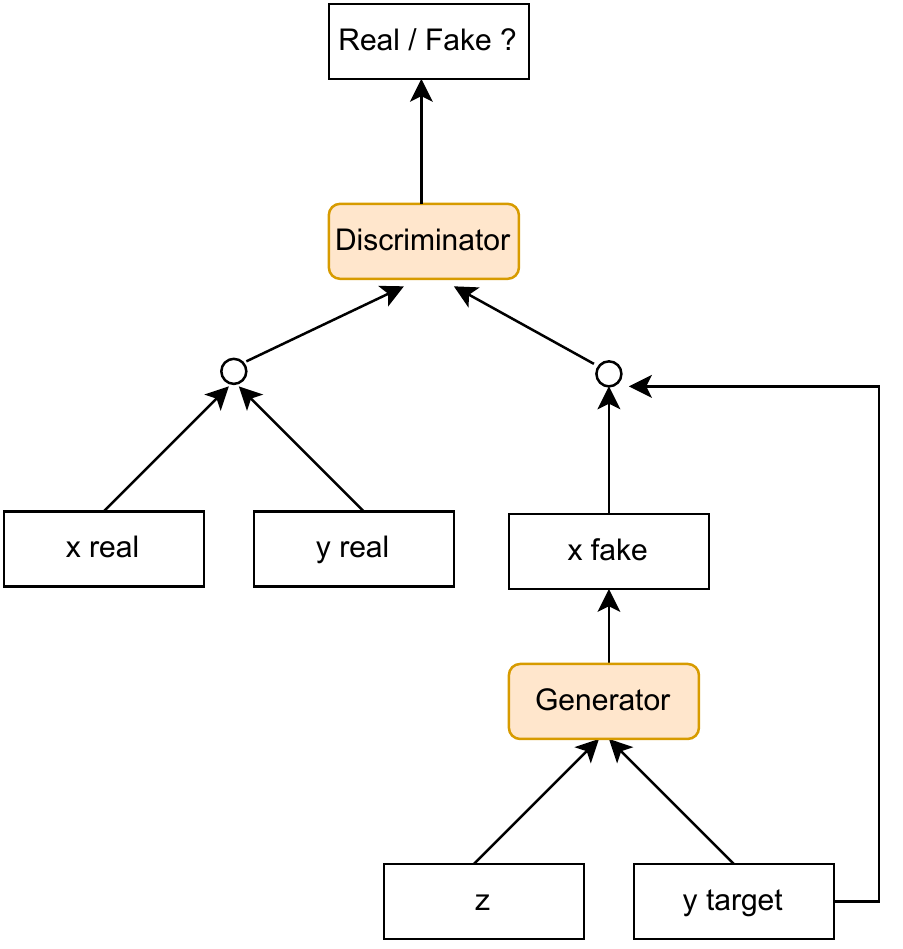}
\label{fig:conventionalCGAN}}
\subfloat[]
{\includegraphics[width=0.42\columnwidth,keepaspectratio]{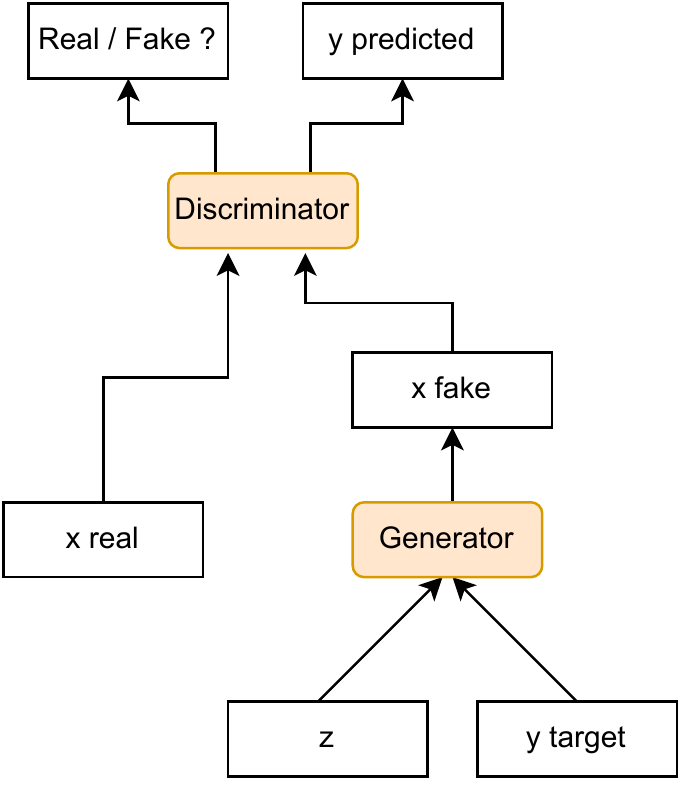}
\label{fig:ourCGAN}}
\caption{Conditional GAN label embedding strategies. (a) Conventional design, (b) Ours. In both structures, z represents random noise inputs; y represents signal labels; x represents signal data.}
\label{fig:Conditional GAN}
\end{figure}

Conventional conditional GAN design for generating different categories of data will provide label information for both generator and discriminator, shown in Fig.~\ref{fig:conventionalCGAN}. We conduct an empirical study on how to embed labels on GAN models in section~\ref{subsec:labelembedding}. We find that the smaller the distortion to the input data, the better is the generative results. Moreover, the transformer-based GAN model seems to be more sensitive to such distortions than other deep learning architectures, such as Convolutional Neural Networks (CNNs). In our experiments, we find that if we follow the common practice of simply embedding labels on both the generator and discriminator, the transformer GAN can not generate any meaningful synthetic data. Inspired by the work in~\cite{choi2018stargan}, we propose a new strategy to embed label information into GAN models. That is to only embed labels on the generator and add another classification head on the discriminator. Let the generator generate multi-category synthetic data and the discriminator will be trained to distinguish the real and synthetic signals as well as their corresponding categories. Fig.~\ref{fig:ourCGAN} shows such an idea. We test this idea on many kinds of GAN model architectures and all of them perform similar to or better than the conventional method. Therefore, we adopt this label embedding idea to our TTS-CGAN model architecture.   


\subsection{Transformer}
The transformer is a state-of-the-art neural network architecture. It only uses attention mechanisms without any Recurrent Neural Networks (RNNs) but improves the results in many NLP tasks.~\cite{vaswani2017attention,devlin2018bert}. 

Given its strong representation capabilities, researchers have also applied transformers to computer vision tasks. In a variety of visual benchmarks, transformer models perform similar to or better than other types of networks, such as CNNs and RNNs. The work in~\cite{dosovitskiy2020image} builds a model named ViT, which applies a pure transformer directly to sequences of image patches. The work in~\cite{jiang2021transgan} builds a pure transformer GAN model to generate synthetic images, where the discriminator's architecture 
is derived from the ViT model. The multi-dimensional time-series data we are dealing with have similarities to both texts and images, meaning a sequence contains both temporal and spatial information. Each timestep in a sequence is like a pixel on an image. The whole sequence can contain an event or multiple events, 
which is similar to a sentence in NLP tasks. 

In this work, we adapt the ideas used in~\cite{dosovitskiy2020image} and~\cite{jiang2021transgan} for images, and view a time-series sequence as a $C \times H \times W $ tuple, where \emph{C} is the number of channels of the time-series data, \emph{H} corresponds to the height of the image, but for time-series that value is set to 1, and \emph{W} corresponds to the width of the image, which for times-series is the number of timesteps in the sequence. 
We divide the tuple into multiple patches on the \emph{W} axis and provide positional encoding to each patch. 

\subsection{Time-series Similarity Analysis}
When judging the quality of GAN-generated images, researchers often post questionnaires to subjects to collect their reactions to the generated images. However, unlike images that are easy to be interpreted with naked eyes, people often find it difficult to interpret the time series by looking at raw signals. 
Dimensionality reduction techniques, such as PCA, and t-SNE, are often used as qualitative methods to distinguish the major feature distributions of time-series datasets. However, we find that even though most of the state-of-the-art GAN models could generate synthetic data with similar PCA and t-SNE distributions their synthetic raw data patterns are not very visually similar. Therefore, there is a need for a more robust technique to quantitatively evaluate the quality of the generated signals.

There exist several time-series similarity metrics, such as point-to-point distance measure, dynamic time warping, longest common sub-sequence, etc~\cite{cassisi2012similarity}. However, absolute similarity (or distance) between real and synthetic signals is not what we are aiming for. Just as there are many kinds of cat images labeled as a cat, a set of time-series data representing the same event could also have different signal patterns. We need a metric that can tell how a set of synthetic signals is similar to a set of real signals. 

The work in ~\cite{yoon2019time} proposed two quantitative metrics, discriminate score and predictive score. Where the discriminate score is computed from training a time-series classification model to distinguish between sequences from the original and generated datasets. And the predictive score is computed from training a sequence-prediction model to predict next-step temporal vectors over each input sequence. Both of the models have an architecture of two LSTM layers. These two scores are restricted by the performance of the LSTM model generalization ability on time-series data. They will fail to evaluate the quality of synthetic signals when such signals are too complicated for the LSTM model to classify or forecast. Our previous work~\cite{li2022tts} proposed two metrics, average cosine similarity, and average Jensen-Shannon distance. These two metrics both require manual feature extraction from the raw sequences for comparison. 

In this paper, we modify the wavelet coherence~\cite{grinsted2004application} metric to be able to measure the similarity of two sets of signals without the need to extract features. The metric is described in detail in sections~\ref{sec:wavelet_coherence} and \ref{subsec:wcoh}. Our experimental results show that our TTS-CGAN-generated data get a much higher wavelet coherence score than other baseline models. And when using this metric to rank these state-of-the-art GAN models, the results are consistent with previous works~\cite{yoon2019time, ni2020conditional}.

\section{Methodology}
\label{sec:methodology}

\subsection{Transformer Time-Series GAN Model Architecture}

\begin{figure}
\centering
\includegraphics[width=.9\columnwidth]{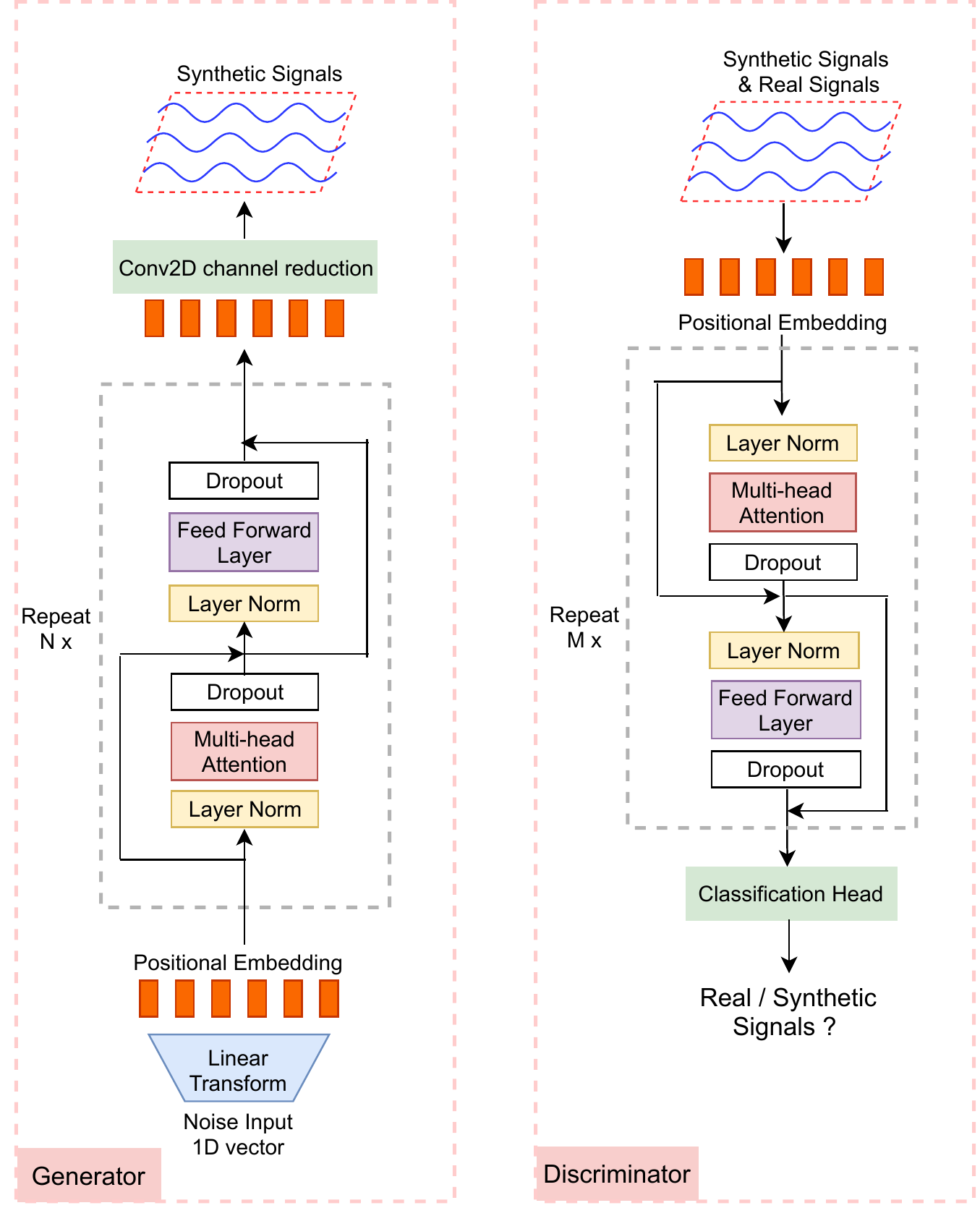}
\caption{TTS-GAN model architecture}
\label{fig:GAN}
\end{figure}

The backbone of our TTS-CGAN architecture is based on our previous TTS-GAN architecture~\cite{li2022tts}. For that reason, we first explain the TTS-GAN and then introduce the modifications necessary to convert it to a conditional GAN. The TTS-GAN model architecture is shown in Fig.~\ref{fig:GAN}. It contains two main components, a generator, and a discriminator. Both of them are built based on the transformer encoder architecture~\cite{vaswani2017attention}. An encoder is a composition of two compound blocks. A multi-head self-attention module constructs the first block and the second block is a feed-forward multi-layer perceptron (MLP) with a GELU activation function. A normalization layer is applied before both of the two blocks and the dropout layer is added after each block. Both blocks employ residual connections. 

The generator first takes in a 1D vector with N uniformly distributed random numbers within the range (0,1), i.e. $N_i \sim U(0,1)$. $N$ represents the latent dimension of the synthetic signals, which is a hyperparameter that can be tuned. The vector is then mapped to a sequence with the same length as the real signals and $M$ hidden dimensions. $M$ is also a hyperparameter that can be changed and is not necessarily equal to the real signal dimensions. Next, the sequence is divided into multiple patches, and a positional encoding value is added to each patch. Those patches are then input to the transformer encoder blocks. Then the encoder blocks outputs are passed through a Conv2D layer to reduce the synthetic data dimensions from $M$ to the real signal dimensions. For example, a synthetic data sequence after the generator transformer encoder layers with a data shape $(hidden\_dimensions, 1, timesteps)$ will be mapped to $(real\_data\_dimensions, 1, timesteps)$. In this way, a random noise vector is transformed into a sequence with the same shape as the real signals. 

The discriminator architecture is similar to the ViT model~\cite{dosovitskiy2020image}, which is a binary classifier to distinguish whether the input sequence is a real signal or a synthetic one. In the ViT model, an image is divided evenly into multiple patches with the same width and height. However, in TTS-GAN, we view any input sequence like an image with a height of 1. The timesteps of the sequence correspond to the image's width. The number of signal channels are treated like the image color channels (RGB). Therefore, to add positional encoding on time series inputs, we only need to divide the width evenly into multiple patches and keep the height of each patch unchanged. This process is explained in detail in section~\ref{sec:processingdata}.

\subsection{Processing Time-Series Data}
\label{sec:processingdata}
\begin{figure}
\centering
\includegraphics[scale=0.6]{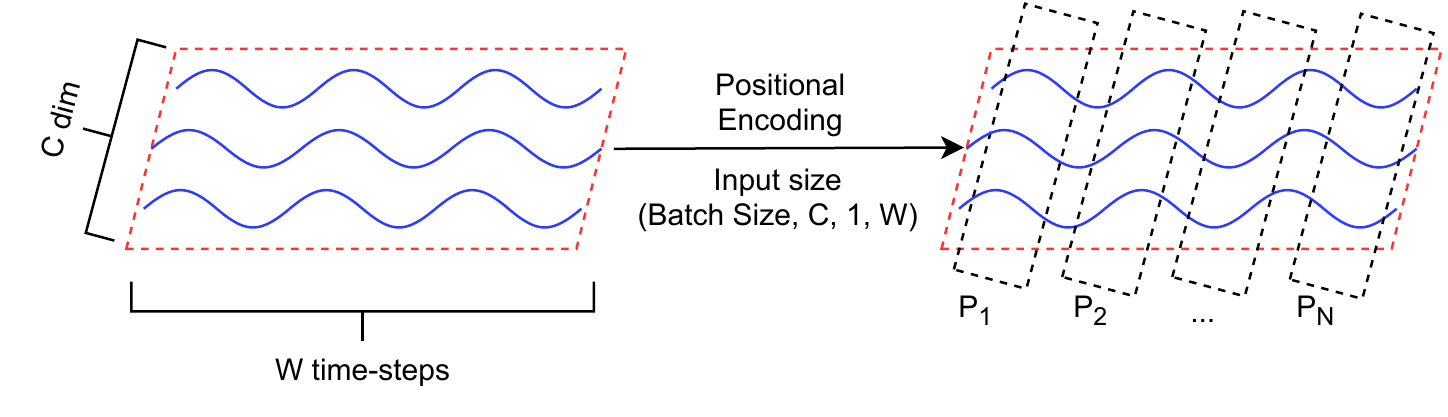}
\caption{Processing time-series data for input to the Transformer. An input sequence of $C$ channels and $W$ time-steps, is split along $W$ into patches patches $P_1, P_2,...,P_N$.}
\label{fig:PositionalEncoding}
\end{figure}

We treat a time-series data sequence like an image with a height equal to 1. The number of timesteps is the width of an image, $W$. A time-series sequence can have a single channel or multiple channels, and those can be viewed as the number of channels (RGB) of an image, $C$. So the input sequences can be represented with the matrix of size $(Batch Size, C, 1, W)$. Then we choose a patch size $N$ to divide a sequence into $W / N$ patches. We then add a soft positional encoding value at the end of each patch. The positional value is learned during model training. Therefore the inputs to the discriminator encoder blocks will have the data shape $(Batch Size, C, 1, (W/N) + 1)$. This process is shown in Fig.~\ref{fig:PositionalEncoding}.

\subsection{Updating TTS-GAN Model Parameters}
\label{sec:loss}
The transformer blocks in the generator and discriminator both use the Mean Squared Error (MSE) loss to update the parameters. The MSE loss measures the average squared difference between the estimated values and the actual values. Here the estimated values are the discriminator outputs and the actual values are the corresponding input labels. To formalize the training loss, we can use $z$ to denote input vectors to the generator. Use $G(z)$ to represent the synthetic data generated by the generator. $D(x)$ is the classification output of the discriminator. $x$ can be the real signals or synthetic signals. $real\_label$ is set to 1 and $fake\_label$ is set to 0. To stabilize the GAN model training, some heuristics can be used when setting label values. For example, we can use soft labels in which $real\_label$ is a float number close to 1 and $fake\_label$ is a float number close to 0. Sometimes, we can also flip the values of the $real\_label$ and the $fake\_label$. In our experiments, whenever a GAN model during the training process can not converge well, we will selectively apply these two strategies to help model training. 
The discriminator loss is the sum of real data loss and fake data (synthetic data) losses. It can be represented as:
\begin{equation} 
L_{D\_real} = MSELoss(D(real), real\_label)
\end{equation}
\begin{equation}
L_{D\_fake} = MSELoss(D(G(z)), fake\_label)
\end{equation}
\begin{equation}
L_{D} = L_{D\_real} + L_{D\_fake}
\end{equation}

The generator wants to generate as real as possible synthetic data. So when updating the generator parameters, we label all synthetic data with real\_label when inputting them to the discriminator. Therefore, the generator loss can be represented as: 
\begin{equation}
L_{G} = MSELoss(D(G(z)), real\_label)
\end{equation}


\subsection{Conditional GAN Model}
\label{subsec:cgan}
One of the limitations of our previous TTS-GAN~\cite{li2022tts} is that it can only be trained on data of one class at a time. That means that for a dataset with $K$ classes, we need to train $K$ different models to generate synthetic data for each class. Training a model on only data from one class limits the amount of data available for training and fails to utilize transfer learning effects among the different classes. This can be a problem for minority classes, when there is not enough data available to train a robust model.

\begin{figure}
\centering
\includegraphics[width=.9\columnwidth]{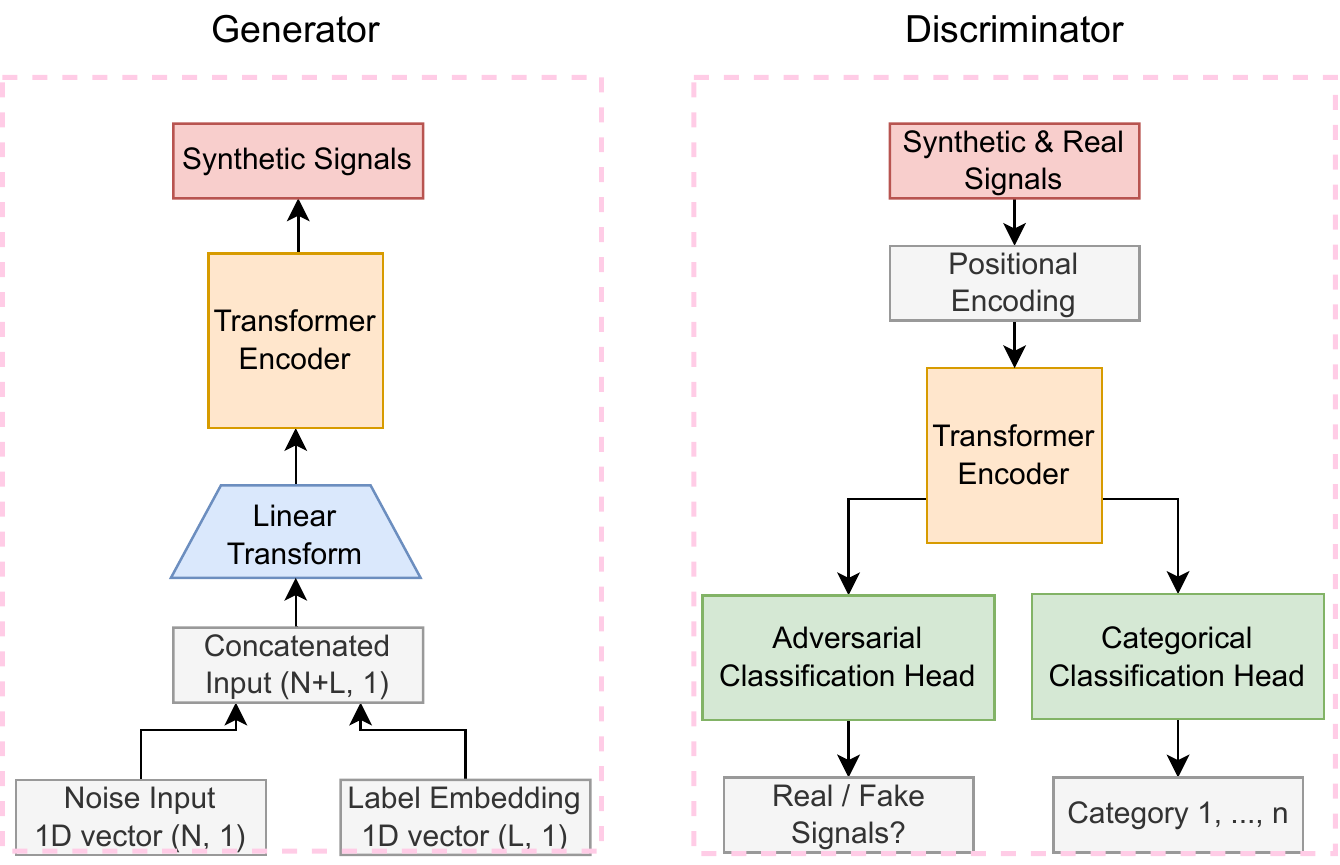}
\caption{Conditional TTS-GAN (TTS-CGAN).}
\label{fig:Conditional TTS-GAN}
\end{figure}

We made several changes to convert the TTS-GAN (Fig.~\ref{fig:GAN}) model to TTS-CGAN (Fig.~\ref{fig:Conditional TTS-GAN}). 
Our goal for the conditional GAN generator $G$ is that it can generate synthetic signals in multiple categories. To achieve this, we train $G$ to generate a synthetic signal from a random vector in latent dimension $z$ and a target categorical label $c, G\left ( z, c \right ) \rightarrow x_{syn}$. The label $c$ is randomly generated so the $G$ learns to flexibly generate different categories of signals. The conditional discriminator $D$ contains two classification heads that will predict whether the input signal is real or fake, $D_{adv}$, and what category this signal belongs to, $D_{cls}$. The discriminator outputs can be written as $D: x\rightarrow \left \{ D_{adv}(x), D_{cls}(x) \right \}$. 

\subsection{Updating TTS-CGAN Model Parameters}
\label{subsec:CGANLoss}
The objective functions to optimize $G$ and $D$ can be written as:
\begin{equation}
\label{equ:L_D}
    L_{D} = -L_{adv} + \lambda L_{cls}^{r},
\end{equation}
\begin{equation}
\label{equ:L_G}
    L_{G} = L_{adv} + \lambda L_{cls}^{f},
\end{equation}

where $L_{adv}$ is the adversarial loss to determine how well $D$ can distinguish real and fake signals. $L_{cls}$ is the classification loss to determine how well $D$ can classify the input signal to its proper categorical label.
$\lambda$ is a hyper-parameter that controls the relative importance of categorical classification loss and adversarial loss. We use $\lambda = 1$ in all of our experiments. 

\subsubsection{Adversarial loss}
To make the generated synthetic signals indistinguishable from real signals, we adopt an adversarial loss
\begin{equation}
\label{equ:advloss}
    L_{adv} = E_{x}[log D_{adv}(x)] + E_{z,c}[log(1-D_{adv}(G(z,c)))]
\end{equation}
where $G$ generates a signal $G(z,c)$ conditioned on both the random noise input $z$ and the target categorical label $c$, while $D$ tries to distinguish between real and fake signals. During the GAN model training, the generator $G$ tries to minimize the adversarial loss, while the discriminator $D$ tries to maximize it. Therefore, in the equation~\ref{equ:L_D}, we add a negative symbol in front of the $L_{adv}$. By minimizing $-L_{adv}$, the discriminator maximizes the $L_{adv}$ loss. 

\subsubsection
{Categorical loss}
For a random noise vector $z$ in latent dimension and a target categorical label $c$, our goal is to map $z$ into an output signal, which is properly classified to the target category $c$. To achieve this goal, we add another classification head on the discriminator $D$ and impose the categorical classification loss when optimizing both $D$ and $G$. We decompose this objective into two terms, a categorical classification loss of real signals used to optimize $D$, it can be defined as:
\begin{equation}
    L_{cls}^{r} = E_{x,c{}'}[-logD_{cls}(c{}'|x)]
\end{equation}
where $L_{cls}^{r}$ is the categorical classification loss on real signals. $c{}'$ is a real signal categorical label. By minimizing this loss function, $D$ learns to classify a real signal $x$ to its corresponding original category $c{}'$. 

Another term, categorical classification loss of fake signals, is used to optimize $G$, which can be defined as:
\begin{equation}
    L_{cls}^{f} = E_{z,c}[-logD_{cls}(c|G(z,c))]
\end{equation}
where $L_{cls}^{f}$ is the categorical classification loss on synthetic(fake) signals. $c$ is a target signal categorical label randomly generated. By minimizing this loss function, $G$ learns to generate signals that can be classified as the target category $c$. 

\subsubsection{Wasserstein loss}
To stabilize the training process and generate higher quality signals, we replace equation~\ref{equ:advloss} with Wasserstein GAN objective with gradient penalty~\cite{arjovsky2017wasserstein, gulrajani2017improved}, defined as
\begin{equation}
\begin{split}
    L_{adv} = E_{x}[D_{adv}(x)] - E_{z,c}[D_{adv}(G(z,c))] \\ - \lambda_{gp}E_{\hat{x}}[(\left \| \nabla_{\hat{x}}D_{adv}(\hat{x}) \right \|_{2} - 1])^{2}]
\end{split}
\label{equ:wassloss}
\end{equation}
where $\hat{x}$ is sampled uniformly along a straight line between a pair of real and synthetic signals. We use $\lambda_{gp} = 10$ for all experiments.

\subsection{A Study of Conditional GAN Label Embedding Strategies}
\label{subsec:labelembedding}
\begin{figure}
    \centering
    
    \subfloat[Image samples from Real MNIST datasets]{\includegraphics[width=\columnwidth, keepaspectratio]{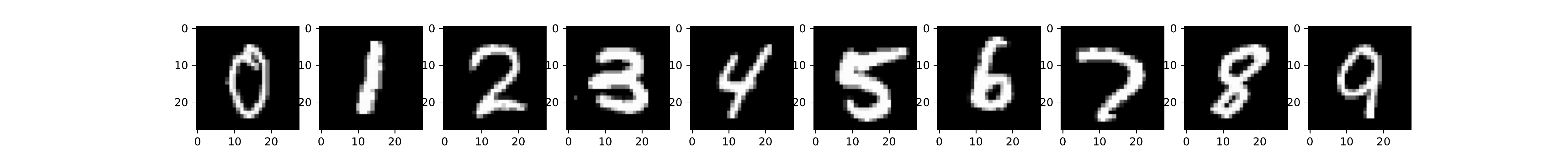}\label{subfig:Real MNIST}}
    
    \subfloat[Strategy 1]{\includegraphics[width=\columnwidth, keepaspectratio]{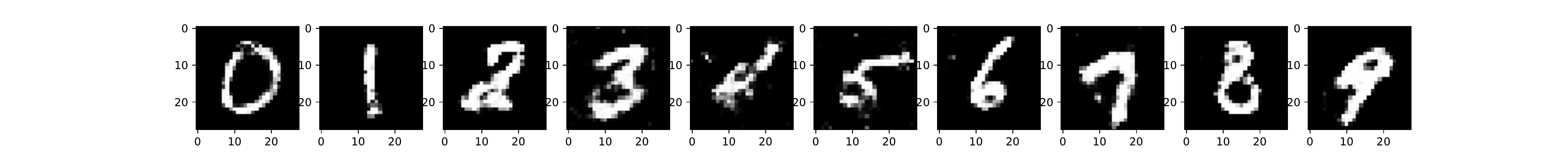}\label{subfig:strategy 1}}
    
    \subfloat[Strategy 2]{\includegraphics[width=\columnwidth, keepaspectratio]{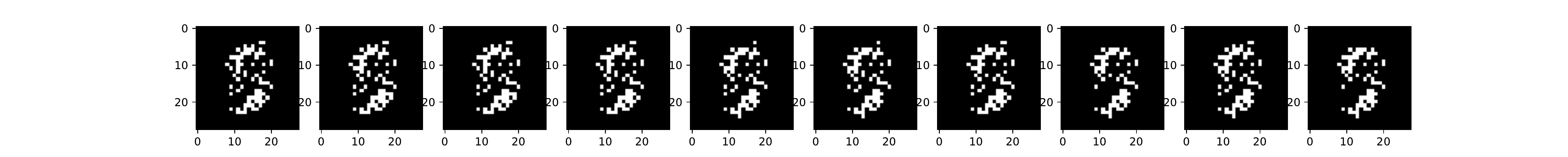}\label{subfig:strategy 2}}
    
    \subfloat[Strategy 3]{\includegraphics[width=\columnwidth, keepaspectratio]{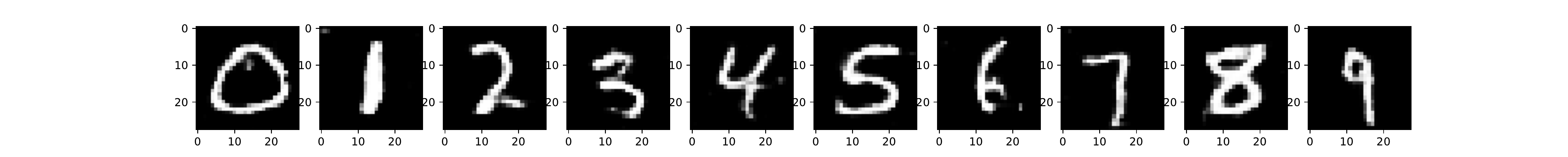}\label{subfig:strategy 3}}
    
    \subfloat[Strategy 4 Our method]{\includegraphics[width=\columnwidth, keepaspectratio]{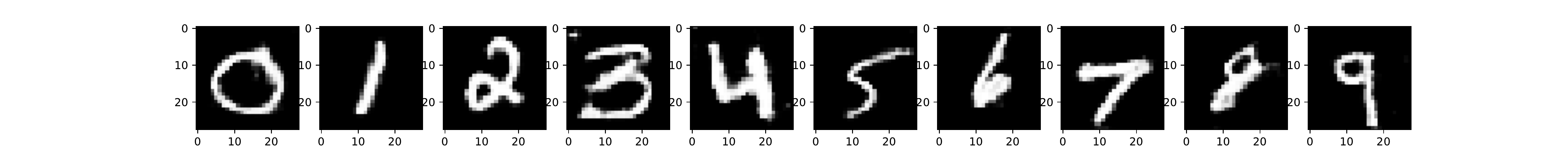}\label{subfig:strategy 4}}
    
    \caption{Conditional Label Embedding strategies.}
    \label{fig:labemb}
\end{figure}

Since synthetic time-series data are hard to interpret, we can not easily know which label embedding strategy will perform best when build a conditional GAN model. Therefore, we choose a simple machine learning dataset, MNIST~\cite{deng2012mnist}, trying to help us derive the best label embedding strategy. To mimic the signal generating process, we flatten the MNIST digit images to 1D vectors. These vectors can be viewed as sequential inputs to the GAN models. 
We test a combination of label embedding methods to generate vectorized digit images. Then we reshape these vectors to the original image size and evaluate their fidelity. We listed a few experimental results of generated digit images with different label embedding strategies in Fig.~\ref{fig:labemb}.  

\textbf{Strategy 1:} Concatenate label embeddings on both Generator and Discriminator inputs on a linear-GAN. In this experiment, we build a GAN model in which both the generator and discriminator only contain a few linear layers. The label embedding strategy is same as the one shown in Fig.~\ref{fig:conventionalCGAN}. In this strategy, the embedded label vectors are concatenated to the end of the input vectors. For example, if the generator input is a 1D vector of size $(100, 1)$ and the embedded label is a 1D vector of size $(10, 1)$. The label embedded input to the generator will be a 1D vector of size $(110, 1)$. The plots in Fig.~\ref{subfig:strategy 1} show that this strategy generated digit images are recognizable but contain a lot of noise.  
    
\textbf{Strategy 2:} Add label embeddings on both Generator and Discriminator inputs on linear-GAN. In this experiment, we use the same linear GAN model as strategy 1. But instead of concatenating label embedding to the input data, we add their values together. For example, if the generator input and embedded label are both 1D vectors of size $(100, 1)$, we add these two vectors together to get another vector also has the size $(100, 1)$. The idea is similar to the strategy commonly used by Transformer models for embedding positional encoding to the input tokens. However, from the images in~\ref{subfig:strategy 2}, we can see such strategy does not work well for label embedding.
    
\textbf{Strategy 3:} Concatenate label embeddings on both Generator and Discriminator inputs on a Convolutional Conditional GAN (CNN-GAN). In this experiment, we build a Conditional GAN with convolution layers to generate categorical digit images. The label embedding information is concatenated to the image feature maps as a separate channel. For example, after the Convolution layer transformation, the input image vector is reshaped to a feature map with the size $(128, 7, 7)$. The label embedding information is also shaped by a linear layer with the size $(1, 7, 7)$. The we concatenate them to a feature map with the size $(129, 7, 7)$. The images in~\ref{subfig:strategy 3} show that this strategy generates acceptable quality of synthetic digit images. 
    
\textbf{Strategy 4 (Our method)} Concatenate label embedding only on the Generator and add a classification head on the Discriminator. We use the same concatenation method and CNN-GAN model architecture as in strategy 3. But instead of adding label embedding information to both the generator and discriminator, we use the method shown in Fig.~\ref{fig:ourCGAN}. The images in~\ref{subfig:strategy 4} show that this strategy produces the best generated image quality over all the other strategies. 

Besides the strategies mentioned above, we also tested many other label embedding combinations and GAN model architectures. We summarize the  experimental findings as follows:
\begin{enumerate}
\item Concatenating label embedding to both the generator and discriminator inputs is a naive way to build a conditional GAN model. It may work but does not perform equally well on different model architectures.
\item Adding label embedding instead of concatenating as the transformer positional encoding does is not a practical solution for label embedding.
\item The minimum distortion rule applies. The label embedding should be concatenated to the input data with as little information as possible. 
\item The more complex the GAN network, the better synthetic data ability it will generate.
\item Concatenating label embedding to the generator and adding a category classification head to the discriminator overall performed best on all kinds of GAN model architectures that we tested.  
\end{enumerate}

\subsection{Wavelet Coherence Similarity Metric}\label{sec:wavelet_coherence}



\begin{figure}
    \centering
    
    \subfloat[NIRS raw signals.]{\includegraphics[width=.445\columnwidth, keepaspectratio]{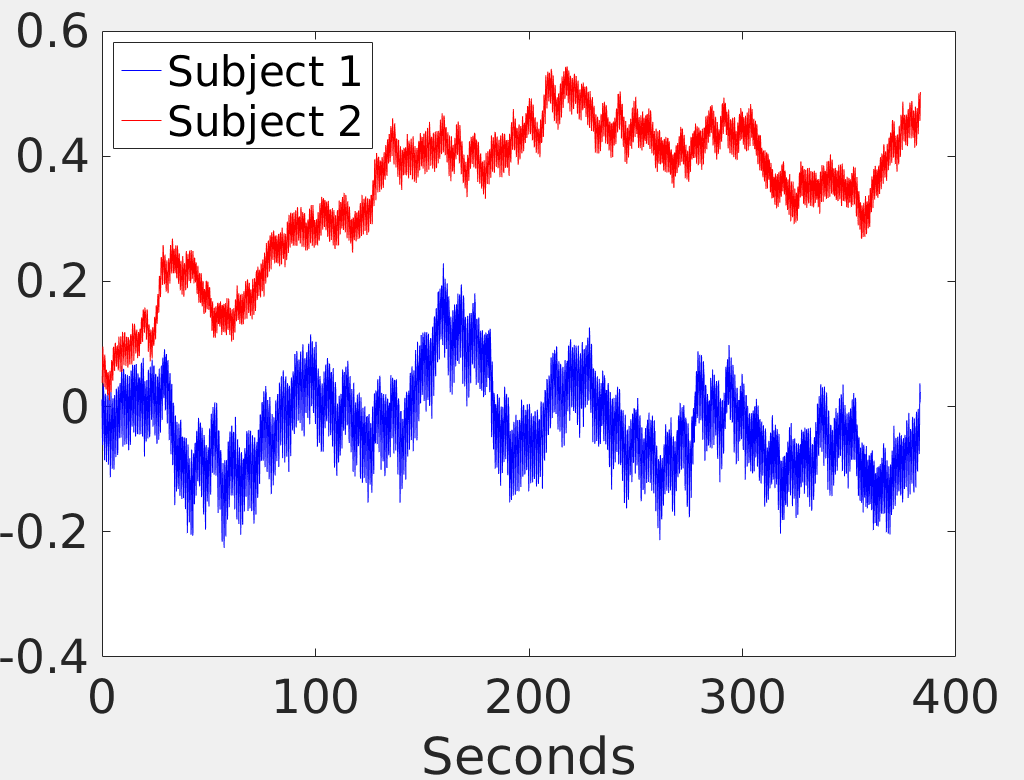}\label{fig:NIRSRaw}}
    \hfill
    \subfloat[NIRS Wavelet Coherence plot.]{\includegraphics[width=.49\columnwidth, keepaspectratio]{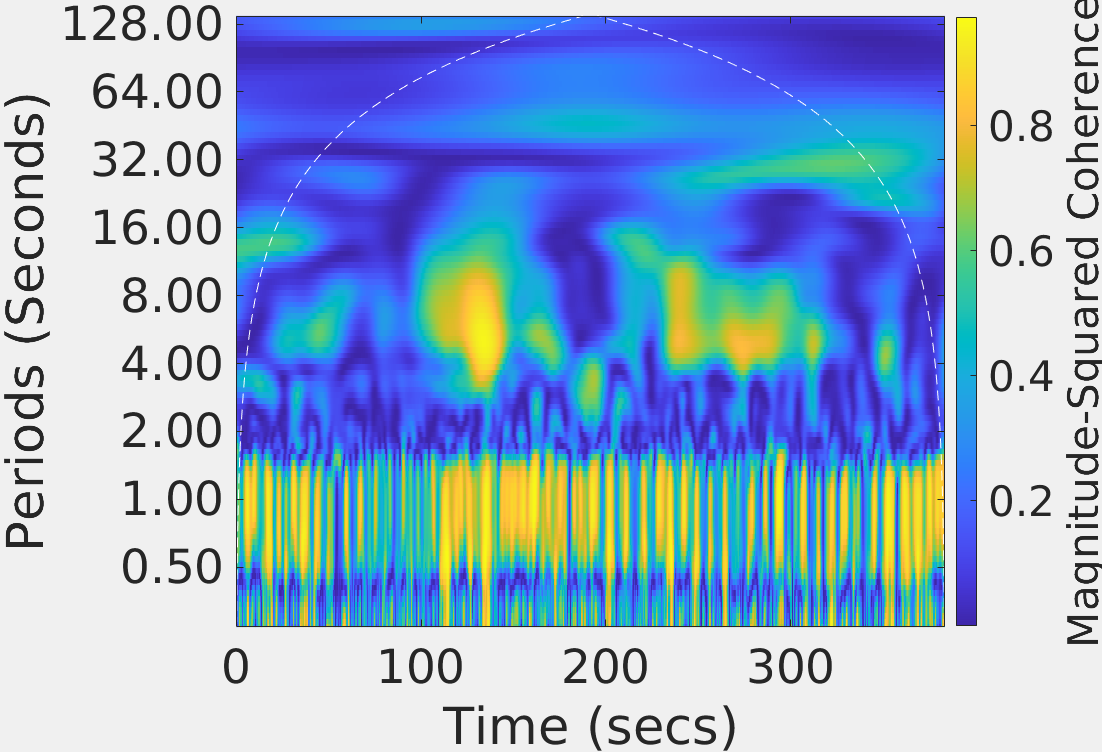}\label{fig:NIRSWcho}}
    
    \caption{Two NIRS signals and their corresponding wavelet coherence plot.}
    \label{fig:wavelet_coh}
\end{figure}

Wavelet coherence~\cite{grinsted2004application} measures the correlation between two signals. The wavelet coherence of two time-series $x$ and $y$ is:
\begin{equation}
    wcoh = \frac{\left | S(C_{x}^{*}(a,b)C_{y}(a,b)) \right |^{2}}{S(\left | C_{x}(a,b) \right |^{2}) \cdot S(\left | C_{y}(a,b) \right |^{2})}
\end{equation}
where $C_{x}(a, b)$ and $C_{y}(a, b)$ denote the continuous wavelet transforms of $x$ and $y$ at scales $a$ and position $b$. The superscript $*$ is the complex conjugate and $S$ is a smoothing operator in time and scale. Wavelet coherence is useful for analyzing non-stationary\footnote{A signal is said to be non-stationary if its frequency of spectral contents is changing with respect to time.} signals. The inputs $x$ and $y$ must be of equal in length, one-dimensional real-valued signals. The coherence is computed using the analytic Morlet wavelet. 

We use an example of finding coherent oscillations in human brain activities to better illustrate the properties of the Wavelet Coherence metric~\cite{wchowebsite}. Fig.~\ref{fig:NIRSRaw} shows two human subjects' brain activities recorded as near-infrared spectroscopy (NIRS) data~\cite{cui2012nirs}. When recording the data, the subjects are cooperating on a task repeated periodically, approximately 7.5 seconds. However, by looking at these two raw signals, we can hardly tell any similarities between them. When using the wavelet coherence metric to compute the correlation between these two signals, we can easily observe some common behaviors that co-exist between these two subjects. Fig.~\ref{fig:NIRSWcho} shows a wavelet coherence plot of the two signals plotted in Fig.~\ref{fig:NIRSRaw}. In Fig.~\ref{fig:NIRSWcho}, the x-axis shows the length of these two signals, which is around 400 seconds. The wavelet coherence computation process will decompose a non-stationary sequence into many stationary sequences, and each stationary sequence has a constant frequency and phase. Therefore, if we look at the y axis of this wavelet coherence plot, we can see that these two signals are decomposed into many sequences with multiple periods\footnote{The period is inversely proportional to frequency.}. We can see that in period 1, these two signals show a high correlation over the whole 400 seconds. It represents the cardiac rhythms of the two subjects, which are commonly similar among human beings. In addition, we can observe a strong correlation at around the period of 7-8 within 100-300 seconds. It represents the two subjects periodically doing the same task every 7.5 seconds when recording such NIRS signals.

\subsection{Wavelet Coherence for twos sets of signals} 
\label{subsec:wcoh}
Since the wavelet coherence of two signals, $wcoh$ is a 2D matrix with the shape $Frequencies \times Timesteps$. We compute the sum over the x-axis and mean over the y-axis of $wcoh$ to get a scalar value of two signals' wavelet coherence $wcoh\_s$ to use as a score. If these two signals are multi-dimensional, we compute a mean $wcoh\_s$ over all dimensions.
For two sets of signals, $A$ and $B$, each set has a total number of $n$ signal samples. We compute $wcoh\_s$ of every signal in $A$ to every signal in $B$ and then compute the overall mean $wcoh\_set$ as the average wavelet coherence for twos sets of signals. This process can be written as:
\begin{equation}
    wcoh\_set = \frac{\sum_{i=1}^{n}({\sum_{j=1}^{n}wcoh\_s(A_{i},B_{j})}/n)}{n}
\end{equation}

\begin{algorithm}
\caption{Compute {wcoh\_set} for two sets of signals}\label{alg:Wcho_set alg}
\textbf{Input}:signal set $A$, signal set $B$\\
\textbf{Output}:$wcoh\_set$ of $A,B$
\begin{algorithmic}
    \State $wcoh\_set = 0$
    \For{signal x: 1..n in $A$}
        \State $wcoh\_set_{x} = 0$
        \For{signal y: 1..n in $B$}
            \State $wcoh = 0$
            \For{signal dimension $c$:1..m in x}
                \State compute $wcoh_{c}$ of $x_{c}, y_{c}$
                \State sum all numbers in $wcoh_{c}$ to a scalar value
                \State $wcoh$ = $wcoh$ + $wcoh_{c}$
            \EndFor
            \State $wcoh$ = $wcoh / m$
            \State $wcoh\_set_{x}$ = $wcoh\_set_{x}$ + $wcoh$
        \EndFor
        \State $wcoh\_set_{x}$ = $wcoh\_set_{x} / n$
        \State $wcoh\_set$ = $wcoh\_set$ + $wcoh\_set_{x}$
    \EndFor
    \State $wcoh\_set$ = $wcoh\_set / n$
    
    \Return $wcoh\_set$
    
\end{algorithmic}
\end{algorithm}

\section{Experiments}
\label{sec:experiments}
\subsection{Datasets}

We evaluate our models on four datasets. Simulated sinusoidal waves, UniMiB human activity recognition (HAR) dataset~\cite{app7101101}, the PTB Diagnostic ECG Database~\cite{bousseljot1995nutzung,goldberger2000physiobank} and MIT-BIH Arrhythmia Database~\cite{moody2001impact, goldberger2000physiobank}.
A few raw data samples for each dataset are shown in Fig.~\ref{fig:real_syn_Signals}.

The \textbf{sinusoidal waves} are simulated with random frequencies $A$ and phases $B$ values between [0, 0.1]. The sequence length is 24 and the number of dimensions is 5. For each dimension $i \in \{1, ..., 5\}$, the sequence can be represented with the formula $x_i(t) = sin(At + B)$, where $A \in (0, 0.1)$ and $B \in (0, 0.1)$. A total number of 10000 simulated sinusoidal waves are used to train the GAN model.

For the \textbf{UniMiB dataset}~\cite{app7101101}, we select two random categories (Jumping and Running) from 24 subjects' recordings to train GAN models. The two classes have 600 and 1572 samples respectively. Every sample has 150 timesteps and three acceleration values (x, y, z) at each timestep. All of the recordings are channel-wise normalized to $\mu=0$ and $\sigma=1$.

The \textbf{PTB Diagnostic ECG dataset}~\cite{bousseljot1995nutzung,goldberger2000physiobank} contains human heartbeat signals in two categories, normal and abnormal with 4046 and 10506 samples respectively. Each sequence represents a heartbeat sampled at 125Hz. The original length of each sequence is 188, padded with zeros at the end to create fixed-length sequences. We only use the timesteps 5 to 55 of each sample, which is the part of the sequence containing the most useful information about a heartbeat.

The \textbf{MIT-BIH Arrhythmia dataset}~\cite{moody2001impact, goldberger2000physiobank} contains 48 half-hour excerpts of two-channel ambulatory ECG recordings, obtained from 47 subjects in 5 different heart health conditions. They are 'Non-Ectopic Beats', 'Superventrical Ectopic', 'Ventricular Beats', 'Unknown', and 'Fusion Beats'. Each sequence is sampled at 125Hz, has a length of 187, and a label in one of five categories. The dataset has a very imbalanced distribution of these five categories. We re-sample the training and testing sets to have a normal distribution in all our experiments.     

\subsection{Experiment Setups}
We conduct all experiments on an Intel server with a 3.40GHz CPU, 377GB RAM, and 2 Nvidia 1080 GPUs. We use the PyTorch deep learning library~\cite{NEURIPS2019_9015} to develop the project. For all datasets, the vector size input to the generator is $(100, 1)$. The transformer blocks in the generator and discriminator are both repeated three times. We adopt a learning rate of $1e-4$ for the generator and $3e-4$ for the discriminator. An Adam optimizer with $\beta_1 = 0.9$ and $\beta_2 = 0.999$, and a batch size of 32 for both generator and discriminator, are used for all experiments. The $\lambda$ in equation~\ref{equ:L_D} and \ref{equ:L_G} is set to 1. The $\lambda_{gp}$ in equation~\ref{equ:wassloss} is set to 10. These parameters are set based on our computation resources capability, practical experience, or GAN model training conventions. A case-by-case parameter tuning may result in better synthetic data quality but our GAN models still perform better than other state-of-the-arts even with such general setups. We save the real data samples and generated synthetic data samples to matrix files and compute the Wavelet Coherence sores from the MATLAB platform.

\subsection{TTS-GAN and TTS-CGAN Evaluation}
We evaluate our proposed models TTS-GAN and TTS-CGAN using several qualitative and quantitative metrics and compare their performance with multiple state-of-the-art time-series GAN models.

\subsubsection{Raw data visualization}

\begin{figure*}
\vspace{-6mm}
\centering

\subfloat{\includegraphics[width=0.4\textwidth, keepaspectratio]{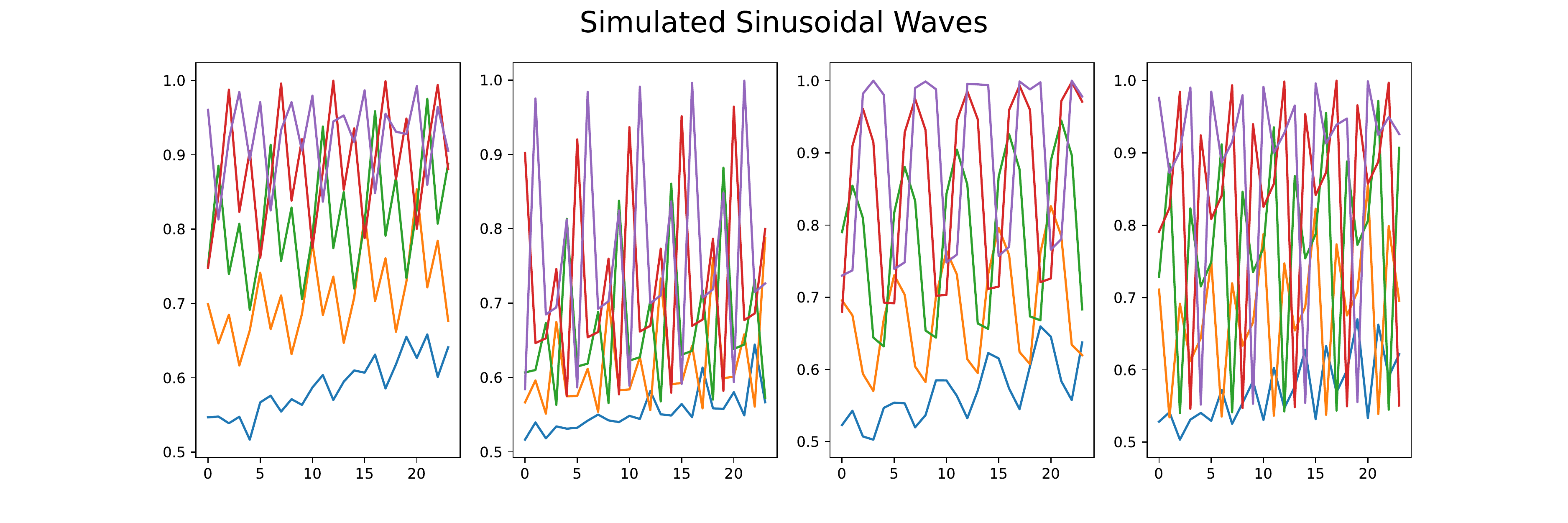}}
\subfloat{\includegraphics[width=0.4\textwidth, keepaspectratio]{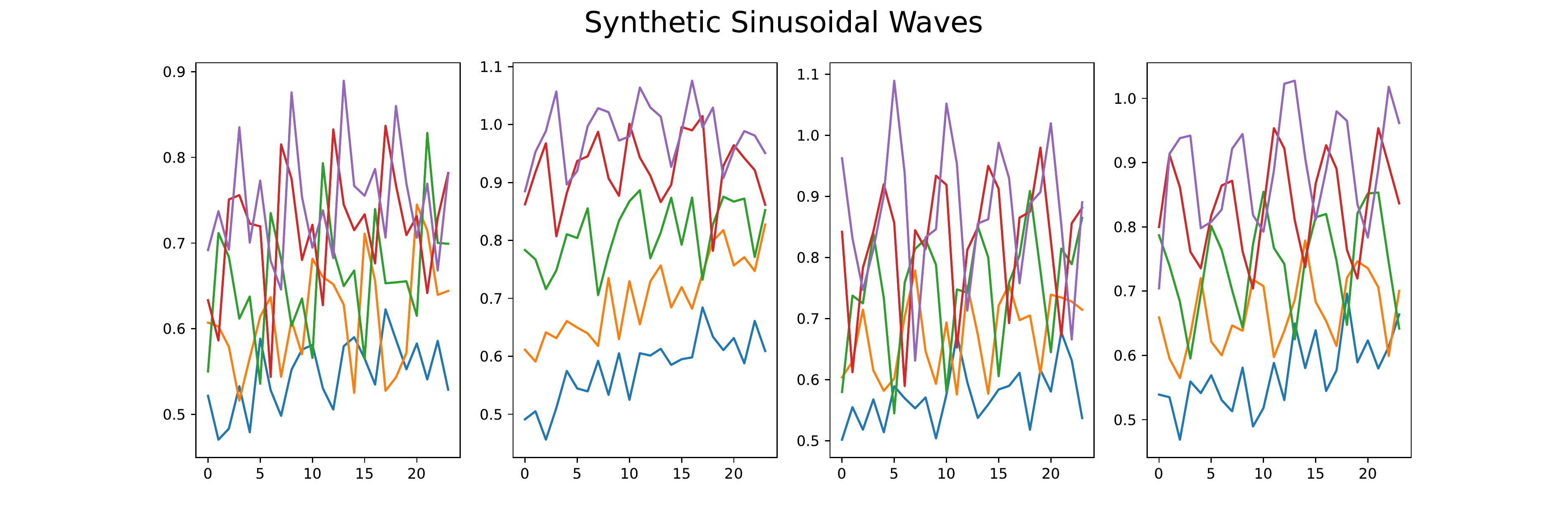}}

\subfloat{\includegraphics[width=0.4\textwidth, keepaspectratio]{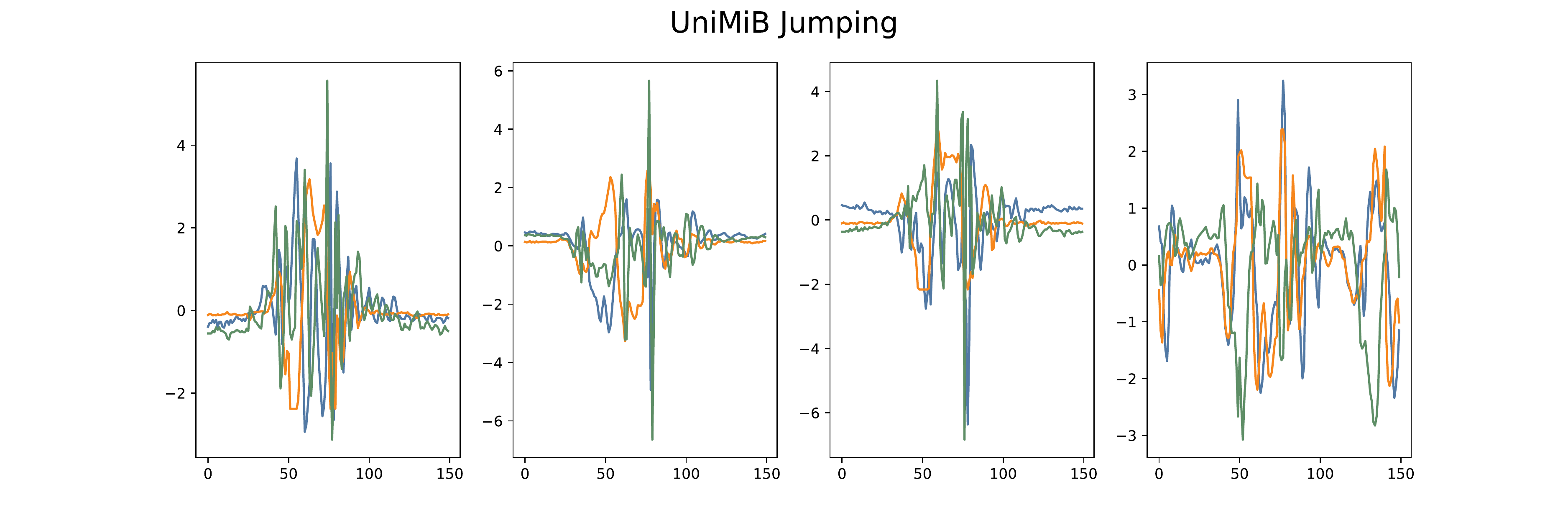}}
\subfloat{\includegraphics[width=0.4\textwidth, keepaspectratio]{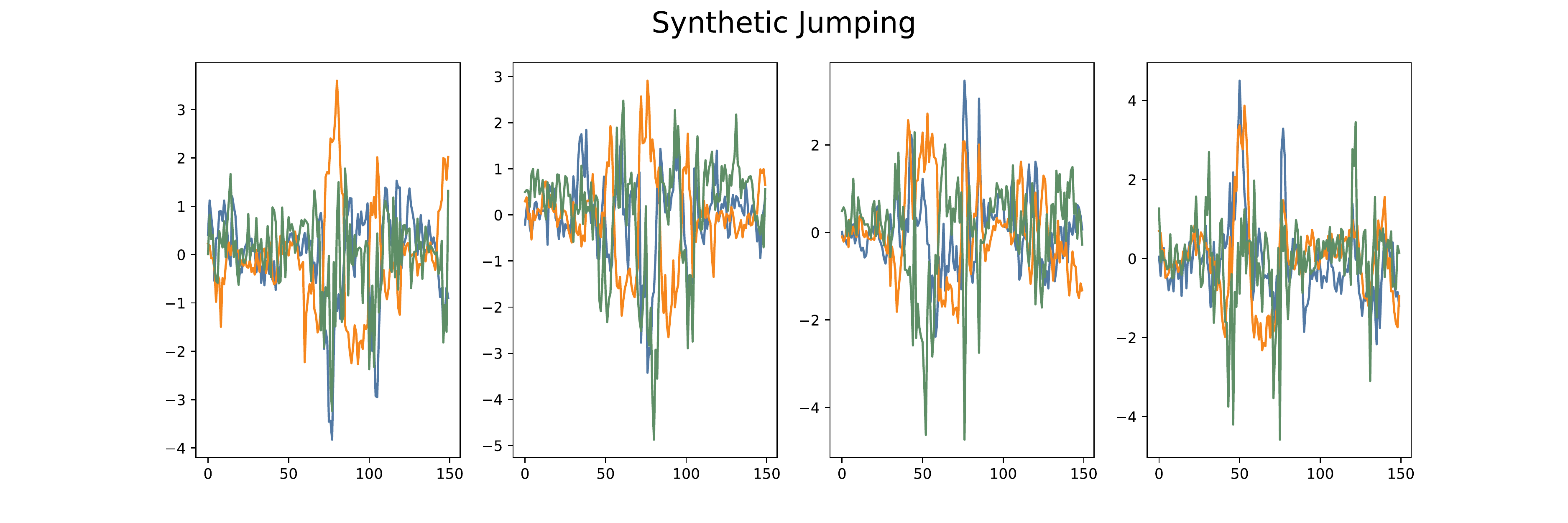}}

\subfloat{\includegraphics[width=0.4\textwidth, keepaspectratio]{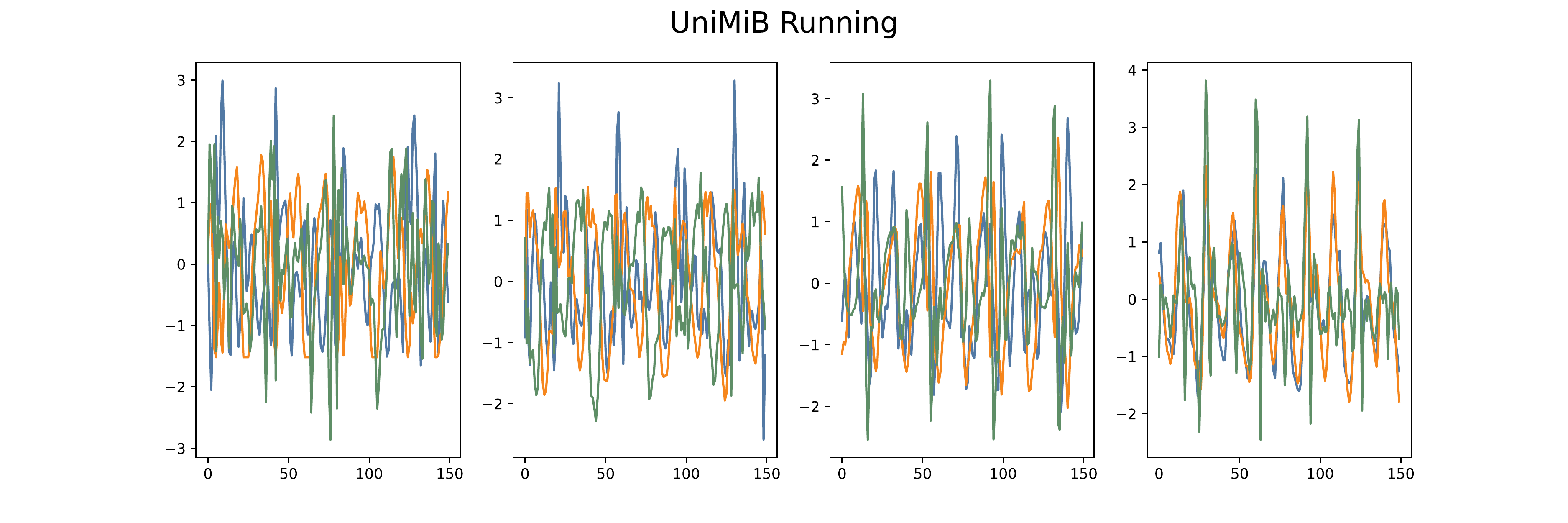}}
\subfloat{\includegraphics[width=0.4\textwidth, keepaspectratio]{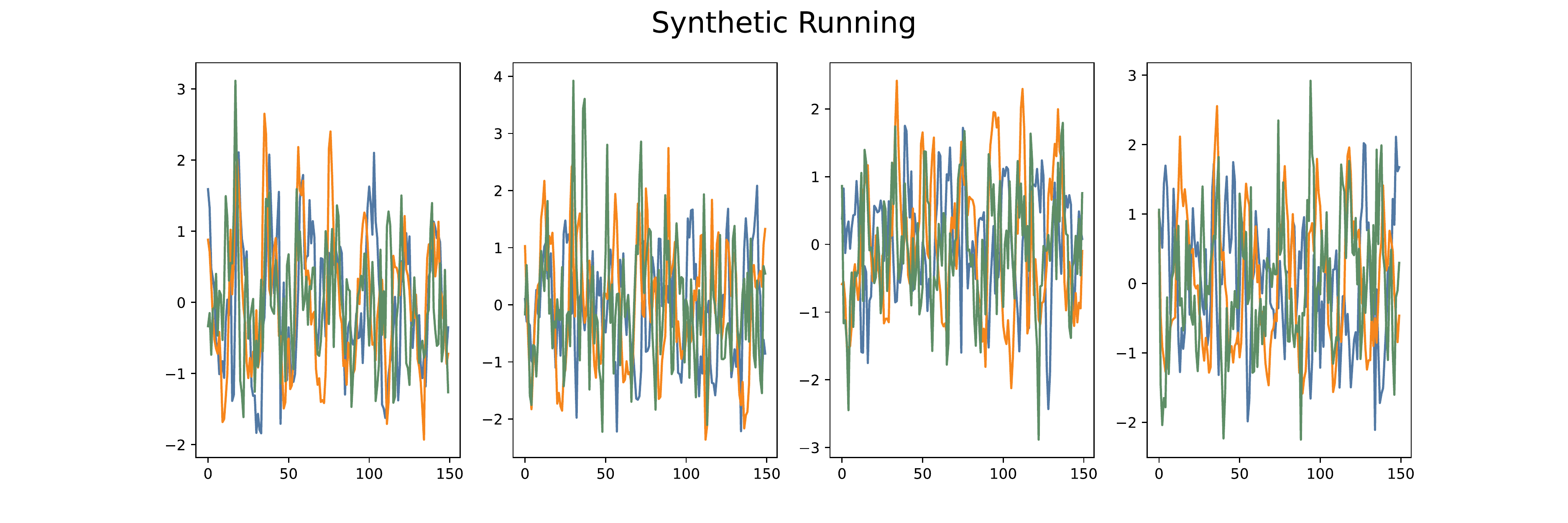}}

\subfloat{\includegraphics[width=0.4\textwidth, keepaspectratio]{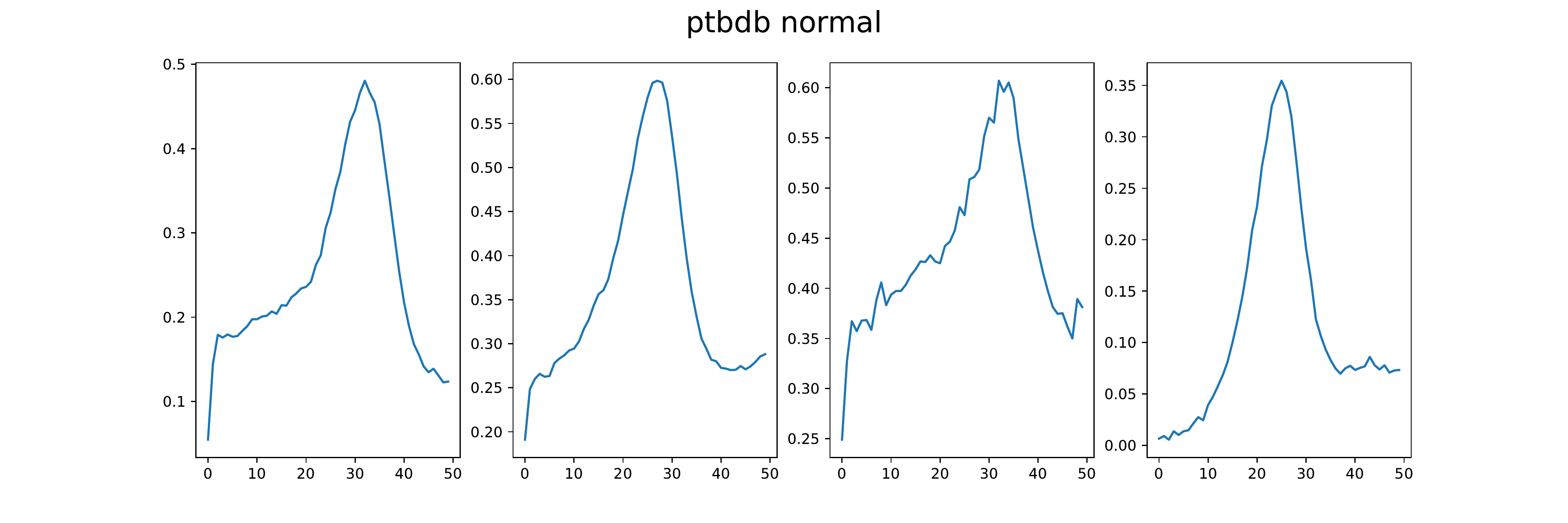}}
\subfloat{\includegraphics[width=0.4\textwidth, keepaspectratio]{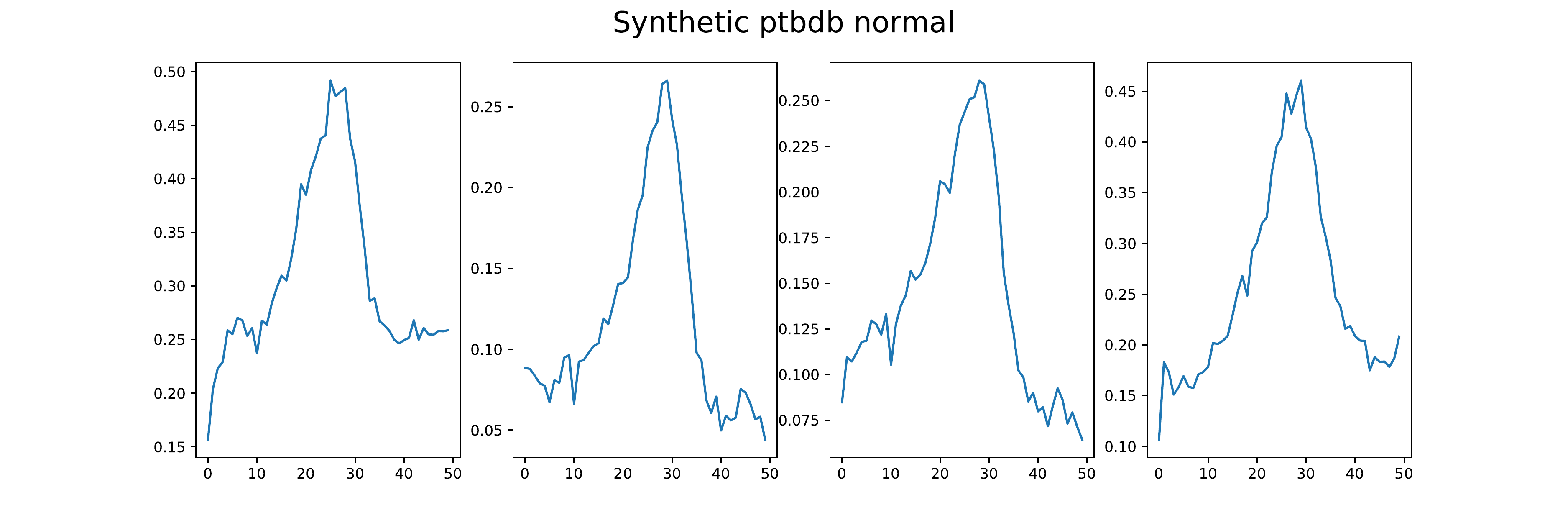}}

\subfloat{\includegraphics[width=0.4\textwidth, keepaspectratio]{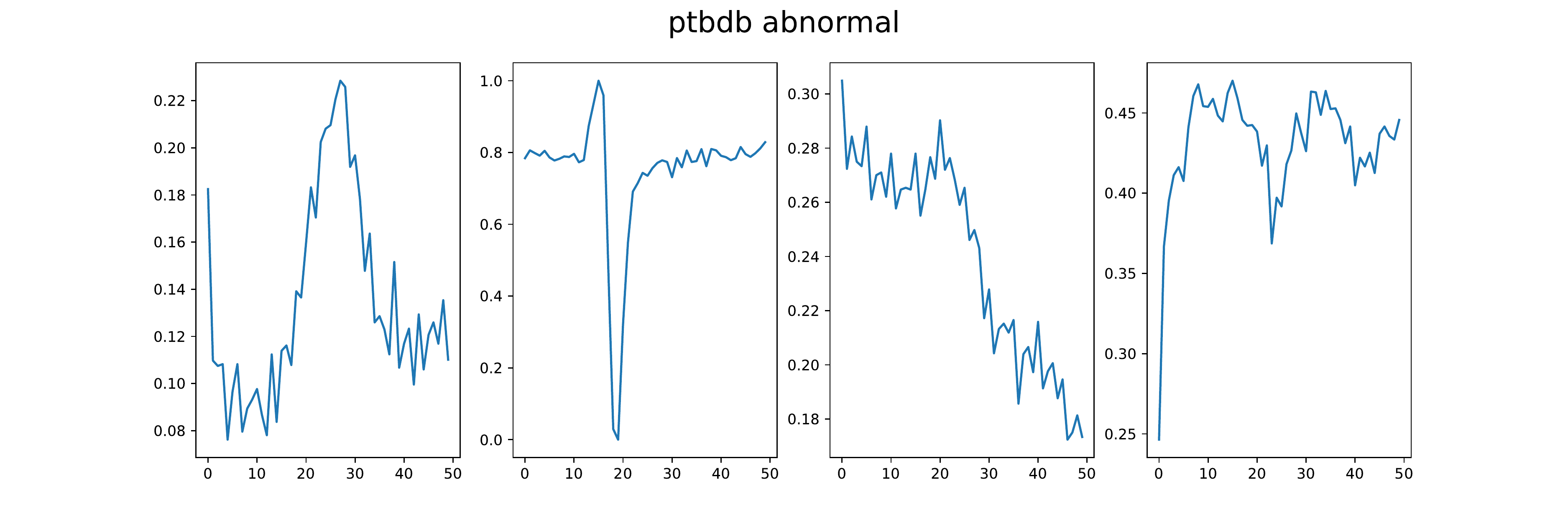}}
\subfloat{\includegraphics[width=0.4\textwidth, keepaspectratio]{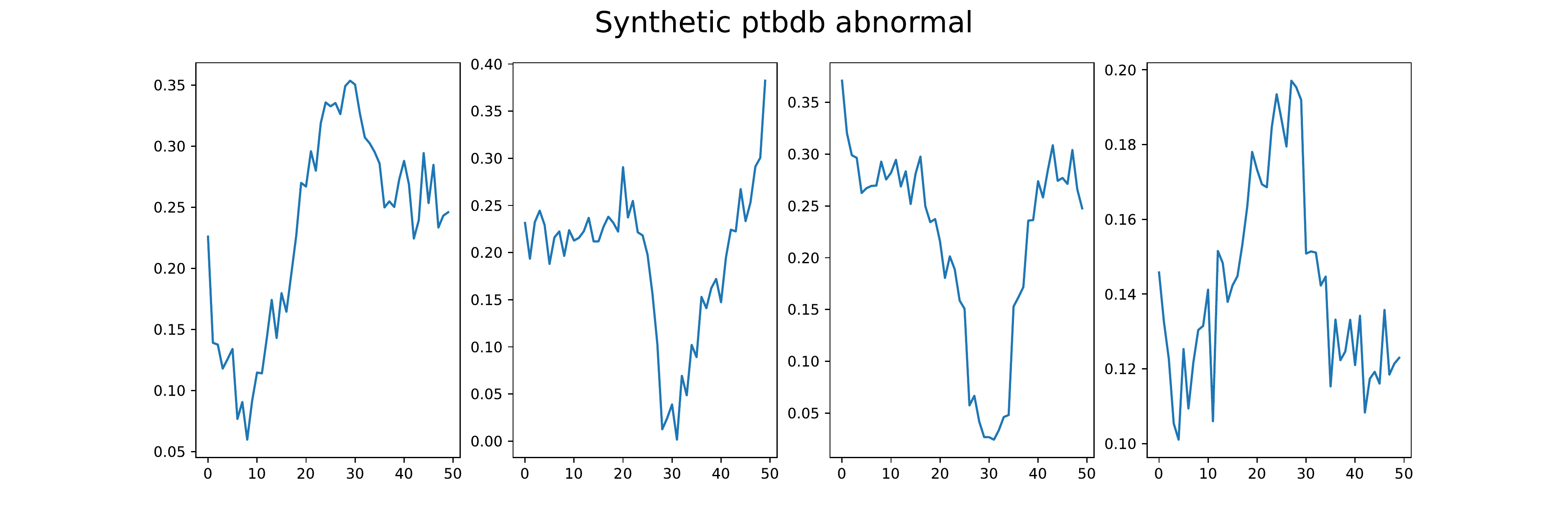}}

\subfloat{\includegraphics[width=0.4\textwidth, keepaspectratio]{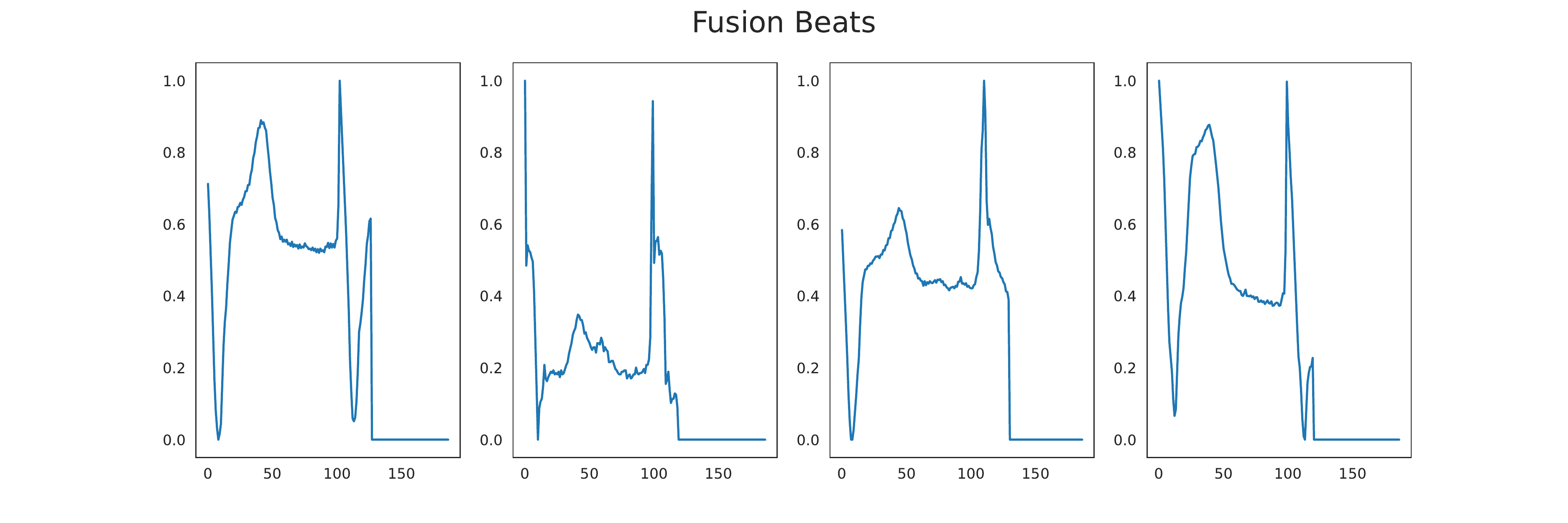}}
\subfloat{\includegraphics[width=0.4\textwidth, keepaspectratio]{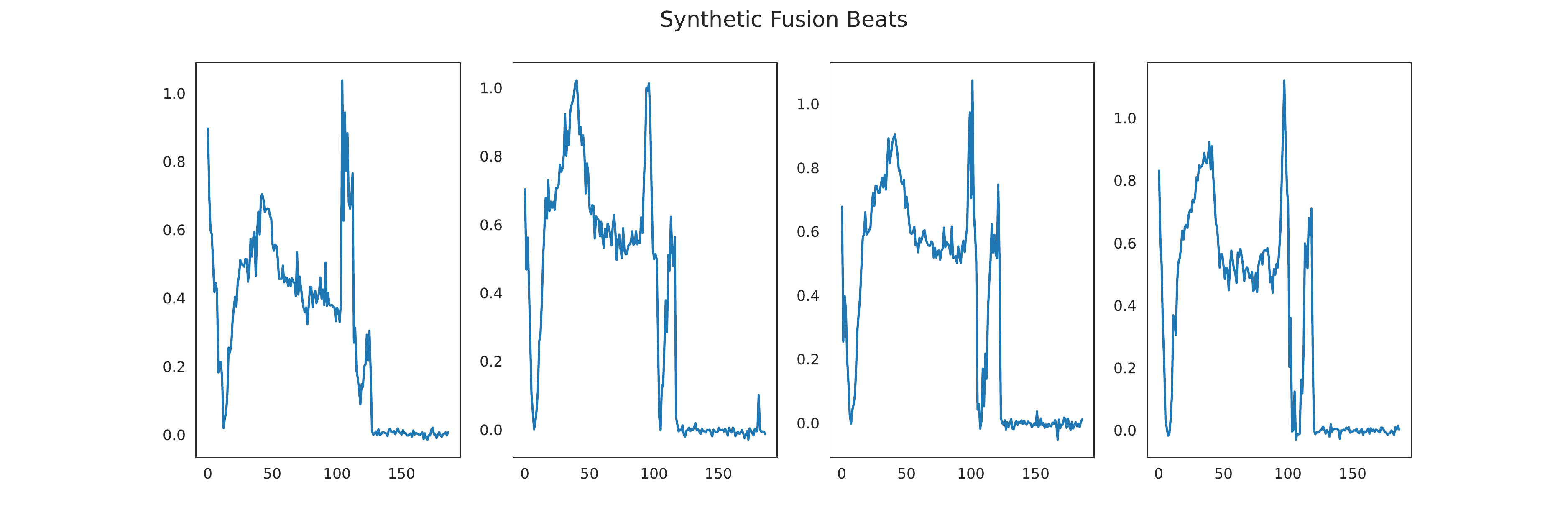}}

\subfloat{\includegraphics[width=0.4\textwidth, keepaspectratio]{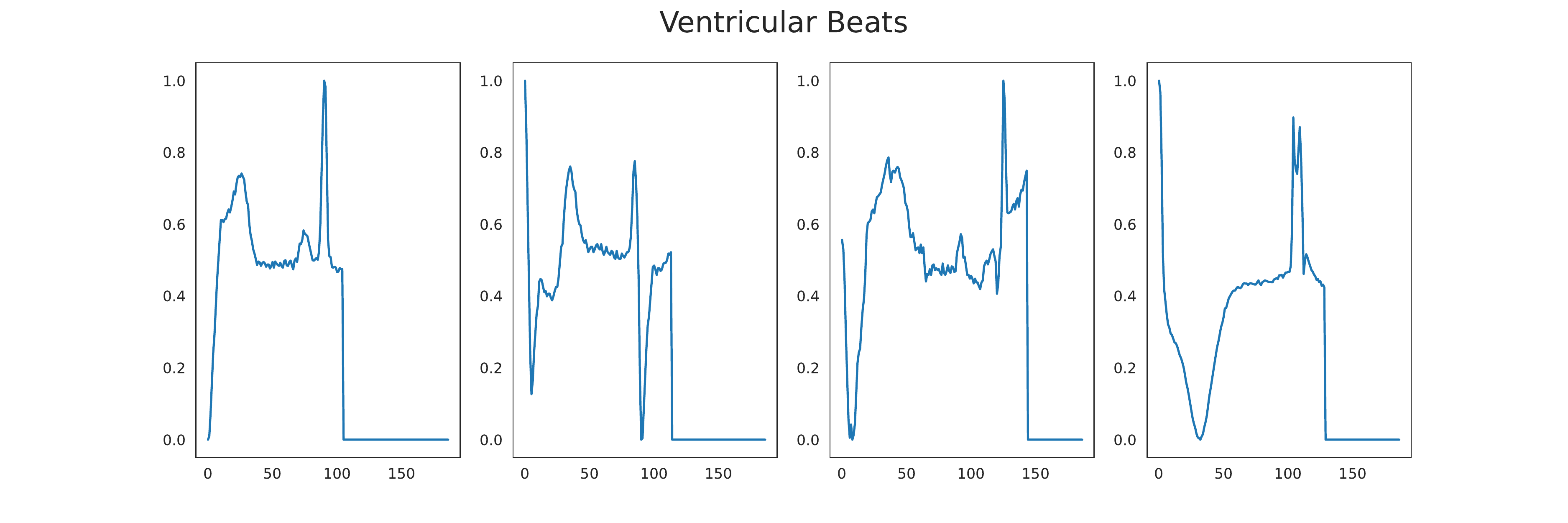}}
\subfloat{\includegraphics[width=0.4\textwidth, keepaspectratio]{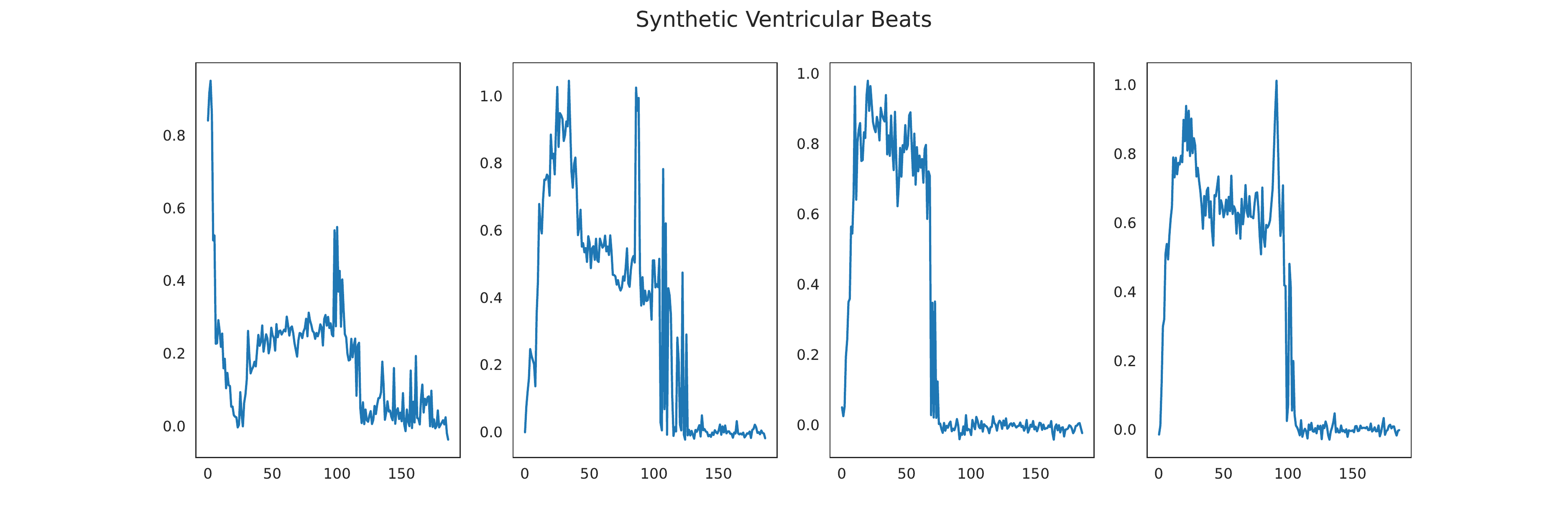}}

\caption{A visual comparison of real data and their corresponding synthetic data generated by TTS-GAN (top 1-5) and by TTS-CGAN (bottom 1-2). The left line plots are real signals and the right line plots are synthetic signals.}
\label{fig:real_syn_Signals}
\vspace{-4mm}
\end{figure*}

The right column of plots in Fig.~\ref{fig:real_syn_Signals} shows some samples of synthetic data generated by TTS-GAN and TTS-CGAN. Their real counterparts are shown in the Fig.~\ref{fig:real_syn_Signals} left column. The top line of plots are the simulated sinusoidal waves and synthetic ones. Such data are stationary simulation signals, where each data sample contains five independent simulated sinusoidal waves with randomly generated frequencies and phases. The GAN model is expected to generate similar stationary data samples by observing the real data samples and learning the data periodically changing rules itself. From the synthetic data plots, we can see that the TTS-GAN model has learned the general rule for generating such stationary data. 

The next two rows of plots are synthetic jumping and running data generated from the UniMiB human activity recognition dataset. These are two examples we selected from the total 9 categories of this dataset. During the experiment, we train a TTS-GAN on each of the 9 categories' real data samples and make comparisons within each category. From the experiment, we find that the real data within a category often contains various data patterns because such data are collected from multiple subjects. Surprisingly, the TTS-GAN model can successfully capture such different patterns and generate a very similar synthetic data set. However, the synthetic data quality may be affected by how complex the original real dataset was. We discuss this influence in section~\ref{sec:futurework}. 

The fourth and fifth rows of plots are the synthetic data from the PTB Diagnostic ECG dataset. This dataset contains two categories of heartbeat signals, normal and abnormal. 'Normal' represents the signals collected from subjects whose heartbeat signals are considered normal, without heart diseases. The abnormal signals are collected from the patients who have a different kinds of heart diseases. The fourth line of plots shows that the TTS-GAN model can learn the normal heartbeat signal patterns and generate very similar synthetic signals. The fifth line of plots shows that the TTS-GAN is able to capture the difference between different abnormal heartbeats patterns and generate a synthetic set with various signal patterns. 

The bottom two rows of plots are from the MIT-BIH Arrhythmia dataset. This dataset contains five categories of human heartbeat ECG signals. We use a single TTS-CGAN model to generate five categories of synthetic data. Due to the page limit, we only plot a few synthetic signals from two categories here. From these plots, we can see that the TTS-CGAN model can generate multiple categories of data at once and all of them are of relatively high quality. From the plots, we can also see that the synthetic signals look nosier than real signals. We will discuss this issue in section~\ref{sec:futurework}.

\subsubsection{Visualizations with PCA, t-SNE, and fusion maps}
\begin{figure*}[ht]
\centering
    \subfloat[Jumping]{\includegraphics[width=0.24\textwidth, keepaspectratio]{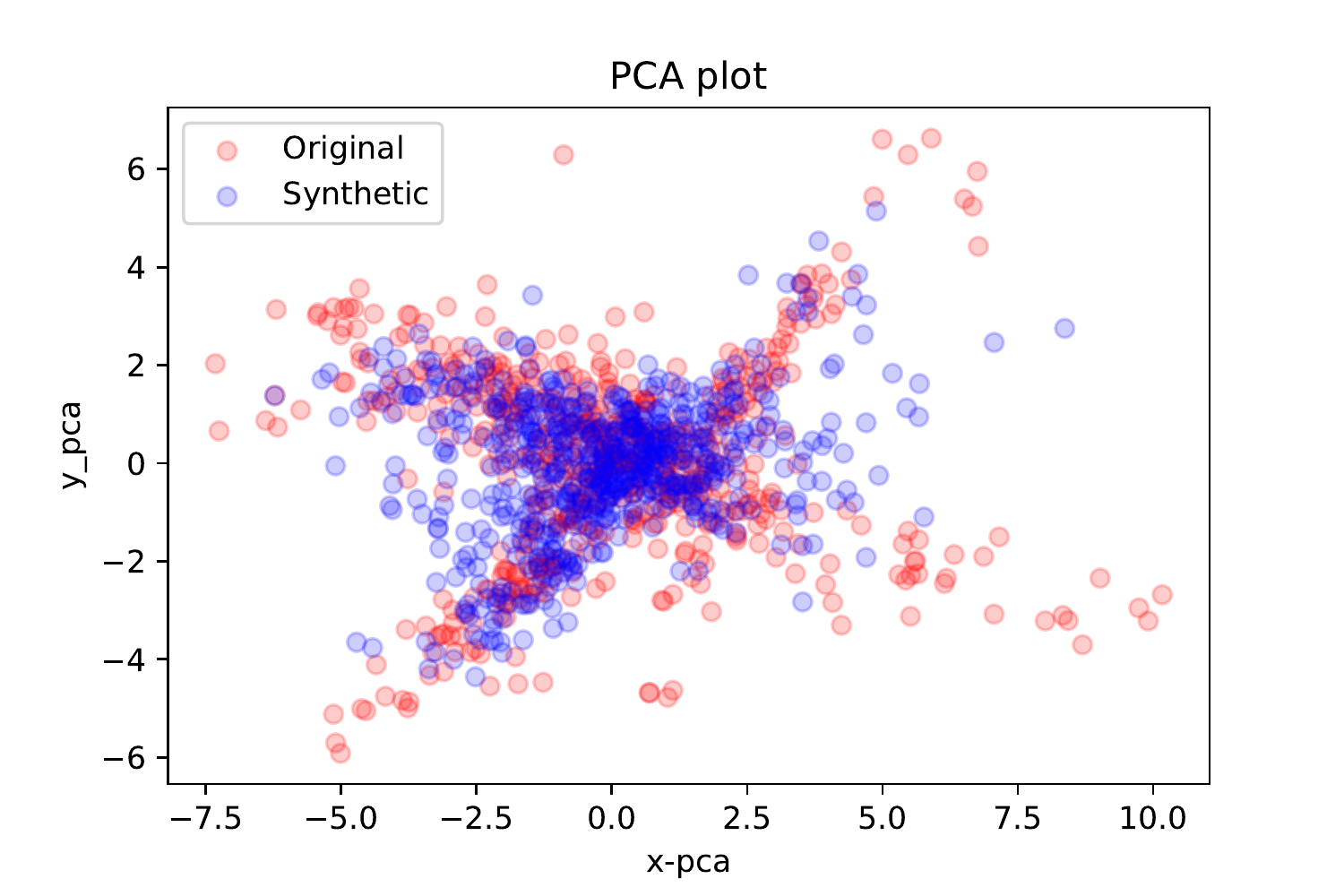}}
    \subfloat[Running]{\includegraphics[width=0.24\textwidth, keepaspectratio]{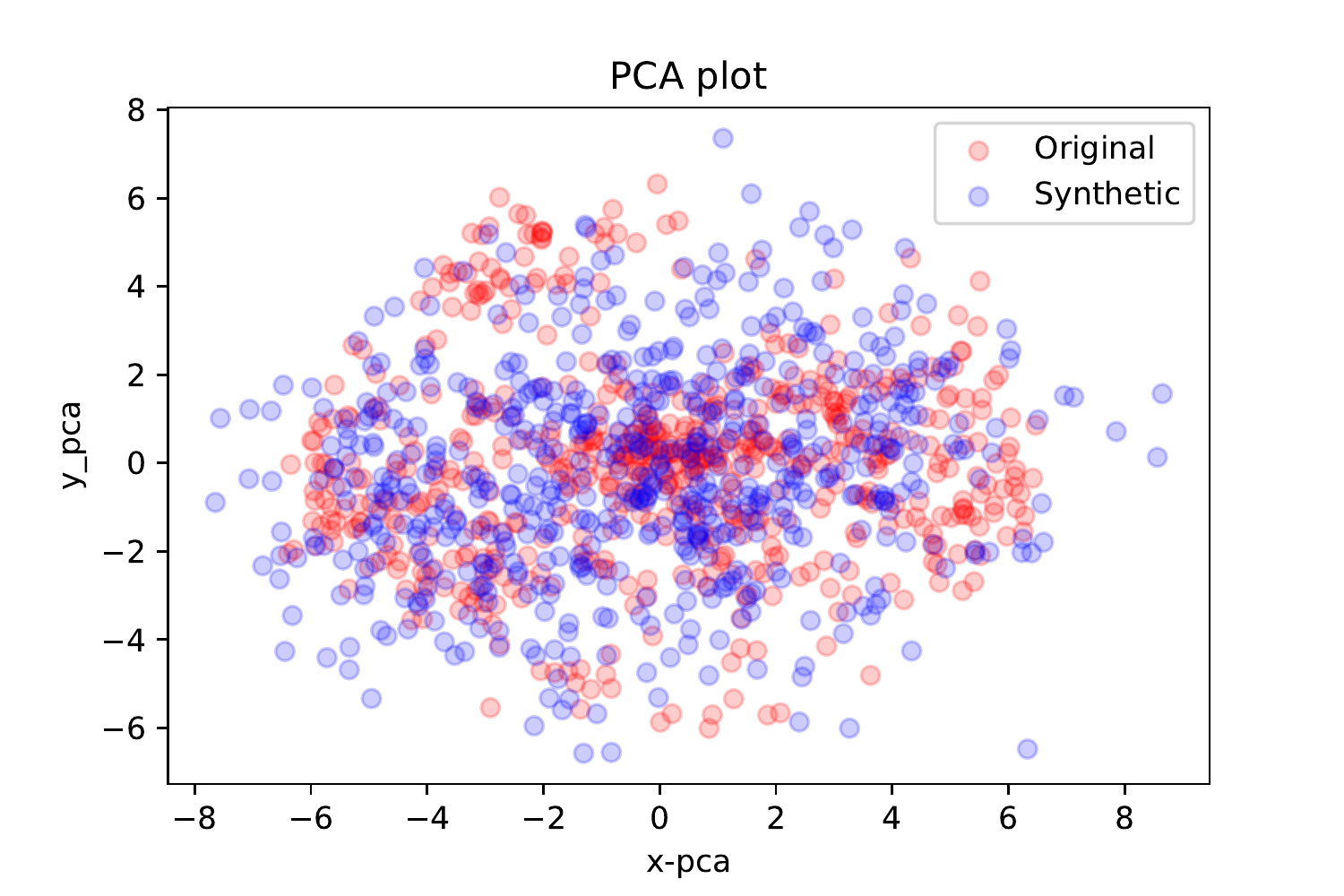}}
    \subfloat[Normal ECG]{\includegraphics[width=0.24\textwidth, keepaspectratio]{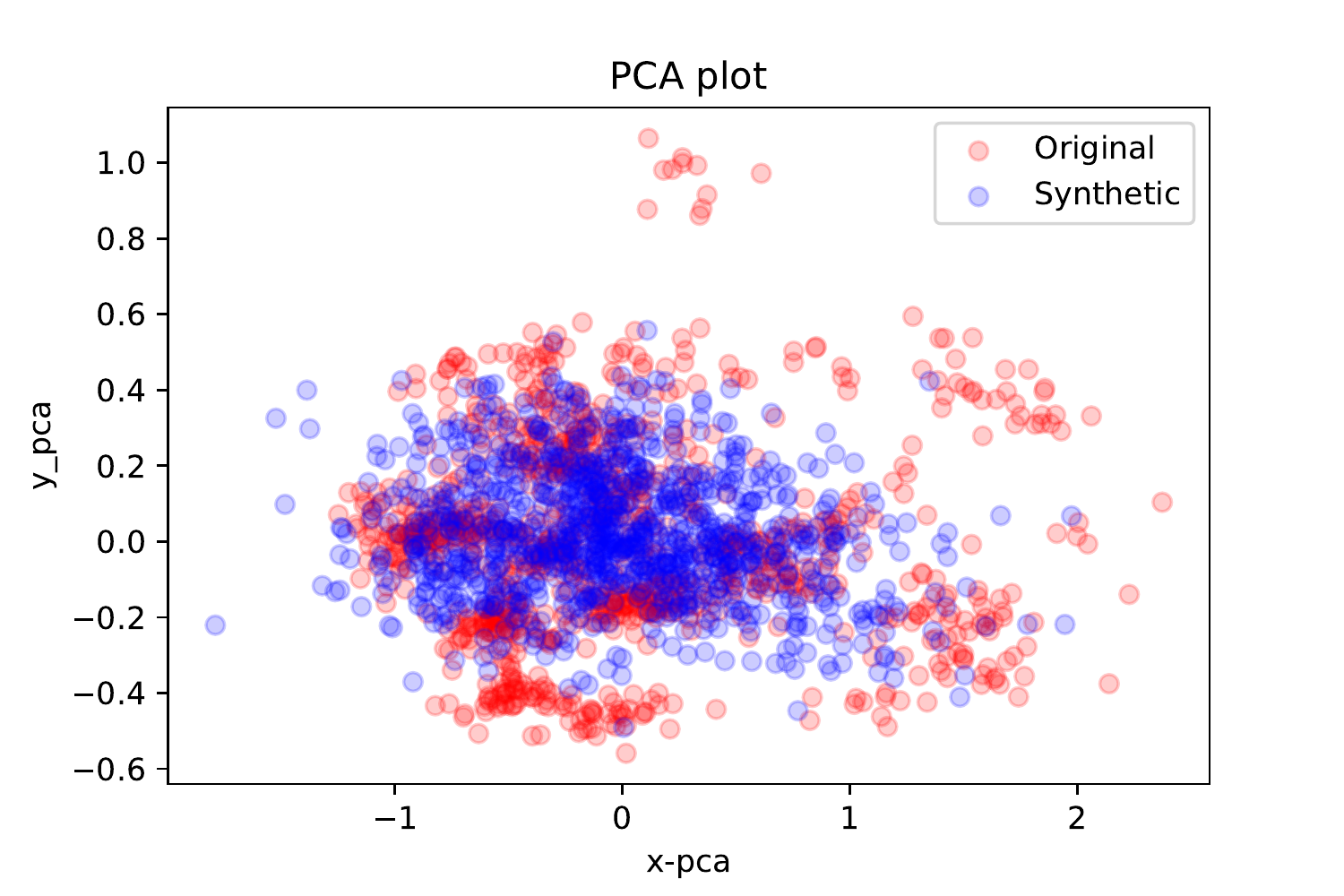}}
    \subfloat[Abnormal ECG]{\includegraphics[width=0.24\textwidth, keepaspectratio]{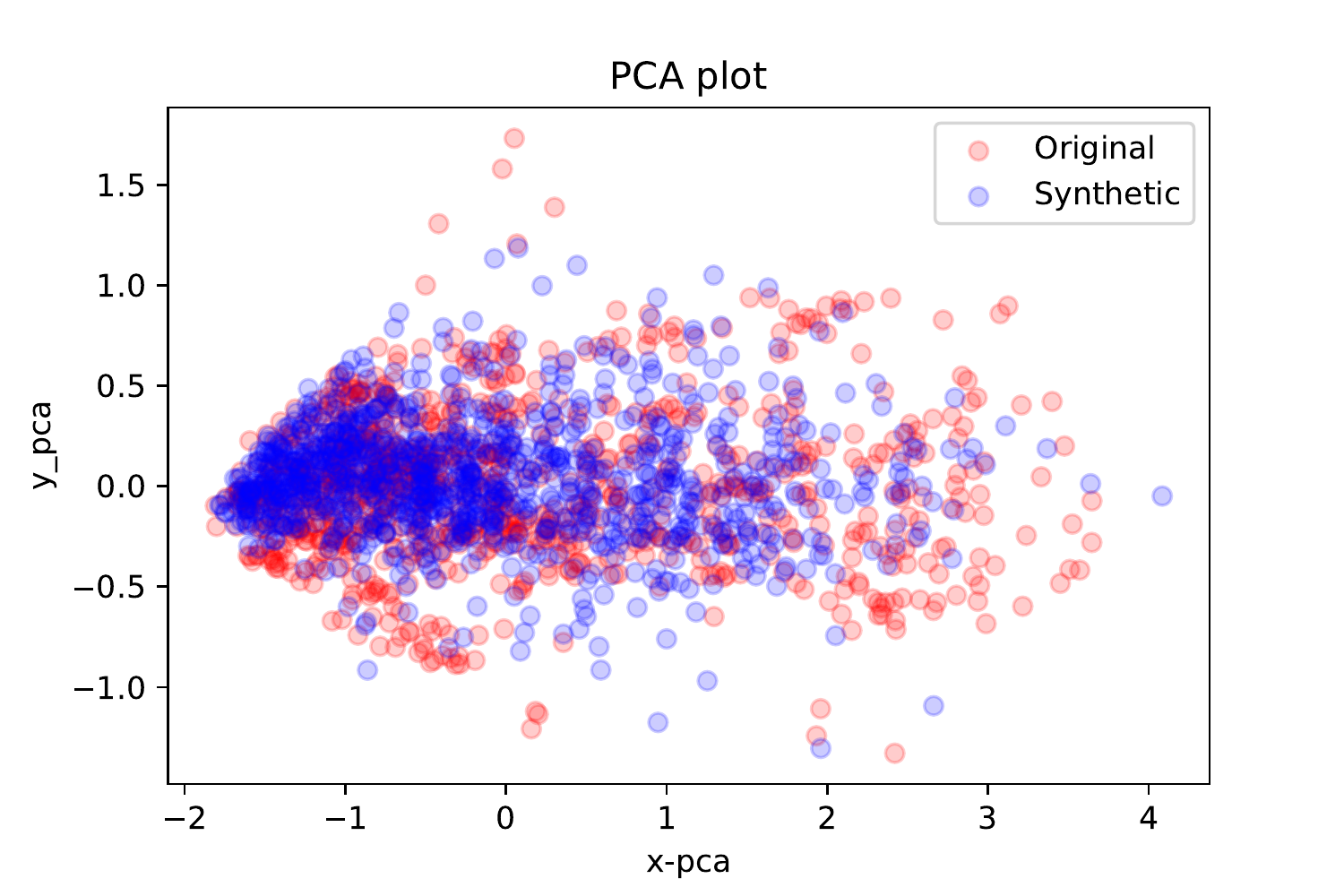}}
    
    \subfloat{\includegraphics[width=0.24\textwidth, keepaspectratio]{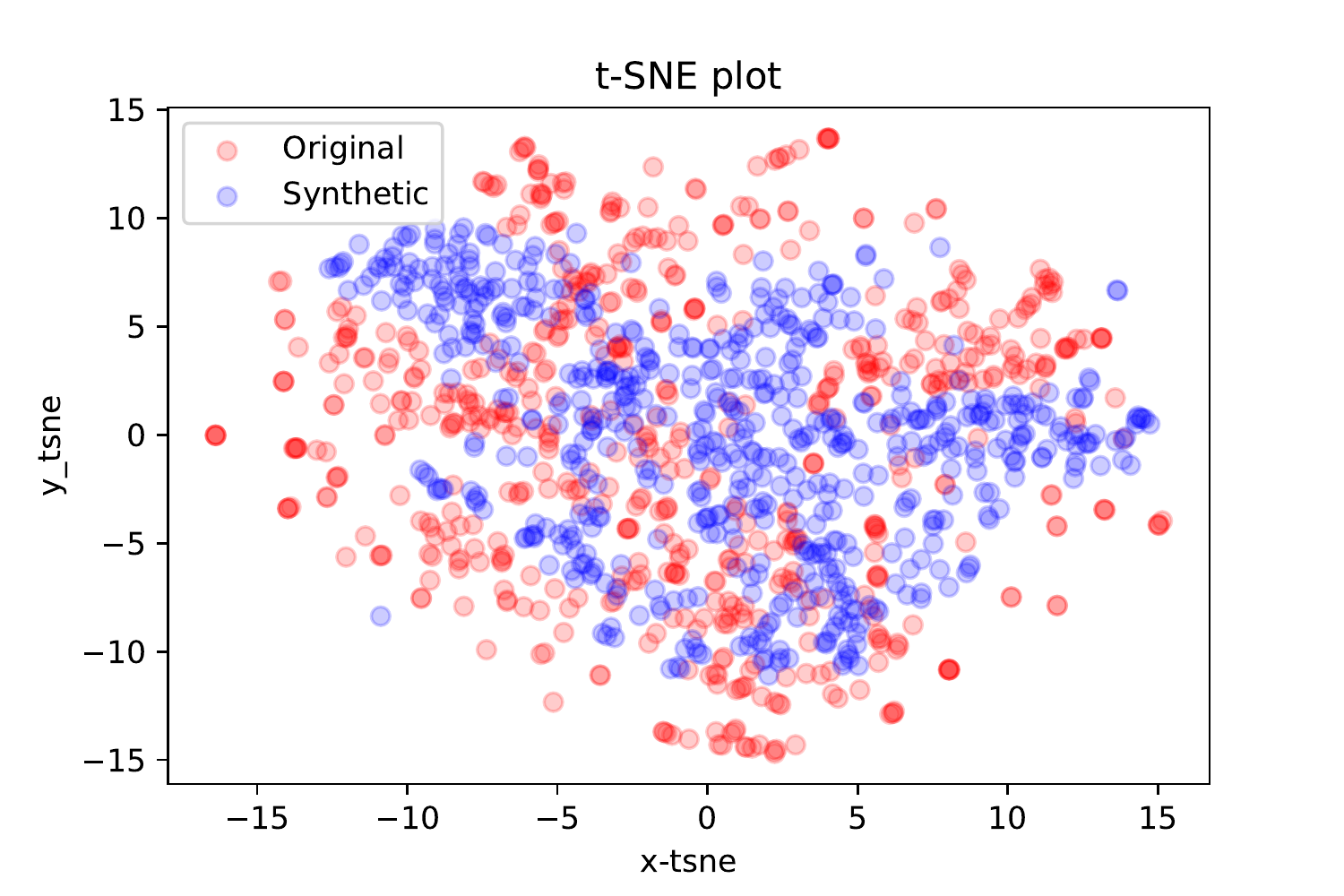}}
    \subfloat{\includegraphics[width=0.24\textwidth, keepaspectratio]{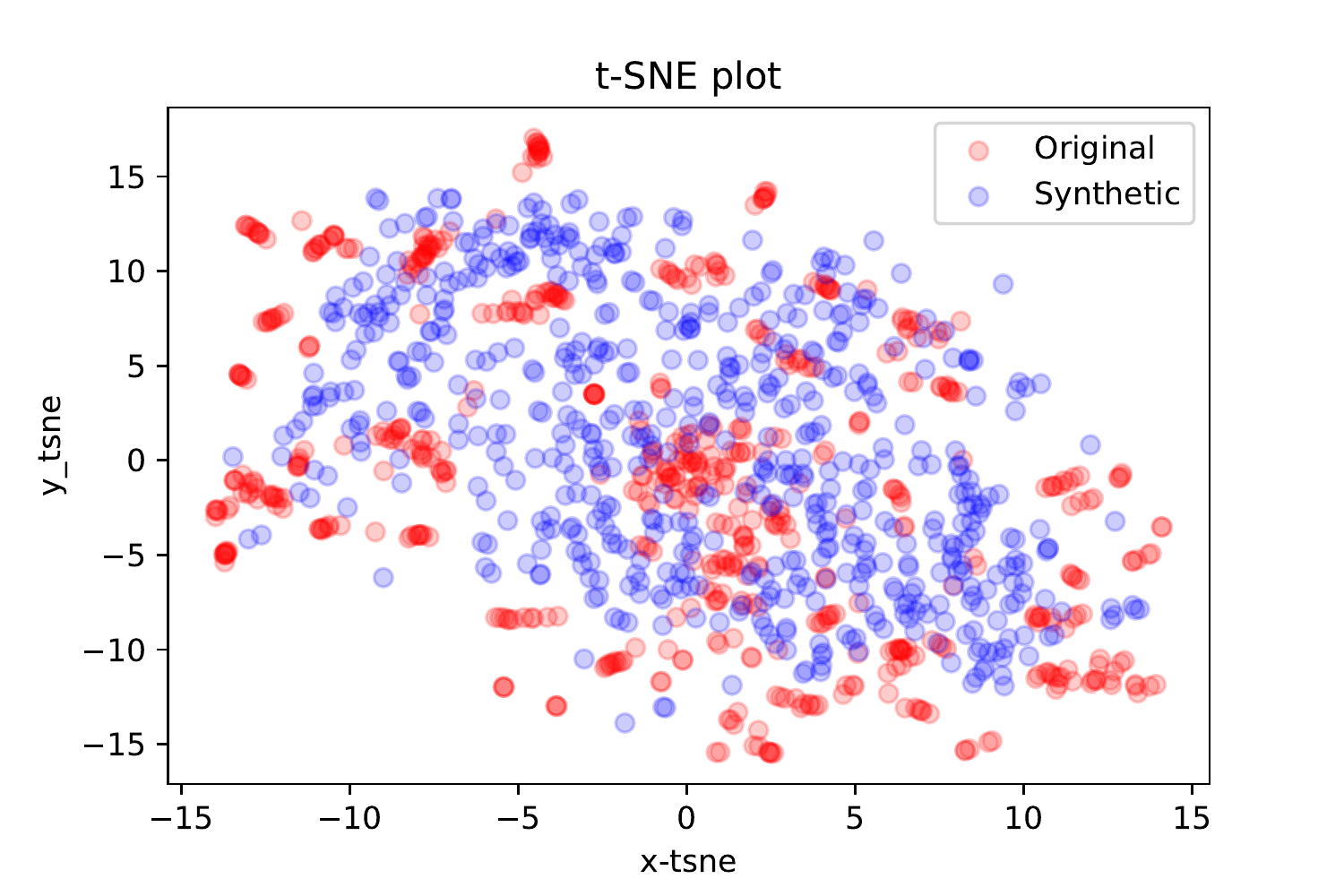}}
    \subfloat{\includegraphics[width=0.24\textwidth, keepaspectratio]{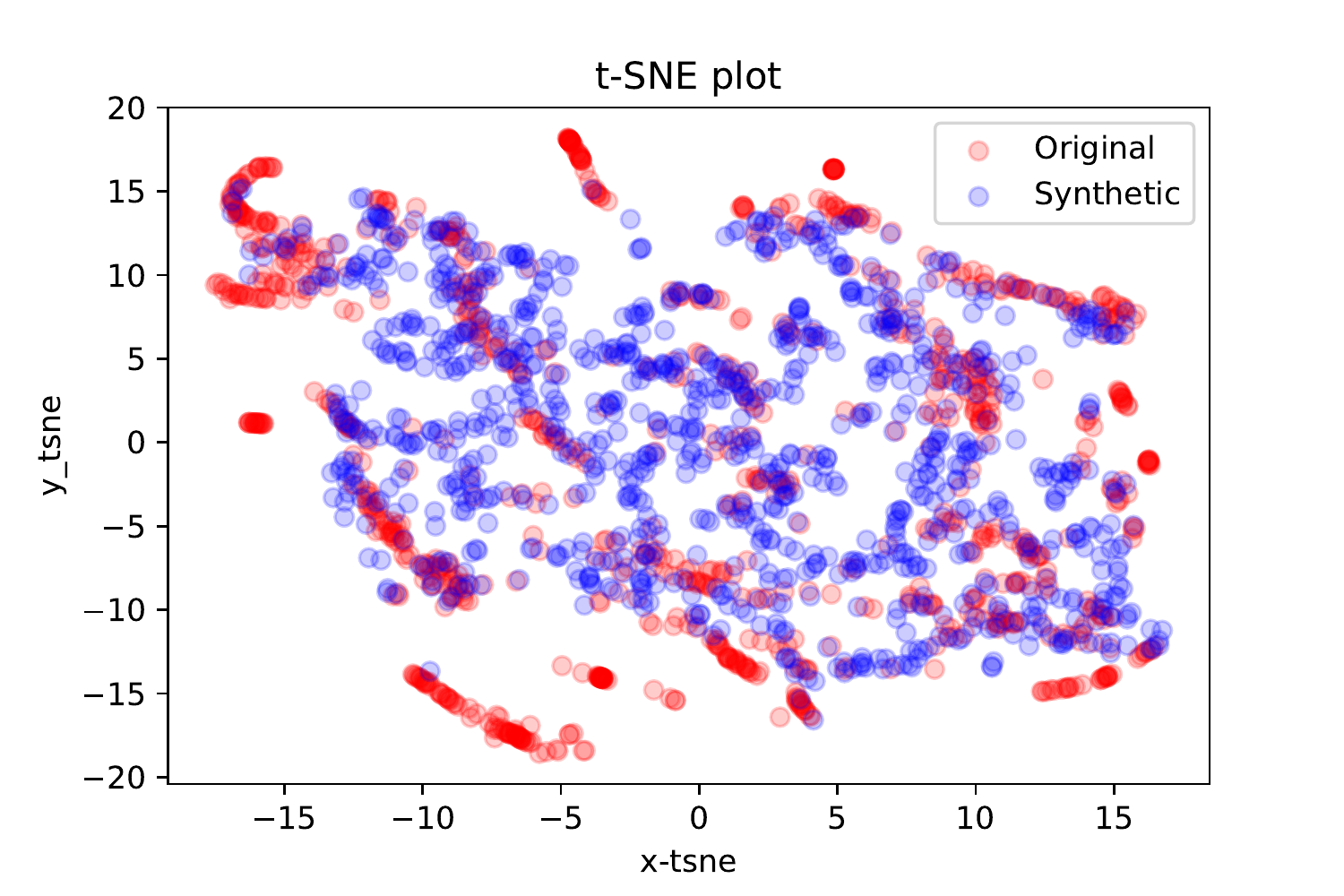}}
    \subfloat{\includegraphics[width=0.24\textwidth, keepaspectratio]{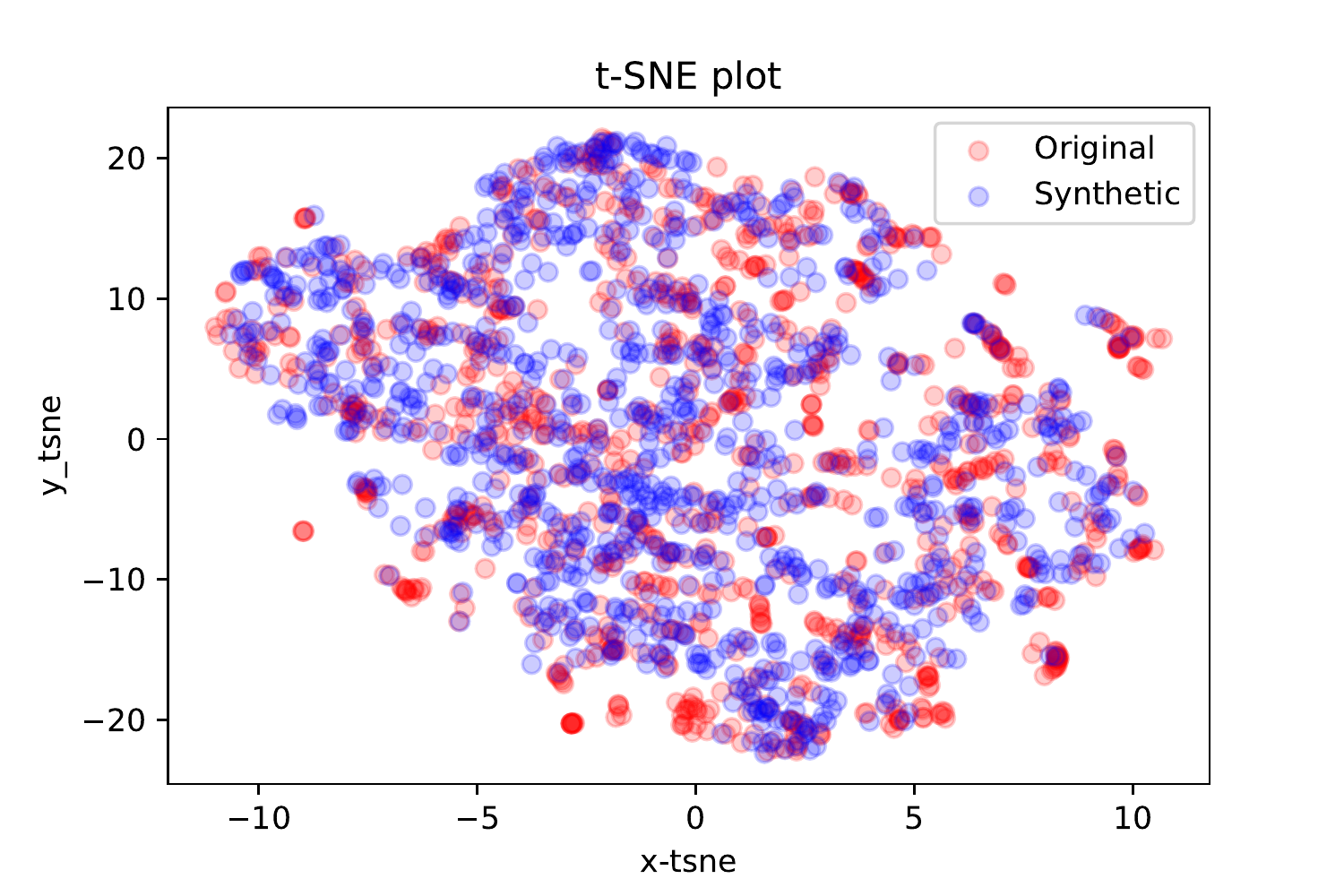}}
    
    \caption{The PCA and t-SNE visualization for real and synthetic data generated by TTS-GAN.}
    \label{fig:smallvisual}
\end{figure*}

To further illustrate the qualitative similarity between the real data and synthetic data, we plot visualization example graphs of data point distributions mapped to two dimensions using Principal Component Analysis (PCA) and t-distributed stochastic neighbor embedding (t-SNE) in Fig.~\ref{fig:smallvisual}. Plots (a) \& (b) are from UniMiB dataset; plots (c) \& (d) are from PTB Diagnostic ECG dataset. 

In these plots, each dot represents a real or synthetic data sample (sequence) value after dimensionality reduction. Red dots denote original data samples, and blue dots denote synthetic data samples generated by TTS-GAN. The top row are PCA plots from each dataset and the bottom row are t-SNE plots. Though PCA and t-SNE reduce data dimensions from different aspects, from these plots, we notice that the dots from real data and synthetic data are following a similar distribution. It indicates that the original high dimensional real and synthetic data samples are also very similar. 

\begin{figure*}[ht]
\centering
    \subfloat[Non-Ectopic Beats]{\includegraphics[width=0.20\textwidth, keepaspectratio]{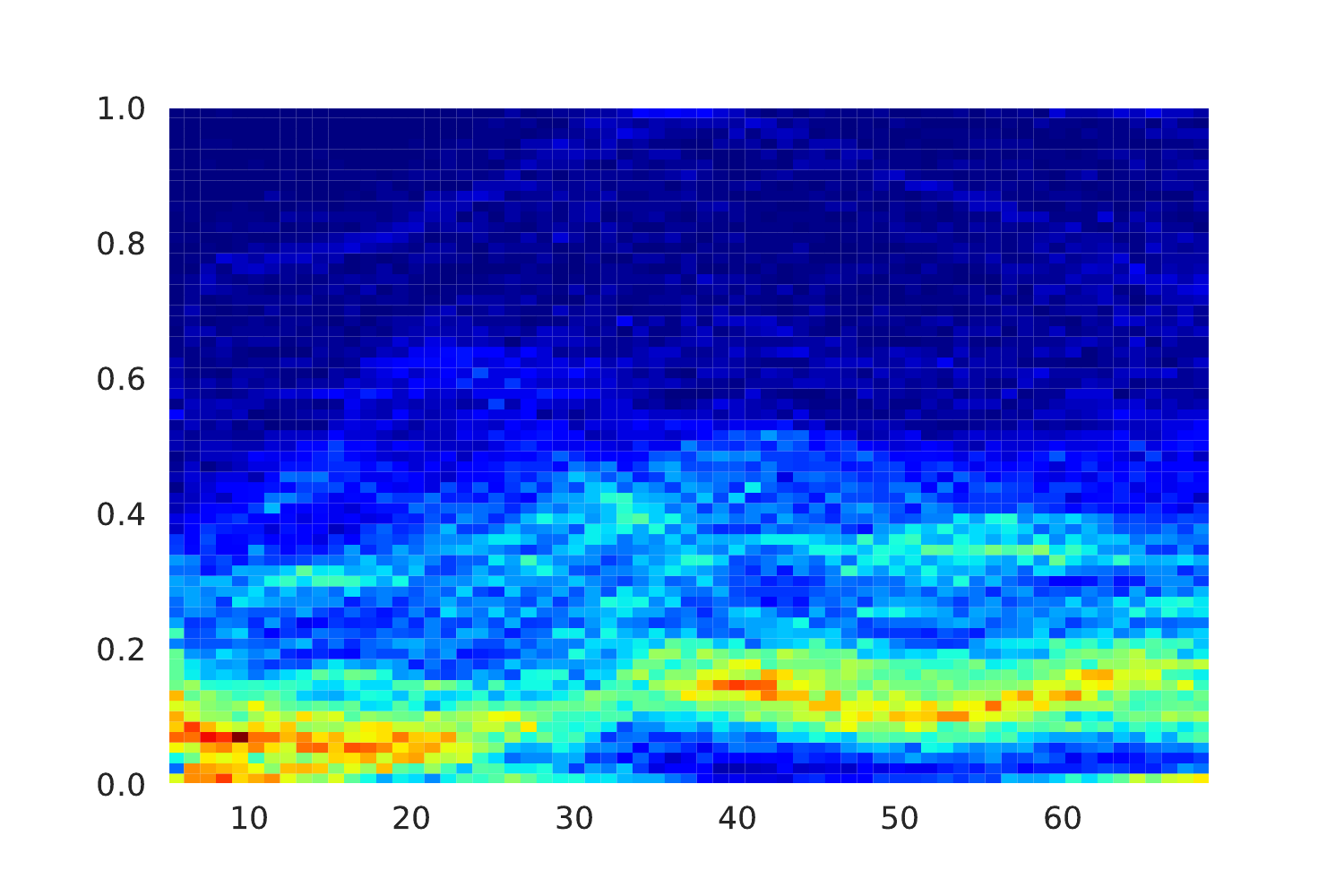}}
    \subfloat[Superventrical Ect. Beats]{\includegraphics[width=0.20\textwidth, keepaspectratio]{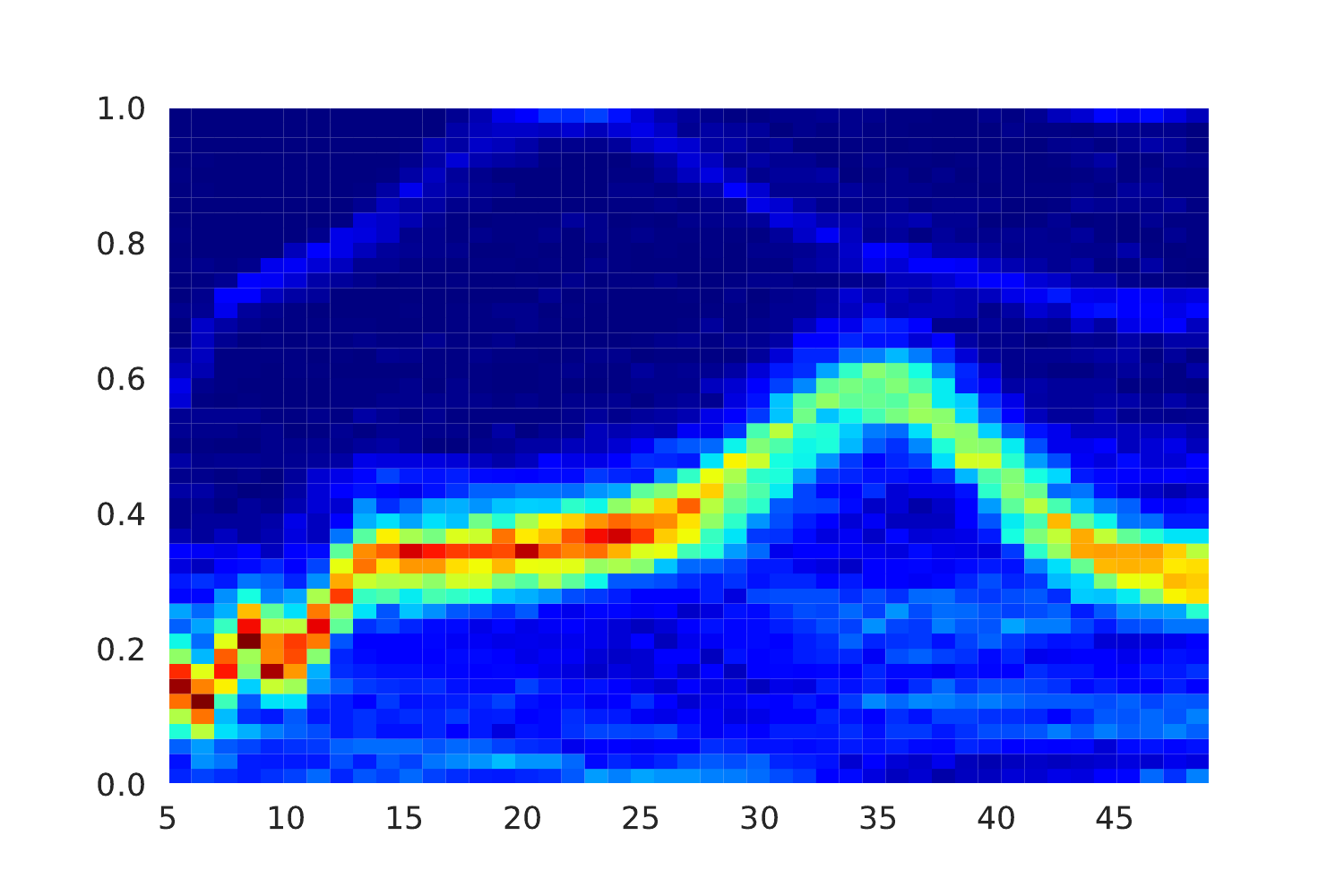}}
    \subfloat[Ventricular Beats]{\includegraphics[width=0.20\textwidth, keepaspectratio]{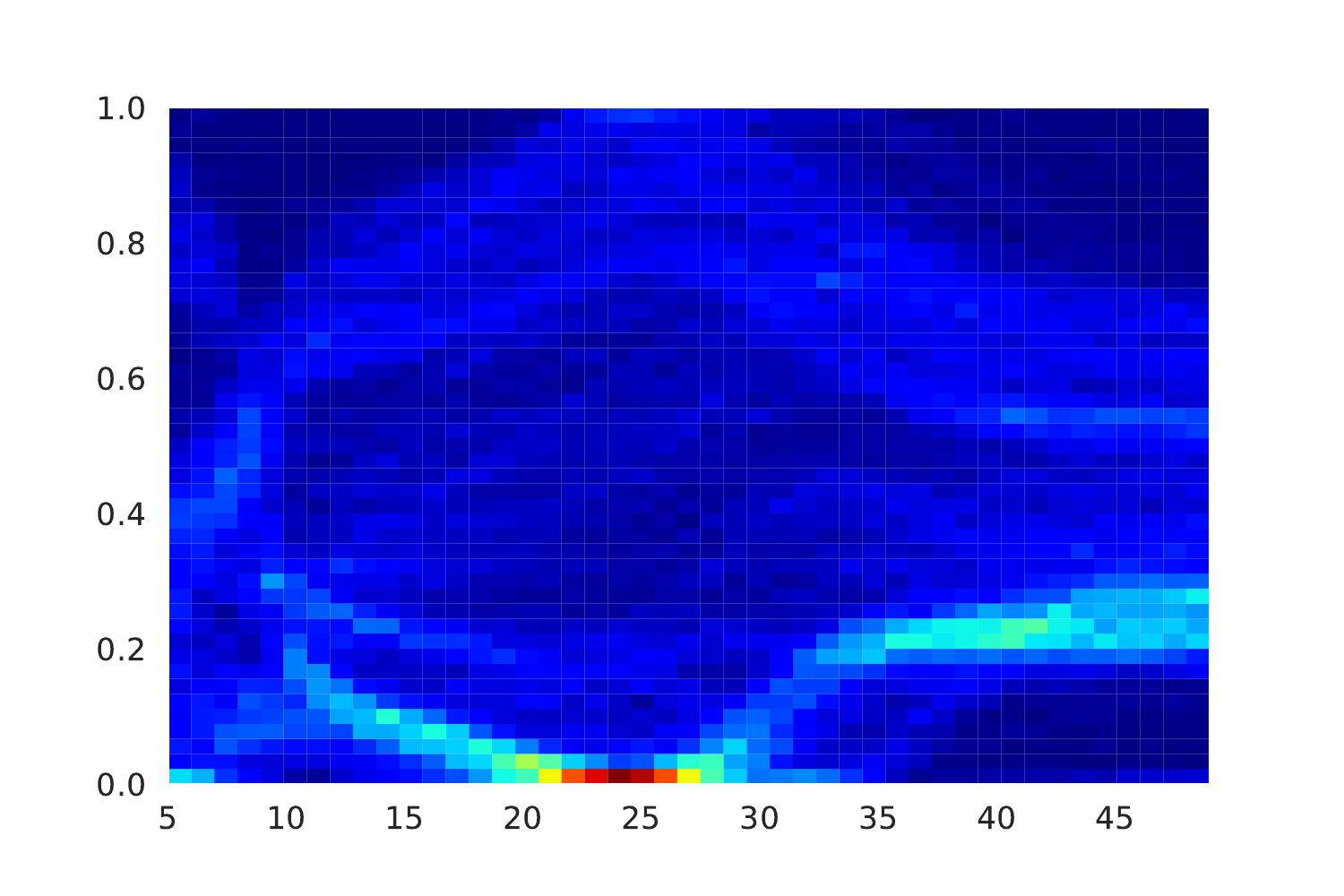}}
    \subfloat[Unknown Beats]{\includegraphics[width=0.20\textwidth, keepaspectratio]{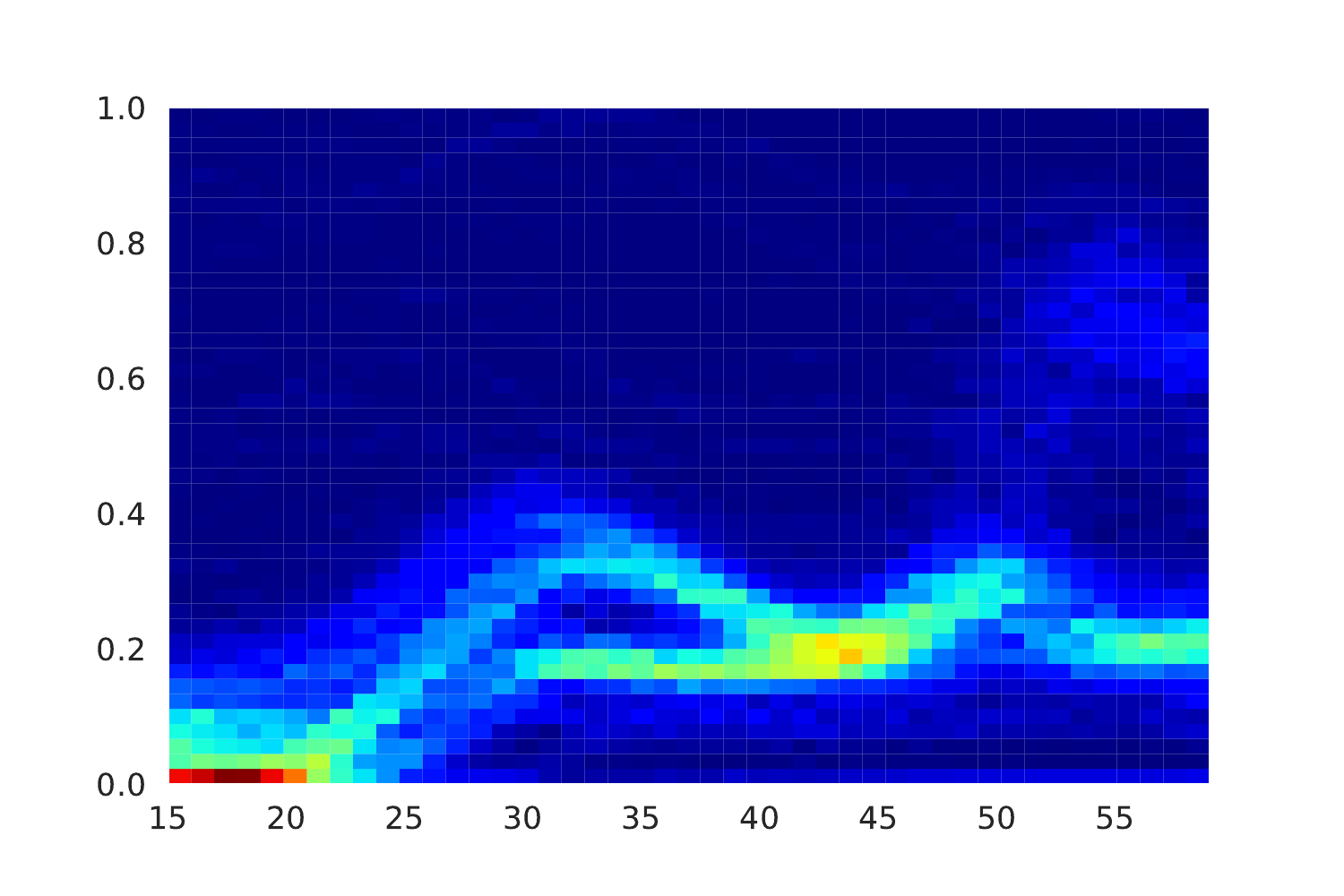}}
    \subfloat[Fusion Beats]{\includegraphics[width=0.20\textwidth, keepaspectratio]{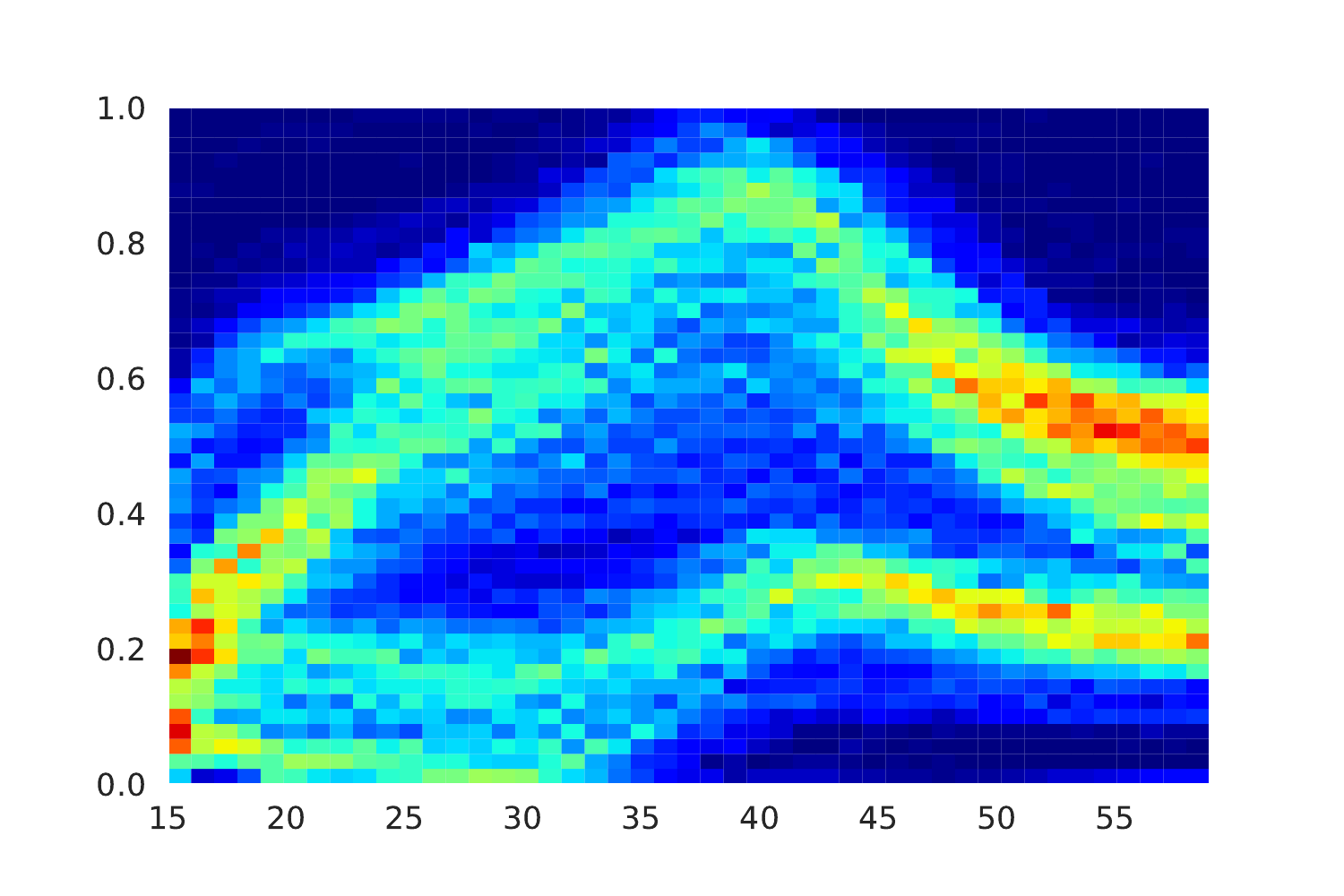}}
    
    \subfloat{\includegraphics[width=0.20\textwidth, keepaspectratio]{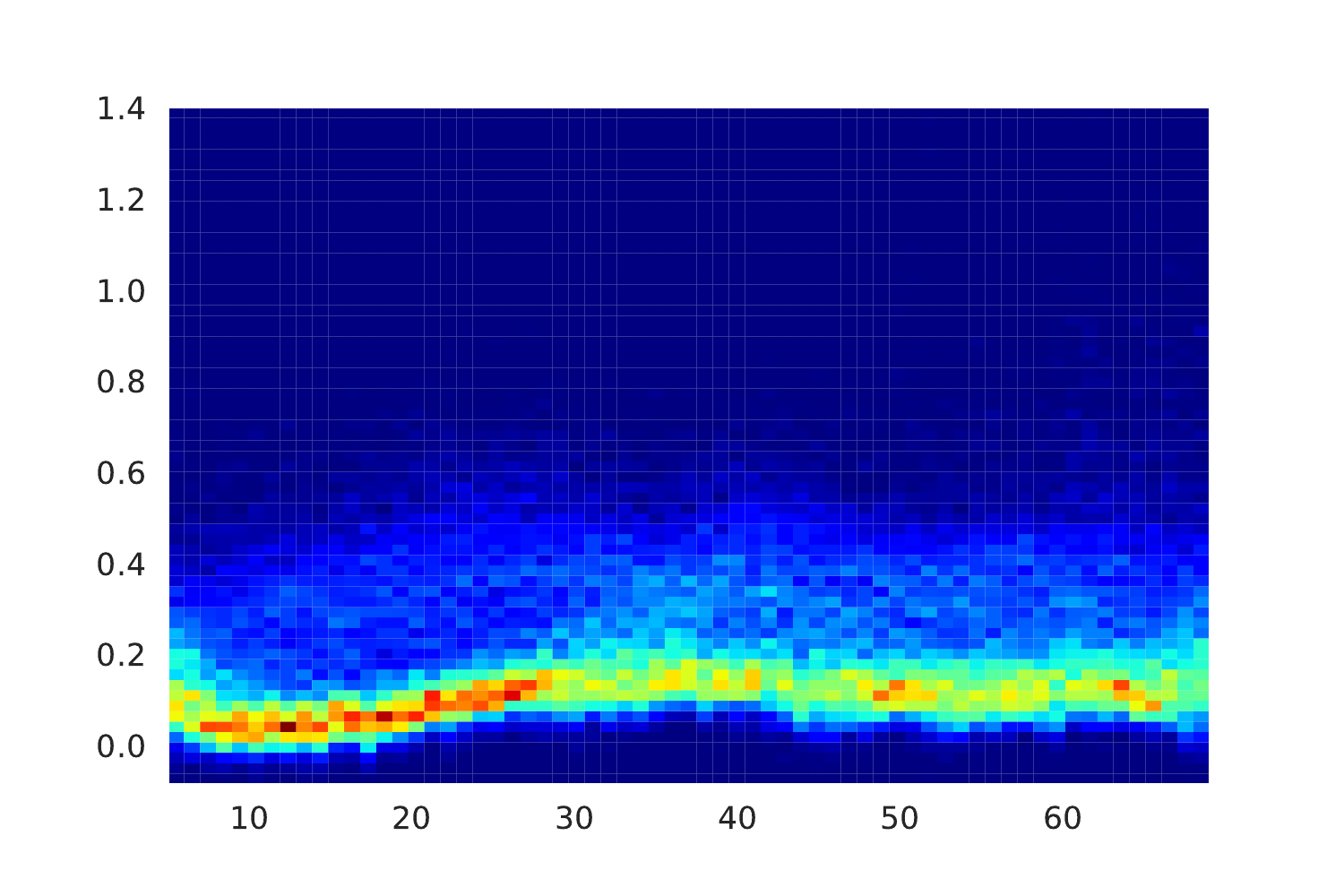}}
    \subfloat{\includegraphics[width=0.20\textwidth, keepaspectratio]{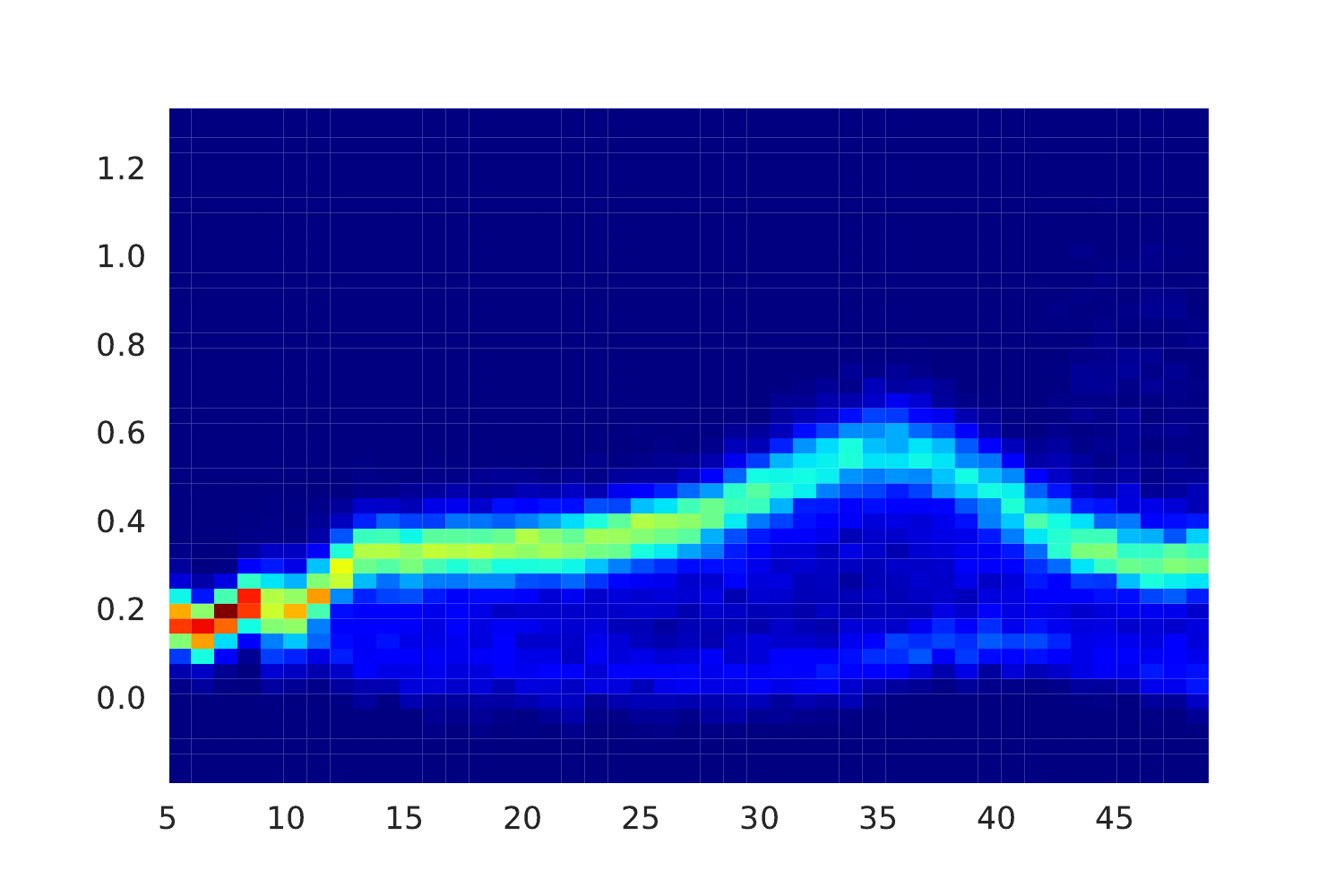}}
    \subfloat{\includegraphics[width=0.20\textwidth, keepaspectratio]{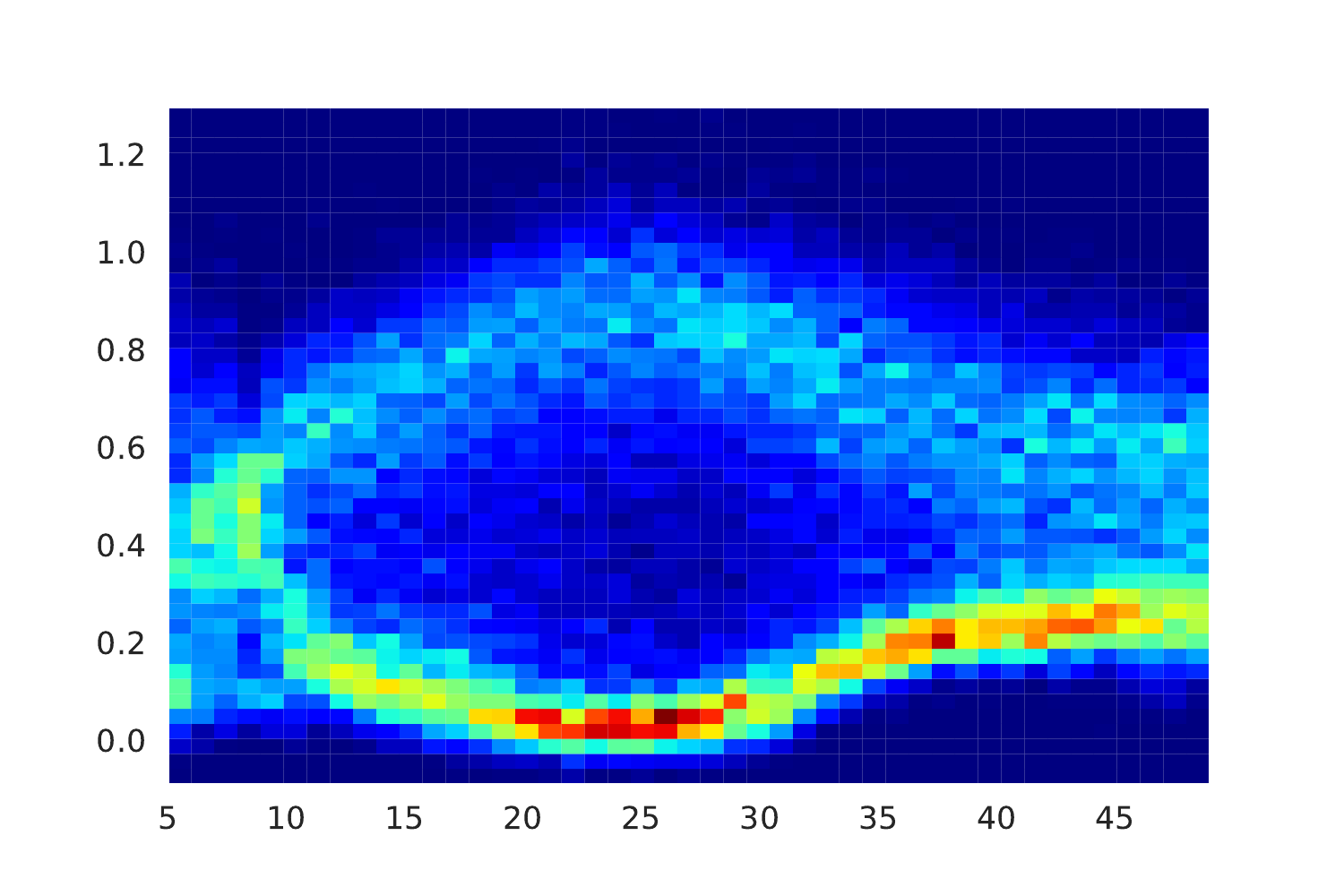}}
    \subfloat{\includegraphics[width=0.20\textwidth, keepaspectratio]{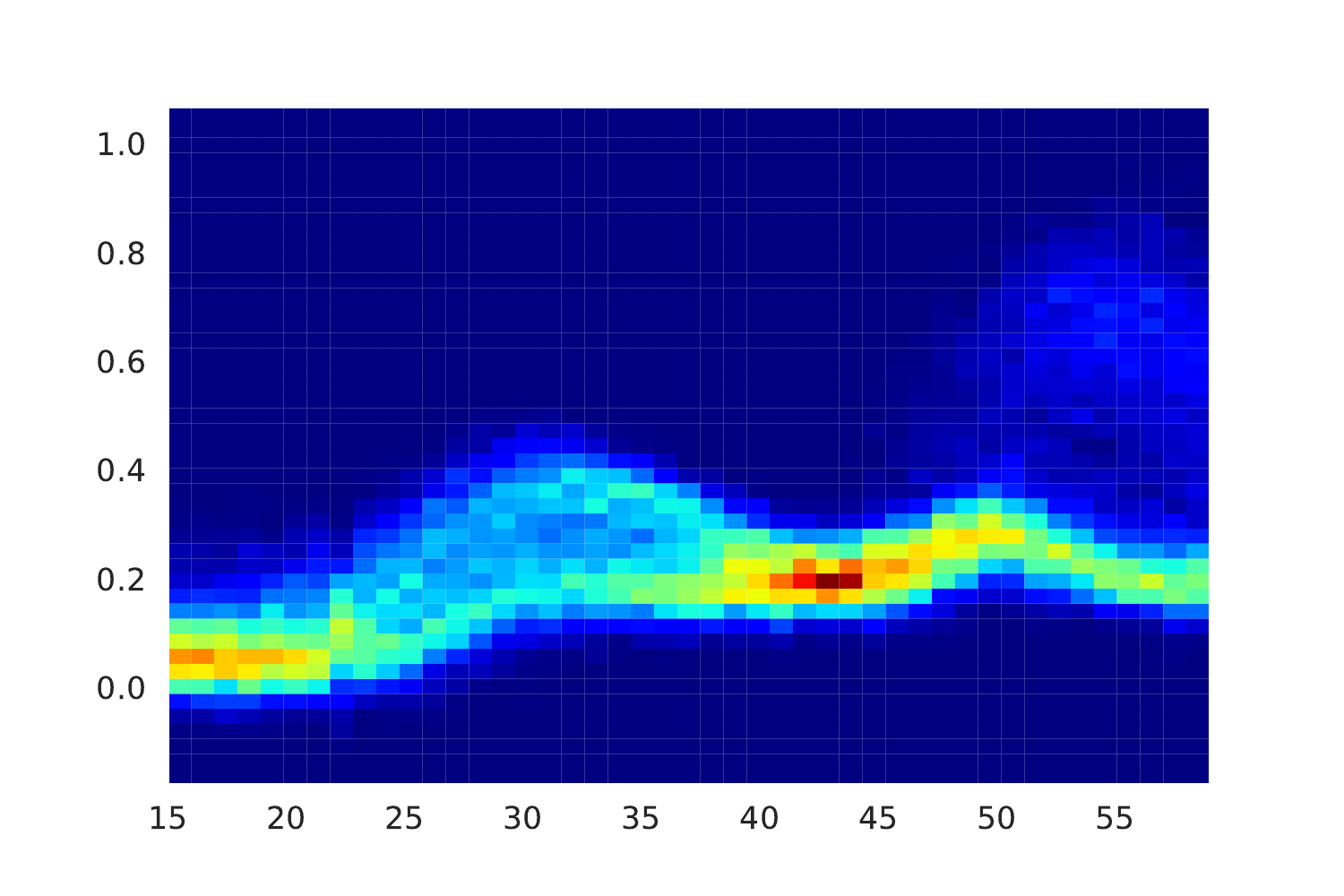}}
    \subfloat{\includegraphics[width=0.20\textwidth, keepaspectratio]{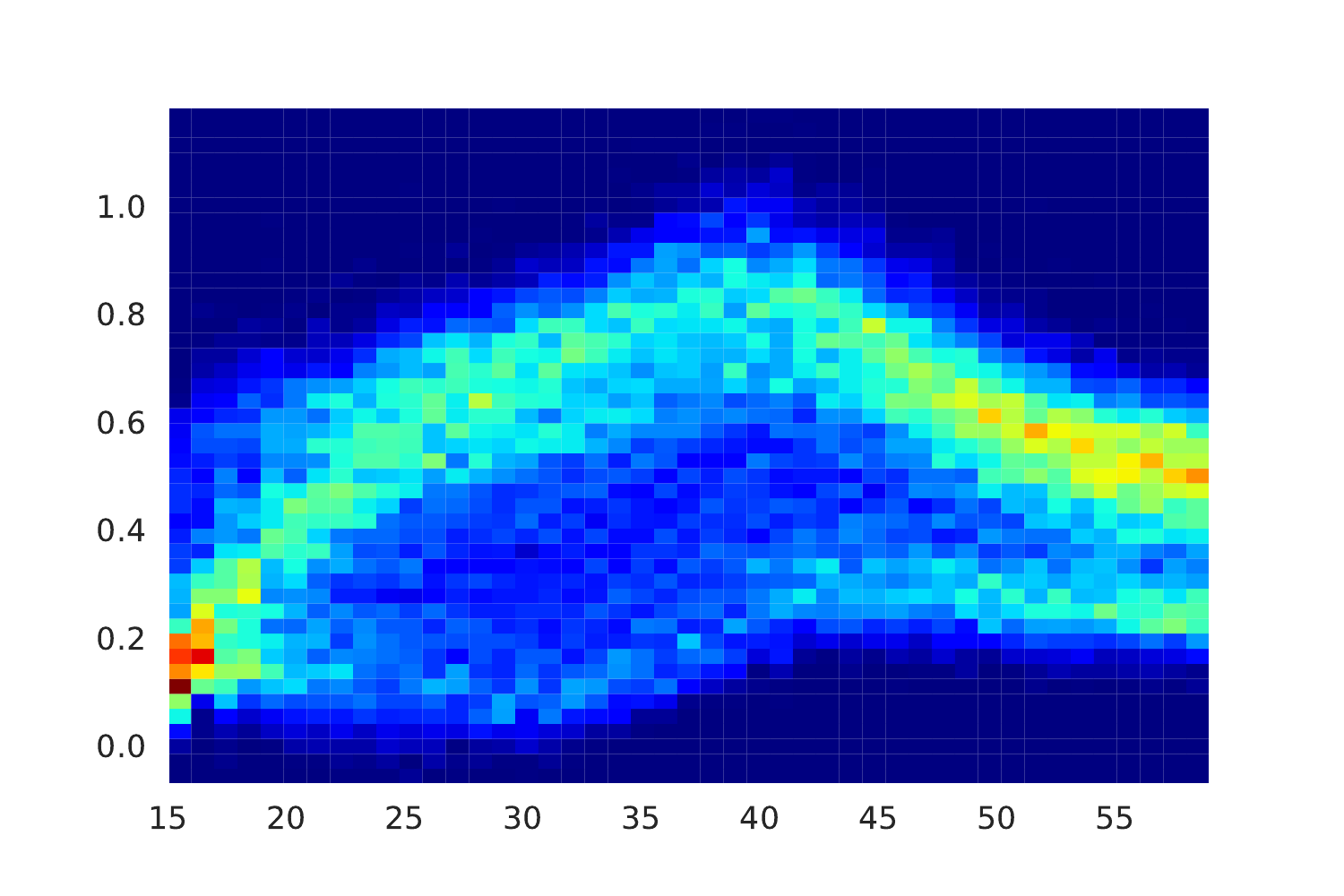}}
    
\caption{PTB Diagnostic ECG dataset real signals and synthetic signals fusion heatmaps. The top row images are from real signals and the bottom row images are from synthetic signals.}
\label{fig:HearBeatVisual}
\end{figure*}

In Fig.~\ref{fig:HearBeatVisual}, we plot TTS-CGAN generated five different categories of heartbeat signals to fusion maps. Each fusion map is a combination of 1000 raw signals from a data category. Those maps are represented as heatmaps where the brighter color indicates more data points having similar values. From these plots we can see a similar data value distribution of real signals and synthetic signals. Since the 1000 signals samples we used to draw the fusion maps are randomly selected from the real and synthetic sets, their signal values distribution may be slightly different. For example, (a) and (c) real and synthetic fusion maps look a little different.

This fusion map strategy is a more direct way to observe data points value distribution from multiple signals. It works well for heart-beat signals, but it may be fail to picture such similarities when a categories of data have multiple dimensions and have various signal patterns, such as the Jumping and Running signals from the UniMiB datasets.

\subsubsection{Quantitative analysis}
We use the Wavelet Coherence score to demonstrate the similarity between a set of synthetic signals and a set of real signals, as introduced in section~\ref{subsec:wcoh}. The higher the score the more similar these two sets. Table~\ref{tbl:wcoh} shows the wavelet coherence scores we get from multiple datasets. We also compare our GAN model performance with several state-of-the-art time-series GAN models: 

\begin{itemize}
    \item C-RNN-GAN~\cite{mogren2016c} It is the first work to tackle the problem of sequential data generation using GAN architectures. This work designed a two-layer LSTM network for both the generator and the discriminator. The generator takes noise vectors as inputs and sequentially generates time-series data from the previous time step.   
    \item RCGAN~\cite{esteban2017real} This work used a similar approach and GAN model architecture as C-RNN-GAN. But it added conditional information to the input to apply some control to the data generated. 
    \item TimeGAN~\cite{yoon2019time} This work introduced a novel framework for time-series data generation. Its GAN model learns to generate synthetic data in a latent space. And to train this framework, the authors also leveraged supervised and unsupervised loss to help control the quality of generated data. 
    \item SigCWGAN~\cite{ni2020conditional} It is the most recent time-series data GAN model generation work. It adds conditional information to control the generated data and leverage Wasserstein loss to stable the GAN model training process.  
\end{itemize}

Our work combines the advantages of the above works and introduces some new strategies to generate to our best knowledge the highest quality time-series data so far. Our model handles the inputs more like the C-RNN-GAN~\cite{mogren2016c} network and adds conditional information like RCGAN~\cite{esteban2017real}. We also use Wasserstein loss like SigCWGAN~\cite{ni2020conditional} to help the model training. A major difference is that the previous works used recurrent networks but we, for the first time, adjust the transformer model to generate time-series data. We also conduct thorough experiments to study the best strategy to add conditional information to the GAN model (Section~\ref{subsec:labelembedding}). The TimeGAN~\cite{yoon2019time} introduced framework is compatible with our GAN models. It is possible to use our GAN model architecture with TimeGAN framework together to generate even higher quality synthetic time-series data.  

\begin{table}
\centering
\begin{tabular}{c|c|c|c|c}
\hline
Models           & Sine           & Jumping        & ECGNormal      & FusionBeats    \\ \hline
C-RNN-GAN~\cite{mogren2016c}            & 9.37           & 40.29          & 13.93          & 25.51          \\
RCGAN~\cite{esteban2017real}            & 6.78           & 38.85          & 13.97          & 22.97          \\
TimeGAN~\cite{yoon2019time}          & 7.17           & 39.42          & 13.97          & 21.98          \\
SigCWGAN~\cite{ni2020conditional}         & 8.75           & 41.02          & 14.92          & 22.87          \\
\textbf{TTS-CGAN} & \textbf{10.03} & \textbf{47.64} & \textbf{23.38} & \textbf{65.64} \\ \hline
\end{tabular}
\caption{Wavelet coherence score, the higher the better. Note: the sores are not cross-comparable among different datasets.}
\label{tbl:wcoh}
\end{table}

We choose several datasets with various data properties to compare our GAN model's performances to other baseline models. Those datasets are Sine (Simulated Sinusoidal Waves), a multi-channel stationary dataset. Jumping set from UniMiB dataset, a multi-channel non-stationary dataset. ECGNormal from PTB Diagnostic ECG dataset and Fusion Beats from MIT-BIH Arrhythmia dataset. They are both human heartbeat ECG signals. We train TTS-GAN model on the first three datasets and train TTS-CGAN model on the last dataset. The wavelet coherence scores are computed based on a set of real data and a set of synthetic data from each dataset. From TABLE~\ref{tbl:wcoh}, we can see the TTS-CGAN models all get higher scores than the other baseline models. It shows our GAN models can generate more realistic synthetic signals than the other GAN models. The ranks of these models performance also align with the results shown in the paper~\cite{ni2020conditional}.

\subsection{A Case Study of the GAN Generated Data}
\begin{figure}[ht]
\centering

    \subfloat[5000 real]{\includegraphics[width=0.49\columnwidth, keepaspectratio]{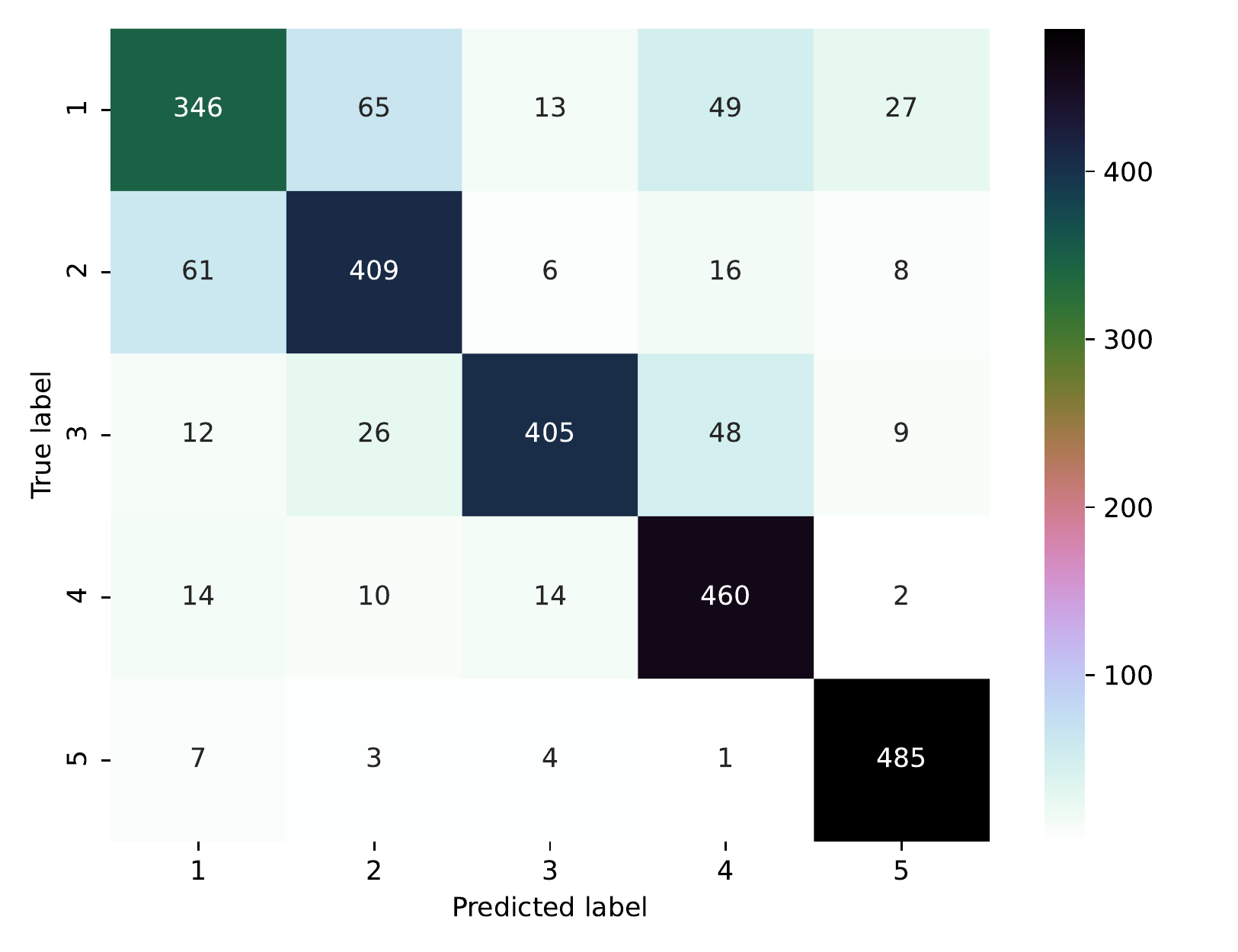}\label{subfig:realCM}}
    \subfloat[5000 synthetic]{\includegraphics[width=0.49\columnwidth, keepaspectratio]{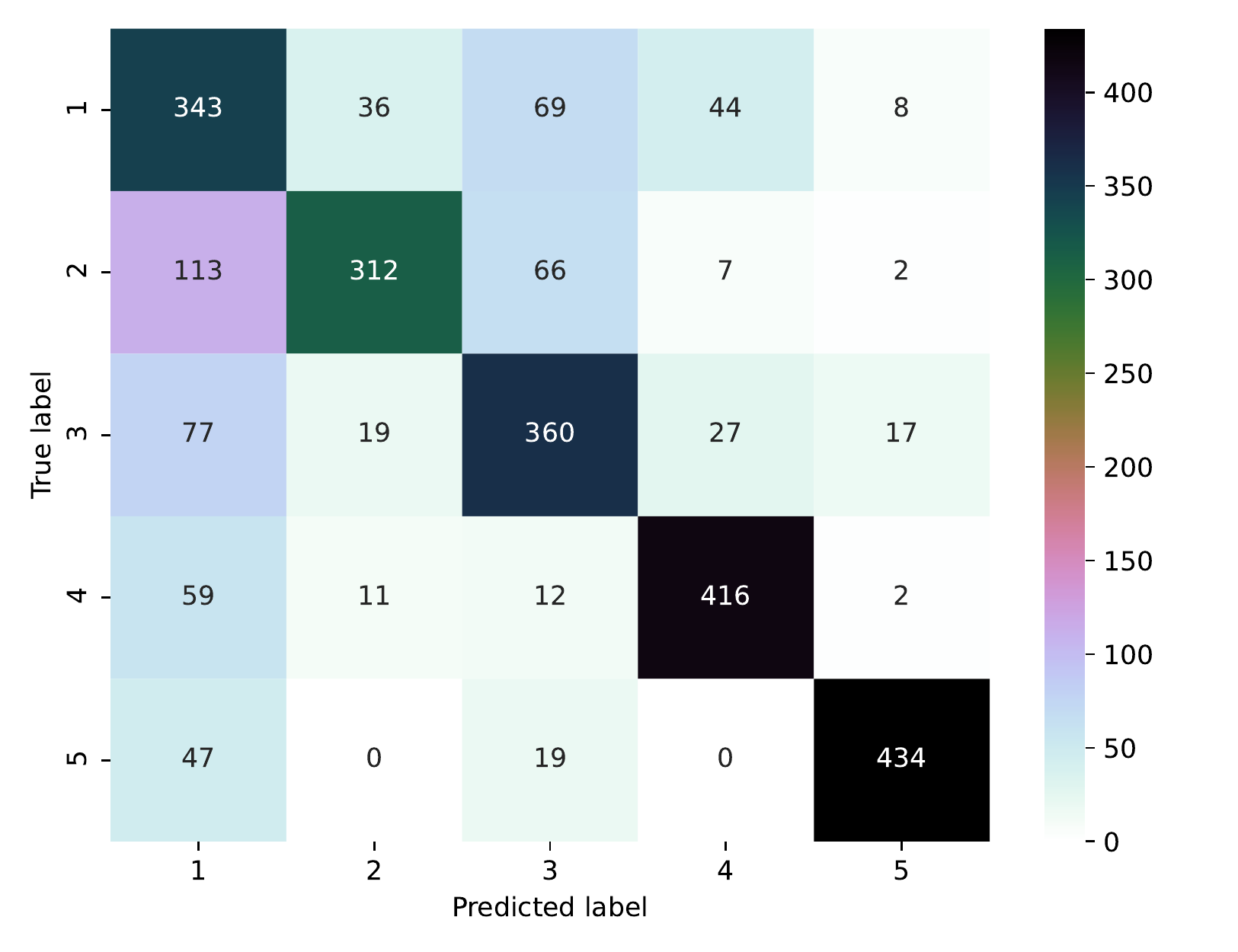}\label{subfig:synCM}}
    
    \subfloat[1000 real]{\includegraphics[width=0.49\columnwidth, keepaspectratio]{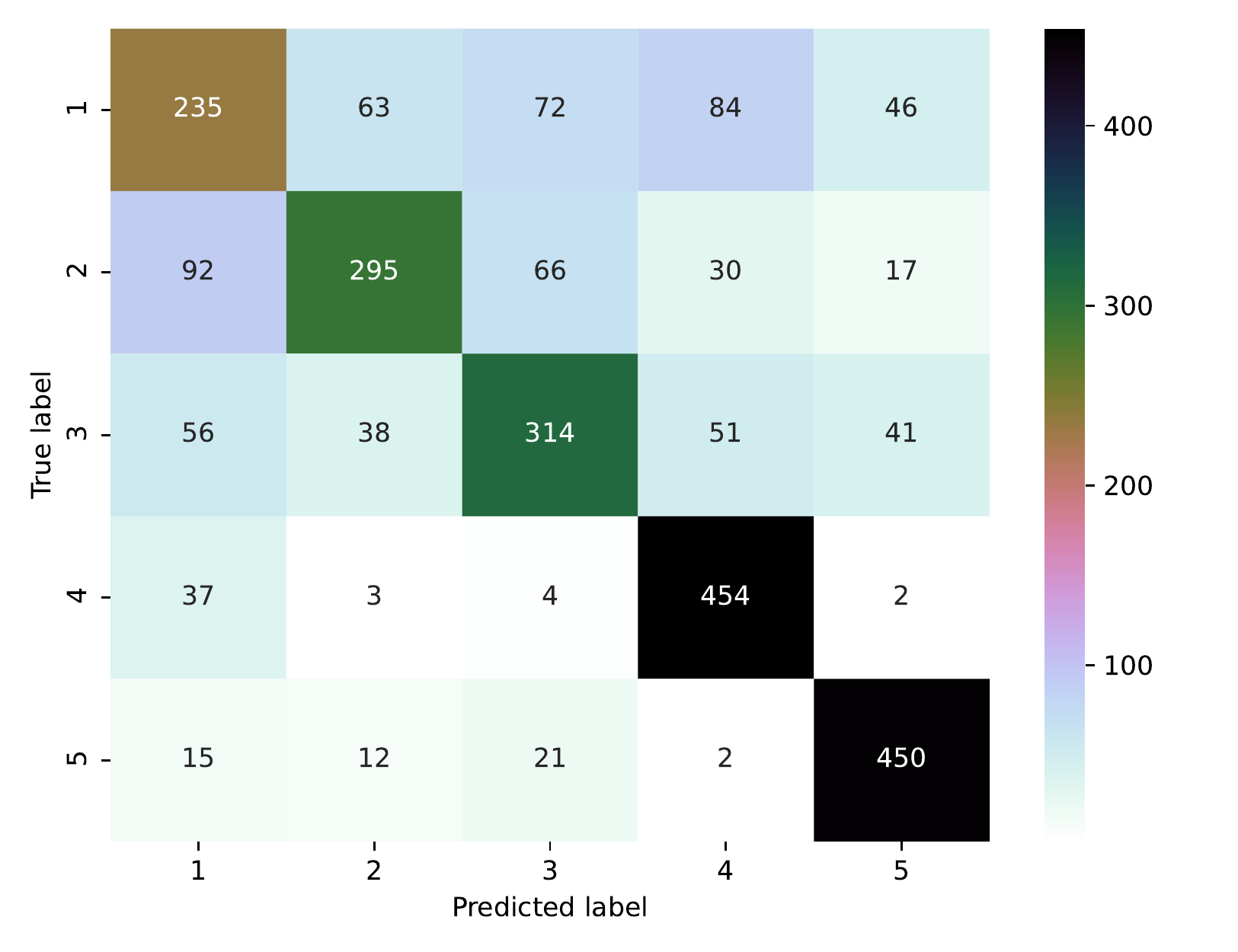}\label{subfig:realsmallCM}}
    \subfloat[1000 real \& 4000 synthetic]{\includegraphics[width=0.49\columnwidth, keepaspectratio]{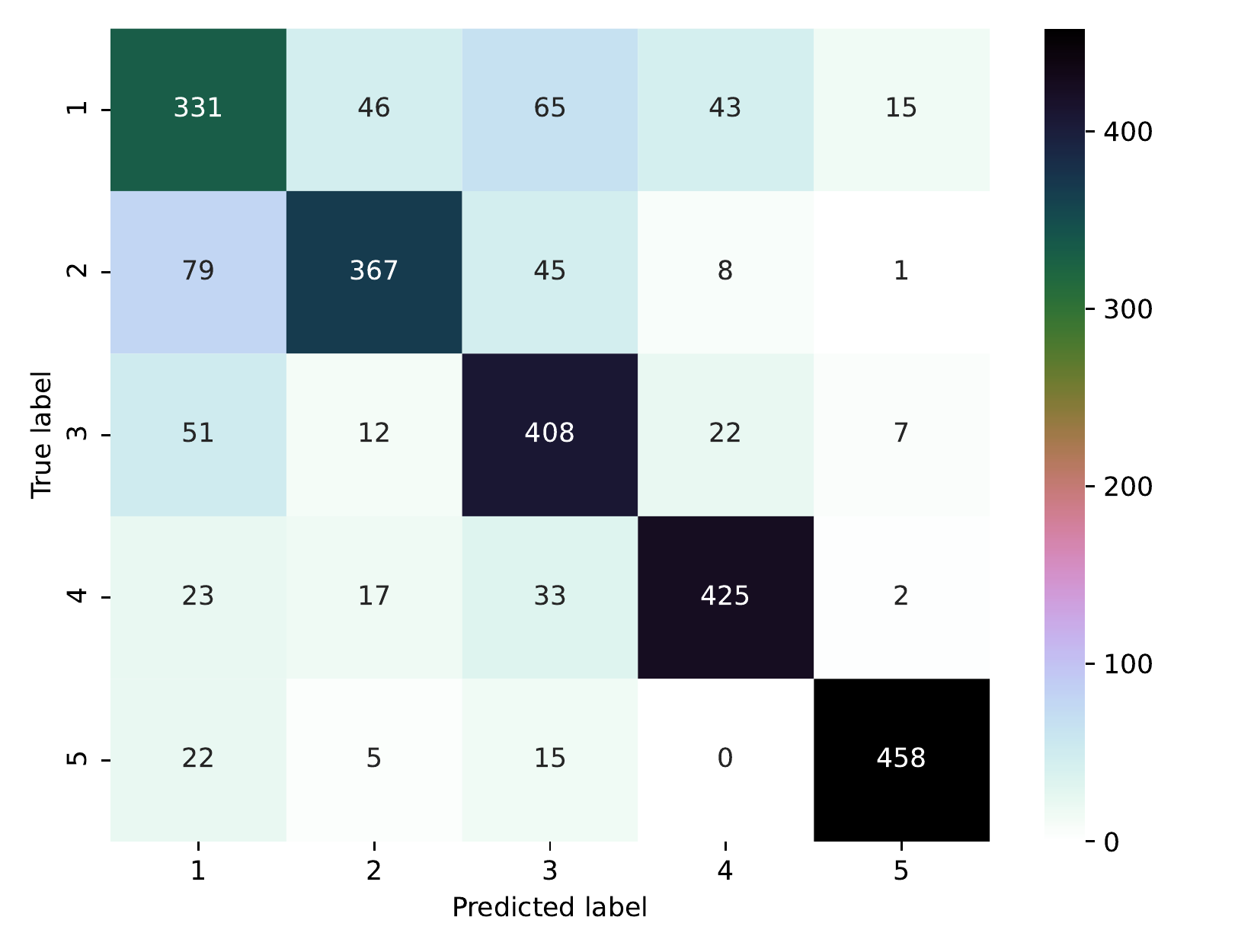}\label{subfig:mixedCM}}
    
\caption{The confusion matrix of training a classifier with different data sources and testing on the same dataset.}
\label{fig:CM}
\end{figure}

We conduct a simple classification experiment on the MIT-BIH Arrhythmia Database. It contains five categories of heartbeat signals. They are (1) Non-Ectopic Beats, (2) Superventrical Ectopic Beats, (3) Ventricular Beats, (4) Unknown Beats, (5) Fusion Beats. We train a TTS-CGAN model to generate different categories of data using a set of real heartbeat signals. We use a set of real data as a testing set which is unseen for either real data classification model or GAN model training. Fig.~\ref{fig:CM} shows the confusion matrix of the testing classification results when using different data resources to train the classification model. And the Table~\ref{tbl:5000synthetic} lists each training method's corresponding classification scores. 

Fig.~\ref{subfig:realCM} shows the classification results for each category when the model was trained with all real signals with 5000 training samples. Fig.~\ref{subfig:synCM} shows the results when the model was trained on pure synthetic signals with also 5000 training samples. Even though the classification performance of method (b) lower than method (a), it still shows the synthetic data can teach the classification model similar information as the real data did. Fig.~\ref{subfig:realsmallCM} shows the results when the model was trained with only 1000 real data samples. It gets the worst classification scores because the training model cannot learn enough information from such small training set. 
Fig.~\ref{subfig:mixedCM} shows the results when we mix the real data and synthetic data. The experiment contains 1000 real data samples and 4000 synthetic data samples. It gets better results than using a small amount (1000 samples) of real data (Fig.~\ref{subfig:realsmallCM}) and only using pure synthetic data (Fig.~\ref{subfig:synCM}). It mimics the situation when we do not have enough amount of real data and use synthetic data to expand the small dataset size. The classification scores show that such synthetic data can be used as substitutes of real data when we do not have enough real data to train the deep learning models. 

\begin{table}[]
\centering
\begin{tabular}{c|cccc}
\hline
Class    & Method        & precision & recall & f1-score      \\ \hline
         & (a)           & 0.79      & 0.69   & 0.74          \\
1        & (b)           & 0.54      & 0.69   & 0.60          \\
         & (c)           & 0.54      & 0.47   & 0.50          \\
         & (d)           & 0.65      & 0.66   & 0.66          \\ \hline
         & (a)           & 0.80      & 0.82   & 0.81          \\
2        & (b)           & 0.83      & 0.62   & 0.71          \\
         & (c)           & 0.72      & 0.59   & 0.65          \\
         & (d)           & 0.82      & 0.73   & 0.78          \\ \hline
         & (a)           & 0.92      & 0.81   & 0.86          \\
3        & (b)           & 0.68      & 0.72   & 0.70          \\
         & (c)           & 0.66      & 0.63   & 0.64          \\
         & (d)           & 0.72      & 0.82   & 0.77          \\ \hline
         & (a)           & 0.80      & 0.92   & 0.86          \\
4        & (b)           & 0.84      & 0.83   & 0.84          \\
         & (c)           & 0.73      & 0.91   & 0.81          \\
         & (d)           & 0.85      & 0.85   & 0.85          \\ \hline
         & (a)           & 0.91      & 0.97   & 0.94          \\
5        & (b)           & 0.94      & 0.87   & 0.90          \\
         & (c)           & 0.81      & 0.90   & 0.85          \\
         & (d)           & 0.95      & 0.92   & 0.93          \\ \hline \hline
accuracy & (a)           & (b)       & (c)    & (d)           \\ \cline{2-5} 
         & \textbf{0.84} & 0.75      & 0.70   & \textbf{0.80} \\ \hline
\end{tabular}
\caption{The classification scores of training a classifier with different data sources and testing on the same dataset.}
\label{tbl:5000synthetic}
\end{table}

\section{Discussion and future study}
\label{sec:futurework}
Though our TTS-GAN and TTS-CGAN models are doing better in generating realistic multi-dimensional time-series data than many state-of-the-art time-series GAN models, compared to other GAN implementation fields, using GAN to generate time-series data is still at its early stage. We have the following questions that need further study. (1)~From Fig.~\ref{fig:real_syn_Signals}, we can clearly obverse that GAN model-generated data are noisier compared to real ones. The reason may be for GAN-generated synthetic data, each timestep is independently predicted but for real time-series, the values of consequent timesteps may imply some restrictions. It is worth to study how can we involve such restrictions on GAN-generated signals. (2)~Based on our empirical experiments, the quality of synthetic signals is different when the GAN models were trained with different types of time-series data.  For example, the stationary signals are much easier to generate, the single-dimensional signals are easier to generate than multi-dimensional ones, and the longer non-stationary signals are much harder to generate than shorter ones, etc. Therefore, it is necessary to have a comprehensive study that tests the generating synthetic data abilities and limitations of time-series GANs considering all types of time-series data. (3)~The more data samples used to train the GAN models, the more realistic synthetic data we usually get. It then becomes a chicken and egg situation, as that the GAN models ought to relieve the data shortage problem. We should study for different types of time-series data what is a proper amount of training data needed to generate good quality synthetic data.

\section{Conclusion}
\label{sec:conclusion}
In this work, we present two variants of a transformer-based GAN model can generate multi-dimensional time-series data of arbitrary length. The first model, TTS-GAN, can be trained on and generate sequences of one class, whereas, the second model, TTS-CGAN, employs label information as a conditional restriction to guide the GAN model to generate time-series data of multiple classes. Several visual comparison methods have been used to show the similarity between the original data and the synthetic data. A quantitative metric is introduced to compare the similarity of two sets of time-series signals. We also conduct a post hoc experiment to illustrate the usefulness of the GAN-generated data. We compare our work with a few state-of-the-art time-series GAN research works and the results of the experiments show that our models outperform all the others. 
 
%
\bibliographystyle{IEEEtran}
\bibliography{paperdraft}

\begin{thebibliography}{10}
\providecommand{\url}[1]{#1}
\csname url@samestyle\endcsname
\providecommand{\newblock}{\relax}
\providecommand{\bibinfo}[2]{#2}
\providecommand{\BIBentrySTDinterwordspacing}{\spaceskip=0pt\relax}
\providecommand{\BIBentryALTinterwordstretchfactor}{4}
\providecommand{\BIBentryALTinterwordspacing}{\spaceskip=\fontdimen2\font plus
\BIBentryALTinterwordstretchfactor\fontdimen3\font minus
  \fontdimen4\font\relax}
\providecommand{\BIBforeignlanguage}[2]{{%
\expandafter\ifx\csname l@#1\endcsname\relax
\typeout{** WARNING: IEEEtran.bst: No hyphenation pattern has been}%
\typeout{** loaded for the language `#1'. Using the pattern for}%
\typeout{** the default language instead.}%
\else
\language=\csname l@#1\endcsname
\fi
#2}}
\providecommand{\BIBdecl}{\relax}
\BIBdecl

\bibitem{lawhern2018eegnet}
V.~J. Lawhern, A.~J. Solon, N.~R. Waytowich, S.~M. Gordon, C.~P. Hung, and
  B.~J. Lance, ``Eegnet: a compact convolutional neural network for eeg-based
  brain--computer interfaces,'' \emph{Journal of neural engineering}, vol.~15,
  no.~5, p. 056013, 2018.

\bibitem{li2022spp}
X.~Li and V.~Metsis, ``Spp-eegnet: An input-agnostic self-supervised eeg
  representation model for inter-dataset transfer learning,'' in
  \emph{International Conference on Computing and Information
  Technology}.\hskip 1em plus 0.5em minus 0.4em\relax Springer, 2022, pp.
  173--182.

\bibitem{ramyachitra2014imbalanced}
D.~Ramyachitra and P.~Manikandan, ``Imbalanced dataset classification and
  solutions: a review,'' \emph{International Journal of Computing and Business
  Research (IJCBR)}, vol.~5, no.~4, pp. 1--29, 2014.

\bibitem{johnson2019survey}
J.~M. Johnson and T.~M. Khoshgoftaar, ``Survey on deep learning with class
  imbalance,'' \emph{Journal of Big Data}, vol.~6, no.~1, pp. 1--54, 2019.

\bibitem{goodfellow2014generative}
I.~Goodfellow, J.~Pouget-Abadie, M.~Mirza, B.~Xu, D.~Warde-Farley, S.~Ozair,
  A.~Courville, and Y.~Bengio, ``Generative adversarial nets,'' \emph{Advances
  in neural information processing systems}, vol.~27, 2014.

\bibitem{ledig2017photo}
C.~Ledig, L.~Theis, F.~Husz{\'a}r, J.~Caballero, A.~Cunningham, A.~Acosta,
  A.~Aitken, A.~Tejani, J.~Totz, Z.~Wang \emph{et~al.}, ``Photo-realistic
  single image super-resolution using a generative adversarial network,'' in
  \emph{Proceedings of the IEEE conference on computer vision and pattern
  recognition}, 2017, pp. 4681--4690.

\bibitem{bousmalis2017unsupervised}
K.~Bousmalis, N.~Silberman, D.~Dohan, D.~Erhan, and D.~Krishnan, ``Unsupervised
  pixel-level domain adaptation with generative adversarial networks,'' in
  \emph{Proceedings of the IEEE conference on computer vision and pattern
  recognition}, 2017, pp. 3722--3731.

\bibitem{zhang2017stackgan}
H.~Zhang, T.~Xu, H.~Li, S.~Zhang, X.~Wang, X.~Huang, and D.~N. Metaxas,
  ``Stackgan: Text to photo-realistic image synthesis with stacked generative
  adversarial networks,'' in \emph{Proceedings of the IEEE international
  conference on computer vision}, 2017, pp. 5907--5915.

\bibitem{brophy2021generative}
E.~Brophy, Z.~Wang, Q.~She, and T.~Ward, ``Generative adversarial networks in
  time series: A survey and taxonomy,'' \emph{arXiv preprint arXiv:2107.11098},
  2021.

\bibitem{mogren2016c}
O.~Mogren, ``C-rnn-gan: Continuous recurrent neural networks with adversarial
  training,'' \emph{arXiv preprint arXiv:1611.09904}, 2016.

\bibitem{esteban2017real}
C.~Esteban, S.~L. Hyland, and G.~R{\"a}tsch, ``Real-valued (medical) time
  series generation with recurrent conditional gans,'' \emph{arXiv preprint
  arXiv:1706.02633}, 2017.

\bibitem{yoon2019time}
J.~Yoon, D.~Jarrett, and M.~Van~der Schaar, ``Time-series generative
  adversarial networks,'' 2019.

\bibitem{ni2020conditional}
H.~Ni, L.~Szpruch, M.~Wiese, S.~Liao, and B.~Xiao, ``Conditional
  sig-wasserstein gans for time series generation,'' \emph{arXiv preprint
  arXiv:2006.05421}, 2020.

\bibitem{vaswani2017attention}
A.~Vaswani, N.~Shazeer, N.~Parmar, J.~Uszkoreit, L.~Jones, A.~N. Gomez,
  {\L}.~Kaiser, and I.~Polosukhin, ``Attention is all you need,'' in
  \emph{Advances in neural information processing systems}, 2017, pp.
  5998--6008.

\bibitem{dosovitskiy2020image}
A.~Dosovitskiy, L.~Beyer, A.~Kolesnikov, D.~Weissenborn, X.~Zhai,
  T.~Unterthiner, M.~Dehghani, M.~Minderer, G.~Heigold, S.~Gelly \emph{et~al.},
  ``An image is worth 16x16 words: Transformers for image recognition at
  scale,'' \emph{arXiv preprint arXiv:2010.11929}, 2020.

\bibitem{devlin2018bert}
J.~Devlin, M.-W. Chang, K.~Lee, and K.~Toutanova, ``Bert: Pre-training of deep
  bidirectional transformers for language understanding,'' \emph{arXiv preprint
  arXiv:1810.04805}, 2018.

\bibitem{lu2021pretrained}
K.~Lu, A.~Grover, P.~Abbeel, and I.~Mordatch, ``Pretrained transformers as
  universal computation engines,'' \emph{arXiv preprint arXiv:2103.05247},
  2021.

\bibitem{jiang2021transgan}
Y.~Jiang, S.~Chang, and Z.~Wang, ``Transgan: Two pure transformers can make one
  strong gan, and that can scale up,'' in \emph{Thirty-Fifth Conference on
  Neural Information Processing Systems}, 2021.

\bibitem{diao2021tilgan}
S.~Diao, X.~Shen, K.~Shum, Y.~Song, and T.~Zhang, ``Tilgan: Transformer-based
  implicit latent gan for diverse and coherent text generation,'' in
  \emph{Findings of the Association for Computational Linguistics: ACL-IJCNLP
  2021}, 2021, pp. 4844--4858.

\bibitem{li2022tts}
X.~Li, V.~Metsis, H.~Wang, and A.~H.~H. Ngu, ``Tts-gan: A transformer-based
  time-series generative adversarial network,'' \emph{arXiv preprint
  arXiv:2202.02691}, 2022.

\bibitem{grinsted2004application}
A.~Grinsted, J.~C. Moore, and S.~Jevrejeva, ``Application of the cross wavelet
  transform and wavelet coherence to geophysical time series,'' \emph{Nonlinear
  processes in geophysics}, vol.~11, no. 5/6, pp. 561--566, 2004.

\bibitem{ratliff2013characterization}
L.~J. Ratliff, S.~A. Burden, and S.~S. Sastry, ``Characterization and
  computation of local nash equilibria in continuous games,'' in \emph{2013
  51st Annual Allerton Conference on Communication, Control, and Computing
  (Allerton)}.\hskip 1em plus 0.5em minus 0.4em\relax IEEE, 2013, pp. 917--924.

\bibitem{goodfellow2020generative}
I.~Goodfellow, J.~Pouget-Abadie, M.~Mirza, B.~Xu, D.~Warde-Farley, S.~Ozair,
  A.~Courville, and Y.~Bengio, ``Generative adversarial networks,''
  \emph{Communications of the ACM}, vol.~63, no.~11, pp. 139--144, 2020.

\bibitem{huang2017beyond}
R.~Huang, S.~Zhang, T.~Li, and R.~He, ``Beyond face rotation: Global and local
  perception gan for photorealistic and identity preserving frontal view
  synthesis,'' in \emph{Proceedings of the IEEE international conference on
  computer vision}, 2017, pp. 2439--2448.

\bibitem{mirza2014conditional}
M.~Mirza and S.~Osindero, ``Conditional generative adversarial nets,''
  \emph{arXiv preprint arXiv:1411.1784}, 2014.

\bibitem{karras2019style}
T.~Karras, S.~Laine, and T.~Aila, ``A style-based generator architecture for
  generative adversarial networks,'' in \emph{Proceedings of the IEEE/CVF
  conference on computer vision and pattern recognition}, 2019, pp. 4401--4410.

\bibitem{zhu2017unpaired}
J.-Y. Zhu, T.~Park, P.~Isola, and A.~A. Efros, ``Unpaired image-to-image
  translation using cycle-consistent adversarial networks,'' in
  \emph{Proceedings of the IEEE international conference on computer vision},
  2017, pp. 2223--2232.

\bibitem{choi2018stargan}
Y.~Choi, M.~Choi, M.~Kim, J.-W. Ha, S.~Kim, and J.~Choo, ``Stargan: Unified
  generative adversarial networks for multi-domain image-to-image
  translation,'' in \emph{Proceedings of the IEEE conference on computer vision
  and pattern recognition}, 2018, pp. 8789--8797.

\bibitem{cassisi2012similarity}
C.~Cassisi, P.~Montalto, M.~Aliotta, A.~Cannata, and A.~Pulvirenti,
  ``Similarity measures and dimensionality reduction techniques for time series
  data mining,'' \emph{Advances in data mining knowledge discovery and
  applications’(InTech, Rijeka, Croatia, 2012,}, pp. 71--96, 2012.

\bibitem{arjovsky2017wasserstein}
M.~Arjovsky, S.~Chintala, and L.~Bottou, ``Wasserstein generative adversarial
  networks,'' in \emph{International conference on machine learning}.\hskip 1em
  plus 0.5em minus 0.4em\relax PMLR, 2017, pp. 214--223.

\bibitem{gulrajani2017improved}
I.~Gulrajani, F.~Ahmed, M.~Arjovsky, V.~Dumoulin, and A.~C. Courville,
  ``Improved training of wasserstein gans,'' \emph{Advances in neural
  information processing systems}, vol.~30, 2017.

\bibitem{deng2012mnist}
L.~Deng, ``The mnist database of handwritten digit images for machine learning
  research,'' \emph{IEEE Signal Processing Magazine}, vol.~29, no.~6, pp.
  141--142, 2012.

\bibitem{wchowebsite}
\BIBentryALTinterwordspacing
Mathworks. Wavelet coherence matlab tutorial. [Online]. Available:
  \url{https://www.mathworks.com/help/wavelet/ug/compare-time-frequency-content-in-signals-with-wavelet-coherence.html}
\BIBentrySTDinterwordspacing

\bibitem{cui2012nirs}
X.~Cui, D.~M. Bryant, and A.~L. Reiss, ``Nirs-based hyperscanning reveals
  increased interpersonal coherence in superior frontal cortex during
  cooperation,'' \emph{Neuroimage}, vol.~59, no.~3, pp. 2430--2437, 2012.

\bibitem{app7101101}
D.~Micucci, M.~Mobilio, and P.~Napoletano, ``Unimib shar: A dataset for human
  activity recognition using acceleration data from smartphones,''
  \emph{Applied Sciences}, vol.~7, no.~10, 2017.

\bibitem{bousseljot1995nutzung}
R.~Bousseljot, D.~Kreiseler, and A.~Schnabel, ``Nutzung der ekg-signaldatenbank
  cardiodat der ptb {\"u}ber das internet,'' 1995.

\bibitem{goldberger2000physiobank}
A.~L. Goldberger, L.~A. Amaral, L.~Glass, J.~M. Hausdorff, P.~C. Ivanov, R.~G.
  Mark, J.~E. Mietus, G.~B. Moody, C.-K. Peng, and H.~E. Stanley, ``Physiobank,
  physiotoolkit, and physionet: components of a new research resource for
  complex physiologic signals,'' \emph{circulation}, vol. 101, no.~23, pp.
  e215--e220, 2000.

\bibitem{moody2001impact}
G.~B. Moody and R.~G. Mark, ``The impact of the mit-bih arrhythmia database,''
  \emph{IEEE Engineering in Medicine and Biology Magazine}, vol.~20, no.~3, pp.
  45--50, 2001.

\bibitem{NEURIPS2019_9015}
A.~Paszke, S.~Gross, F.~Massa, A.~Lerer, J.~Bradbury, G.~Chanan, T.~Killeen,
  Z.~Lin, N.~Gimelshein, L.~Antiga, A.~Desmaison, A.~Kopf, E.~Yang, Z.~DeVito,
  M.~Raison, A.~Tejani, S.~Chilamkurthy, B.~Steiner, L.~Fang, J.~Bai, and
  S.~Chintala, ``Pytorch: An imperative style, high-performance deep learning
  library,'' in \emph{Advances in Neural Information Processing Systems 32},
  H.~Wallach, H.~Larochelle, A.~Beygelzimer, F.~d\textquotesingle
  Alch\'{e}-Buc, E.~Fox, and R.~Garnett, Eds.\hskip 1em plus 0.5em minus
  0.4em\relax Curran Associates, Inc., 2019, pp. 8024--8035.

\end{thebibliography}

\section{Biography Section}

\begin{IEEEbiography}[{\includegraphics[width=1in,height=1.25in,clip,keepaspectratio]{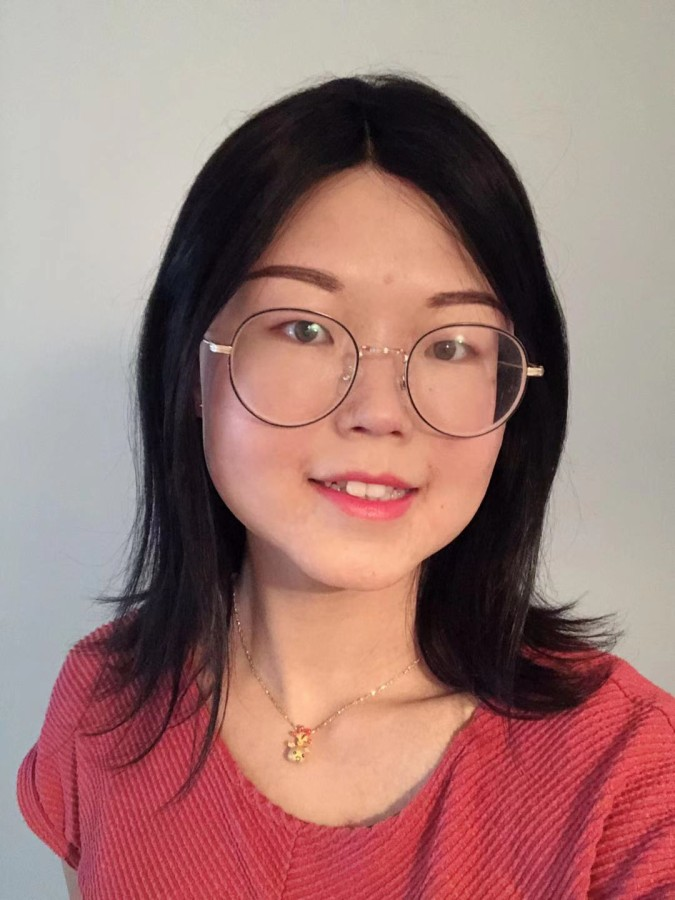}}]{Xiaomin Li}
She is a Ph.D. candidate at Texas State University. She received her Bachelor of Science degree in Computer Science in 2018, from Lanzhou University in China. Her research interest is in time-series data analysis using machine learning algorithms, specifically on bio-signals and human health data. She is familiar with many state-of-the-art deep learning algorithms and models, such as self-supervised learning, transfer learning, generative adversarial networks, transformers, etc. She also has a deep understanding of the fields  of computer vision, edge computing, high-performance computing, etc. 
\end{IEEEbiography}

\begin{IEEEbiography}[{\includegraphics[width=1in,height=1.25in,clip,keepaspectratio]{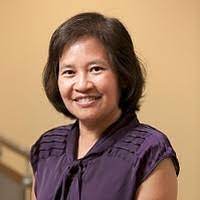}}]{Anne Hee Hiong Ngu} Dr. Ngu is a professor and the Ph.D. Program director with the Department of Computer Science at Texas State University. She earned her Ph.D. in Computer Science from the University of Western Australia in 1990. 
She was the recipient of the Undergraduate Research Mentoring Award from 
NCWIT, for her outstanding mentorship to encourage and advance undergraduates in computing fields in 2013. Her main research interests are in information integration, Internet of Things, health informatics, scientific workflows, service computing, databases, and agent technologies.
\end{IEEEbiography}

\begin{IEEEbiography}[{\includegraphics[width=1in,height=1.25in,clip,keepaspectratio]{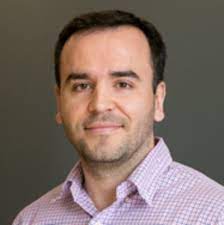}}]{Vangelis Metsis} Dr. Metsis is an Associate Professor at the Department of Computer Science at Texas State University. He received his Bachelor of Science degree in Computer Science in 2005, from the Department of Informatics of Athens University of Economics and Business in Greece, and his Doctoral degree in 2011 from the Department of Computer Science and Engineering of The University of Texas at Arlington. His research interests span the areas of machine learning and computer vision with focus in applications of smart health, pervasive computing, and VR/AR.
\end{IEEEbiography}


\end{document}